\documentclass[11pt]{article}

\usepackage[final]{acl}
\usepackage{hyperref}
\usepackage{url}
\usepackage{booktabs}
\usepackage{capt-of}
\usepackage{amsmath}
\usepackage{multirow}
\usepackage{booktabs}
\usepackage{rotating}

\usepackage[table]{xcolor}  
\usepackage{colortbl}       

\usepackage[utf8]{inputenc}
\usepackage{algorithm}
\usepackage{algpseudocode} 
\usepackage{amsmath}
\usepackage{amssymb}
\usepackage{xcolor}
\usepackage{wrapfig}
\usepackage{subcaption}
\usepackage{tikz}
\usetikzlibrary{matrix, positioning}

\usepackage{times}
\usepackage{latexsym}

\usepackage[T1]{fontenc}

\usepackage[utf8]{inputenc}

\usepackage{microtype}

\usepackage{inconsolata}

\usepackage{graphicx}

\usepackage{placeins}
\usepackage{makecell}

\usepackage{lineno}
\usepackage[hide]{todo}

\definecolor{darkblue}{rgb}{0, 0, 0.5}
\hypersetup{colorlinks=true, citecolor=darkblue, linkcolor=darkblue, urlcolor=darkblue}

\defcitealias{nllbteam2022languageleftbehindscaling}{NLLB Team, 2022}
\defcitealias{gemmateam2025gemma3technicalreport}{GemmaTeam, 2025}
\defcitealias{gemmateam2024gemma2improvingopen}{Gemma Team, 2024}

\title{To Each Language Its Tokenizer: Modular Tokenizers for Efficient Multilingual LLMs}

\author{
 \textbf{Franck Signe\textsuperscript{1,2}},
 \textbf{Hippolyte Pilchen\textsuperscript{1,3}},
 \textbf{François Yvon\textsuperscript{2}},
 \textbf{\'Edouard Grave\textsuperscript{1}}
\\
\\
 \textsuperscript{1}Kyutai, Paris\\
 \textsuperscript{2}Sorbonne Université, CNRS, Institute of Intelligent Systems and Robotics, Paris\\
 \textsuperscript{3}Univ. Grenoble Alpes, CNRS, Grenoble INP, LIG, Grenoble\\
\\
 \small{
   \textbf{Correspondence:} \href{mailto:email@domain}{franck.signe-talla@kyutai.org}
 }
}

\newcommand{\Nosampling}{\emph{Mono-only}}
\newcommand{\Full}{\emph{$\text{Samp}_{\operatorname{Full}}$}}
\newcommand{\Data}[1]{\emph{$\text{Samp}_{\operatorname{Sub}#1}$}}

\begin{document}
\maketitle
\begin{abstract}
Multilingual Large Language Models (LLMs) traditionally rely on a single vocabulary shared by all supported languages, which can lead to uneven compression across them. Moreover, their large embedding and output matrices increase memory usage and slow inference, notably for small-scale models. It is also wasteful as models are often used for only a subset of languages. To address these issues, we introduce a modular framework for multilingual model training. First, we propose methods to learn large modular BPE and Unigram tokenizers that enable extraction of subtokenizers tailored to any language subset. These subtokenizers achieve compression on par with monolingual tokenizers and improve cross-lingual fairness. Second, we design a pretraining strategy that samples subtokenizers to form batches, restricting predictions to the relevant vocabulary subset and allowing efficient training despite a large vocabulary. This supports efficient inference with any combination of language-specific vocabularies. Therefore, it reduces memory usage and speeds up inference in models without sacrificing performance.\footnote{The code is available at \url{https://github.com/kyutai-labs/modular-tokenization}.}

\end{abstract}

\section{Introduction \label{sec:introduction}}

\begin{figure}[t]
        \centering
        \includegraphics[width=\linewidth, trim={0.5em 0.5em 0.5em 0}, clip]
        {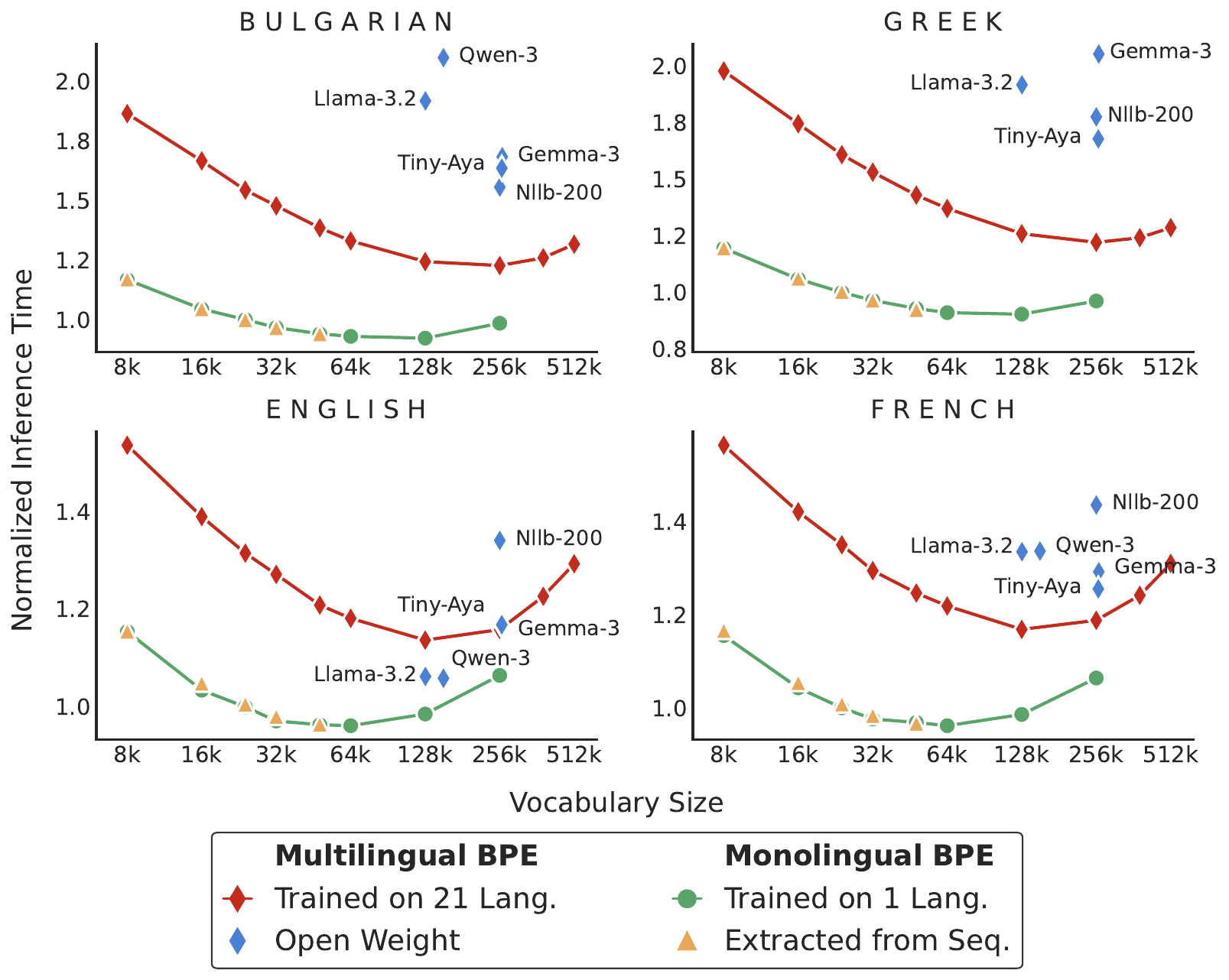}
        \captionof{figure}{Normalized inference time on FLORES 200 for a 2.5B non-embedding parameter model using monolingual and multilingual tokenizers of varying sizes. $24k$-token monolingual tokenizers are used as reference. For each size, \textit{Extracted} denotes monolingual tokenizers derived from a sequentially trained multilingual BPE model (\textsection{}~\ref{sec:bpe_tokenizer}), with a per-language budget of that size across 21 languages. Baseline open-weight tokenizers cover all reported languages.}      
        \label{fig:BPE inference cost}
\end{figure}

Tokenization is the interface between raw text and the embedding space of Large Language Models (LLMs), directly influencing vocabulary coverage, sequence length, and computational efficiency. In multilingual settings, tokenizers must cover diverse scripts and linguistic patterns, often requiring large vocabularies. In smaller models, the embedding matrix can account for a disproportionate share of total parameters — for example, Gemma~2 2.6B \citep{gemmateam2024gemma2improvingopen} devotes 22\% of its parameters to a 256k-token vocabulary, even with shared input and output embeddings. This is particularly wasteful in resource-constrained settings, where models are often deployed for specific language(s) and the full multilingual output matrix need not be loaded, leaving most tokens redundant and wasting memory and compute. Beyond computational inefficiencies, standard multilingual tokenizers also induce compression disparities across languages: identical content can yield drastically different tokenization lengths depending on the language \citep{petrov2023language}, leading to unequal effective context windows and higher computational costs for non-English text. Together, these issues make inference with large multilingual tokenizers covering many languages slower and more expensive than with language-specific ones, as illustrated in Fig.~\ref{fig:BPE inference cost}.

To address both challenges at pretraining, we propose two complementary contributions. First, a method to construct a modular multilingual tokenizer from which specialized subtokenizers can be extracted, each covering a specific language or subset of languages, with compression rates comparable to independently trained monolingual tokenizers, provided for both BPE \citep{gage-etal-1994-new, sennrich-etal-2016-neural} and Unigram \citep{kudo-2018-subword}. Second, a pretraining procedure that trains the model to operate with any combination of subtokenizers tailored to the language(s) of interest, allowing irrelevant tokens to be excluded from the output matrix at inference. By maintaining high compression for target languages, this yields faster inference than standard multilingual tokenizers while retaining competitive performance on multiple-choice evaluation, summarization, and translation.

\section{Related Work \label{sec:related}} 

\paragraph{Unfairness in Tokenization.} 

Multilingual tokenizers often compress languages unevenly: identical content translated from English can yield widely varying token lengths, with some low-resource languages requiring up to 15~times more tokens \citep{petrov2023language}, raising computational costs and penalizing them in token-based pricing schemes \citep{ahia-etal-2023-languages}. These disparities are not limited to multilingual settings: \citet{arnett2025explaining} show that monolingual BPE tokenizers trained under identical conditions (parallel content-matched corpora, identical vocabulary sizes and training procedures) still exhibit different compression rates across languages. They attribute this largely to whitespace-based pre-tokenization which prevents merges across word boundaries, a limitation that superword tokenizers \citep{liu2025superbpe, schmidt2025boundless} address by allowing merges across whitespace boundaries.

To address these disparities, \citet{foroutan2025parityawarebytepairencodingimproving} propose a BPE variant that selects, at each step, the merge maximizing compression for the worst-compressed language. Other work leverages byte-level models: \citet{ahia2024magnet} propose adaptive gradient-based tokenization to group bytes, and \citet{limisiewicz-etal-2024-myte} modify UTF-8 representations based on morphology to balance compression across languages. \citet{zheng-etal-2021-allocating} explicitly allocate vocabulary capacity across languages, increasing the share of underrepresented languages. \citet{chung-etal-2020-improving, liang-etal-2023-xlm} group linguistically similar languages to reduce harmful vocabulary sharing, training tokenizers for each cluster before merging them for multilingual coverage, and use these different tokenizers to train encoder-based models. In contrast, our approach enables flexible inference with any combination of specialized vocabularies, without sacrificing performance, notably on generation tasks.

\paragraph{Vocabulary Adaptation.}

The computational efficiency of a pretrained model on a target domain depends heavily on vocabulary suitability, motivating two adaptation strategies: vocabulary expansion to reduce sequence length, and vocabulary reduction to lower computational overhead.

Expansion methods append language-specific tokens to pretrained vocabularies \citep{tejaswi-etal-2024-exploring, csaki-etal-2024-sambalingo, cui2024efficienteffectivetextencoding, kiulian-etal-2025-english, fujii2024continual, lin2024mala500massivelanguageadaptation, dagan2024getting}, achieving more compact representations, lower tokenizer fertility, and higher throughput. Usually, a dedicated tokenizer is trained on the target distribution and its new tokens are appended to the original vocabulary, but existing merge rules can prevent these tokens from being formed, making this approach suboptimal. 
\citet{purason-etal-2026-teaching} address this by continuing the training of the original tokenizer on the target distribution, refining its merge table rather than simply appending new tokens.

Reduction methods adapt vocabularies to target language(s) by pruning irrelevant tokens \citep{ushio-etal-2023-efficient, williams-aletras-2025-vocabulary,pilana-liyanage-yvon-2026-adaptbpe,zhang2026vocabtailordynamicvocabularyselection} or replacing them with domain-specific ones \citep{nakash2025adaptivocab, chijiwa2026lossless}, reducing memory and improving inference speed. The latter requires careful embedding initialization \citep{minixhofer-etal-2022-wechsel, gee-etal-2022-fast, dobler-de-melo-2023-focus, samuel-etal-2025-small, nakash2025adaptivocab}, and both require a fine-tuning step to maintain performance. Unlike prior work, we tackle these issues during pretraining rather than post-hoc, enabling vocabulary reduction gains to be realized effortlessly at inference.

\section{Inefficiencies of Multilingual Tokenizers 
\label{sec:inefficiency}}
A tokenizer $\mathcal{T}$ is formally defined as a tuple $(\mathcal{V}, \tau, \delta)$, where $\mathcal{V}$ is a finite vocabulary of tokens \citep{zouhar-etal-2023-formal}. The tokenization function $\tau: \Sigma^* \to \mathcal{V}^*$ maps a string $S$ defined  over a finite alphabet $\Sigma$ to a sequence of tokens, which we denote for simplicity as $\mathcal{T}(S) = \tau(S)$. The detokenization function $\delta: \mathcal{V}^* \to \Sigma^*$ maps a token sequence back to a string over the original alphabet, with most modern tokenizers satisfying the \textit{lossless constraint} $\delta(\mathcal{T}(S)) = S$ for all $S \in \Sigma^*$.

\paragraph{Tokenizer Training.} BPE tokenizers are trained with the Hugging Face tokenizers library\footnote{\url{https://huggingface.co/docs/transformers/main_classes/tokenizer}} and Unigram tokenizers with SentencePiece,\footnote{\url{https://github.com/google/sentencepiece}} both using whitespace pre-tokenization with byte-level fallback and no normalization. Monolingual tokenizers are trained on language-specific CommonCrawl subsets;\footnote{\url{https://data.commoncrawl.org/crawl-data/CC-MAIN-2024-10/index.html}} multilingual ones on the concatenation of these subsets, ensuring identical per-language data exposure in both settings.

\subsection{Monolingual Tokenizers Better Compress than Multilingual Ones}

Multilingual LLMs require tokenizers with broad language coverage to handle diverse scripts and linguistic structures, but larger vocabularies increase the computational cost of the output layer, forcing tokenizers to operate under limited capacity. This constraint disproportionately affects under-represented languages, which tend to be poorly compressed \citep{zheng-etal-2021-allocating}. To quantify the compression efficiency of a tokenizer on a given language $\mathcal{L}$, we use the \textbf{Normalized Sequence Length (NSL)} \citep{dagan2024getting}, defined as the ratio of tokenized sequence length to that of a reference tokenizer. For a dataset $\mathcal{D}$ with $M$ examples in language $\mathcal{L}$, NSL is defined as:
\begin{align}
\text{NSL} & = 
\frac{\sum_{i=1}^{M} \mathrm{length}\big(\mathcal{T}(\mathcal{D}[i])\big)}
     {\sum_{i=1}^{M} \mathrm{length}\big(\mathcal{T}_{\text{ref}}^l(\mathcal{D}[i])\big)}, \label{eq:nsl}
\end{align}
where $\mathcal{T}_{\text{ref}}^l$ is a monolingual reference tokenizer of fixed vocabulary size trained exclusively on $\mathcal{L}$.

\begin{table}[t]
\footnotesize
\renewcommand{\arraystretch}{1.2}
\centering

\begin{tabular}{lcccccc}
\toprule
\multicolumn{2}{c}{} & \multicolumn{5}{c}{NSL ($\downarrow$)}\\ 
Model & Size & BG & EL & EN & FI & FR \\
\midrule
TinyAya & 261k & 1.38 & 1.42 & 0.98 & 1.35 & 1.06 \\
NLLB    & 256k & 1.33 & 1.51 & 1.14 & 1.30 & 1.21 \\
Gemma3  & 262k & 1.42 & 1.73 & 0.98 & 1.54 & 1.09 \\
Llama3  & 128k & 1.78 & 1.78 & 0.98 & 1.83 & 1.24 \\
Qwen3   & 152k & 1.92 & 3.90 & 0.96 & 1.82 & 1.22 \\
\midrule
\multirow{2}{*}{Multi 21} 
& 128k & 1.16 & 1.17 & 1.06 & 1.14 & 1.09 \\
& 256k & 1.04 & 1.04 & 0.98 & 1.02 & 1.01\\
\bottomrule
\end{tabular}

\caption{NSL relative to a 24k-token monolingual BPE reference, for open-weight tokenizers and a standard multilingual BPE jointly trained on our 21 languages (\emph{Multi 21}). More languages in Appendix~\ref{appendix:bpe_tokenizer}.}
\label{tab:nsl_table}

\end{table}

Tab.~\ref{tab:nsl_table} reports $\text{NSL}$ on FLORES-200 (devtest, 1012 lines) for multilingual tokenizers relative to monolingual reference with 24k tokens. We focus on BPE, since the open-weight tokenizers are BPE-based, ensuring a uniform evaluation; the same conclusions hold for Unigram tokenizers (see Appendix~\ref{appendix:unigram_nsl}). Open-weight tokenizers — TinyAya~\citep{salamanca2026tinyayabridgingscale}, Nllb-200~\citep{costajussa-etal-2024-scaling}, Gemma-3~\citep{gemmateam2025gemma3technicalreport}, Llama-3~\citep{grattafiori2024llama3herdmodels}, and Qwen-3~\citep{yang2025qwen3technicalreport} — achieve compression comparable to monolingual references for English but underperform for most other languages. By spreading a fixed vocabulary across many languages, these tokenizers end up providing poor compression for the majority of languages they aim to support.

Our evaluation set is deliberately conservative: 21 European languages (Appendix~\ref{appendix:list_of_languages}), medium- to high-resource and well covered by open-weight multilingual tokenizers — unfairness thus appears even in a setting that favors these tokenizers, and it widens sharply on low-resource languages (Tab.~\ref{tab:nsl_lowres}). Covering fewer languages leaves more vocabulary per language: \emph{Multi 21}, a standard BPE trained on our 21 languages only, comes within 4\% of monolingual compression at 256k tokens — a size comparable to the largest open-weight tokenizers in Tab.~\ref{tab:nsl_table}. This \emph{trade-off} between coverage and per-language compression motivates specialized tokenizers over a single large multilingual one.

While better compression allows more data to be processed for the same compute budget at training, its impact on model quality remains contentious: some work finds a positive correlation with downstream performance \citep{galle-2019-investigating, goldman-etal-2024-unpacking}, while others find that increased compression fails to yield performance gains \citep{schmidt-etal-2024-tokenization, dagan2024getting}. In this work, we therefore pursue improved compression primarily for efficiency rather than performance gains.

\tikzset{
    tokenstyle/.style 2 args={
        fill=#2,
        draw=gray!40,
        rounded corners=1pt,
        inner sep=2pt,
        font=\ttfamily\small,
        baseline=(char.base),
        minimum height=1.4em, 
        text height=1ex,
        text depth=0.3ex,
        minimum width=#1
    }
}

\newcommand{\token}[2][0pt]{%
    \begin{tikzpicture}[remember picture, baseline=(char.base)]
        \node[tokenstyle={#1}{gray!15}] (char) {#2};
    \end{tikzpicture}\kern2pt
}


\newcommand{\tokenblue}[2][0pt]{%
    \begin{tikzpicture}[remember picture, baseline=(char.base)]
        \node[tokenstyle={#1}{blue!15}] (char) {#2};
    \end{tikzpicture}\kern2pt
}

\begin{figure*}
\centering
\begin{tikzpicture}
    \matrix (m) [
        matrix of nodes,
        column sep=0.6cm, 
        row sep=-0.2em, 
        nodes={anchor=west}, 
    ]{
        \hspace{6em} \textbf{Finnish BPE } & \hspace{4em} \textbf{English + Finnish BPE} \\
        
        \token{\_} \hspace{0.01em} \token{a} \hspace{0.01em} \token{i} \hspace{0.01em} \token{k} \hspace{0.01em} \token{a} \hspace{0.01em} \token{i} \hspace{0.01em} \token{s} \hspace{0.01em} \token{e} \hspace{0.01em} \token{m} \hspace{0.01em} \token{m} \hspace{0.01em} \token{i} \hspace{0.01em} \token{n} & 
        
        \token{\_} \hspace{0.01em} \token{a} \hspace{0.01em} \token{i} \hspace{0.01em} \token{k} \hspace{0.01em} \token{a} \hspace{0.01em} \token{i} \hspace{0.01em} \token{s} \hspace{0.01em} \token{e} \hspace{0.01em} \token{m} \hspace{0.01em} \token{m} \hspace{0.01em} \token{i} \hspace{0.01em} \token{n} \\
        \token[2.43em]{\_a} \hspace{0.01em} \tokenblue{i} \hspace{0.01em} \tokenblue[2.43em]{ka} \hspace{0.01em} \token[2.43em]{is} \hspace{0.01em} \tokenblue{e} \hspace{0.01em} \tokenblue[2.43em]{mm} \hspace{0.01em} \token[2.43em]{in} & 
        \token[2.43em]{\_a} \hspace{0.01em} \tokenblue[2.43em]{ik} \hspace{0.01em} \tokenblue{a} \hspace{0.01em} \token[2.43em]{is} \hspace{0.01em} \tokenblue[2.43em]{em} \hspace{0.01em}  \tokenblue{m} \hspace{0.01em}  \token[2.43em]{in} \\
        \token[2.43em]{\_a} \hspace{0.01em} \token[4.1em]{ika} \hspace{0.01em} \tokenblue[4.1em]{ise} \hspace{0.01em} \tokenblue[5.5em]{mmin} & 
        \token[2.43em]{\_a} \hspace{0.01em} \token[4.1em]{ika} \hspace{0.01em} \tokenblue[2.43em]{is} \hspace{0.01em} \tokenblue[2.43em]{em} \hspace{0.01em} \tokenblue[4em]{min} \\
        \token[7.4em]{\_aika} \hspace{0.01em} \tokenblue[10.2em]{isemmin} & 
        \token[7.4em]{\_aika} \hspace{0.01em} \tokenblue[2.43em]{is} \hspace{0.01em} \tokenblue[2.43em]{em} \hspace{0.01em} \tokenblue[4em]{min} \\
        \tokenblue[18.5em]{\_aikaisemmin} & 
        \tokenblue[7.4em]{\_aika} \hspace{0.01em} \tokenblue[2.43em]{is} \hspace{0.01em} \tokenblue[2.43em]{em} \hspace{0.01em} \tokenblue[4em]{min} \\
    };

\end{tikzpicture}
\caption{\textbf{Comparison of BPE merge hierarchies.} The Finnish-only tokenizer (left) forms the token \textit{\_aikaisemmin}. In contrast, the joint English--Finnish tokenizer (right), where Finnish rules are appended to English ones, produces suboptimal units: early English merges \textit{e+m} and \textit{m+in} prevent \textit{\_aikaisemmin} from forming despite its presence in the vocabulary. Our \textit{Sequential training} mitigates this by continuing training the English tokenizer on Finnish data, creating merges \textit{em+min}, \textit{is+emmin}, and \textit{\_aika+isemmin} that ultimately recover the token \textit{\_aikaisemmin}.}
\label{fig:inefficient_bpe_merge}
\end{figure*}
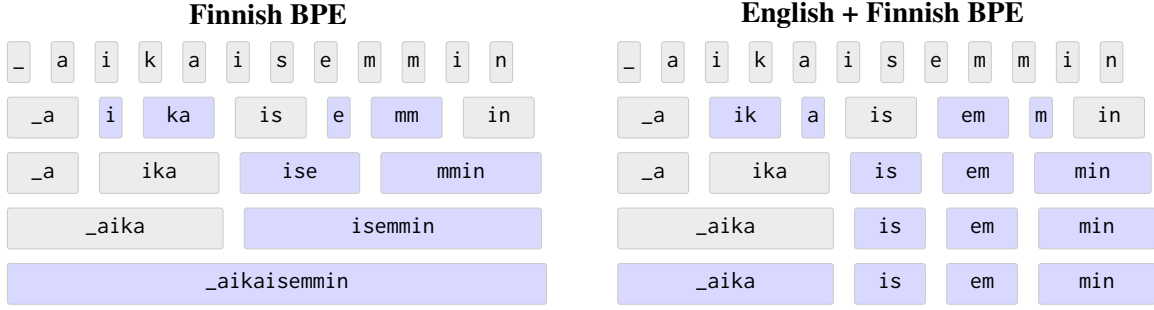

\subsection{Monolingual Tokenizers Enable Cheaper Inference than Multilingual Ones}

The previously discussed \emph{trade-off} in multilingual coverage often forces large-coverage tokenizers to adopt extensive vocabularies to maintain competitive inference speeds. As shown in Fig.~\ref{fig:BPE inference cost}, initial vocabulary expansions bring substantial speed gains, but these quickly plateau as marginal compression improvements are offset by a growing output-layer cost. This cost is especially pronounced in small-scale models, where the embedding and output matrices represent a large share of total parameters, amplifying both memory overhead and inference latency. Fig.~\ref{fig:BPE inference cost} suggests that monolingual tokenizers sidestep this issue, achieving faster inference with much smaller vocabularies.

\section{Building Modular Tokenizers \label{sec:modular}}

Given a global tokenizer $\mathcal{T}=(\mathcal{V}, \tau, \delta)$, we define a \textbf{subtokenizer} $\mathcal{T}_s=(\mathcal{V}_s, \tau_s, \delta_s)$ where $\mathcal{V}_s \subseteq \mathcal{V}$. To leverage the advantages of monolingual tokenizers within a multilingual framework, we aim to design $\mathcal{T}$ covering languages $\{\mathcal{L}_1, \dots, \mathcal{L}_N\}$ that satisfies: \textbf{(i) Extractability:} for each $\mathcal{L}_i$, a specialized subtokenizer $\mathcal{T}_i=(\mathcal{V}_i, \tau_i, \delta_i)$ can be explicitly derived; \textbf{(ii) Compression Efficiency:} each $\mathcal{T}_i$ matches the compression of a tokenizer trained only on $\mathcal{L}_i$; and \textbf{(iii) Compositionality:} for any subset $\{\mathcal{L}_1, \dots, \mathcal{L}_k\}$, a unified subtokenizer $\mathcal{T}_s$ can be built via $\mathcal{V}_s = \bigcup_{i=1}^k \mathcal{V}_i$, maintaining competitive compression across all included  
languages.

\subsection{UnigramLM Tokenizer \label{sec:unigram}}

The UnigramLM tokenization function assigns to each token in $\mathcal{V}$ a precomputed probability, assuming independence between token occurrences \citep{kudo-2018-subword, meister2025unigramlm}. Given a string $S$, it finds the most probable sequence whose concatenation equals $S$, via the Viterbi algorithm \citep{1054010}. A subtokenizer $\mathcal{T}_s$ restricts this to $\mathcal{V}_s$ by assigning zero probability to tokens $v \notin \mathcal{V}_s$, while keeping original probabilities for $v \in \mathcal{V}_s$. No renormalization is needed, as rescaling does not affect the optimal Viterbi path. To build the vocabularies $\mathcal{V}_i$, we consider two main approaches.

\paragraph{Frequency-Based Extraction.} Given a multilingual tokenizer $\mathcal{T}$, a language-specific vocabulary $\mathcal{V}_i$ of size $k$ is built by retaining the $k$ most frequent tokens on a corpus in $\mathcal{L}_i$. Even with $k$ much smaller than $|\mathcal{V}|$, this approach preserves compression comparable to $\mathcal{T}$ (Appendix~\ref{appendix:frequency_based}, Tab~\ref{tab:more_nsl_table_freq}), and the reduced vocabulary size directly translates into lower inference latency (Appendix~\ref{appendix:inference_time_comparison_unigram}). However, the compression of the resulting subtokenizer is inherently bounded by that of $\mathcal{T}$.

\paragraph{Merging Monolingual Tokenizers.} Monolingual tokenizers are trained on each corpus $\mathcal{D}_i$, and their vocabularies are merged into $\mathcal{V} = \bigcup \mathcal{V}_i$. Token probabilities are then re-estimated over the combined corpus $\mathcal{D} = \bigcup \mathcal{D}_i$ using the EM algorithm (Algorithm~\ref{algo:unigram_em}, Appendix~\ref{appendix:unig_tokenization}). Subtokenizers extracted for each $\mathcal{L}_i$ use the original $\mathcal{V}_i$, achieving compression rates comparable to independently trained monolingual tokenizers (Appendix~\ref{appendix:unigram_nsl}, Tab~\ref{tab:more_nsl_table_unigram}). As this approach achieves better compression and faster inference than frequency-based extraction, it is used as the default Unigram setting.

\subsection{BPE Tokenizer}
\label{sec:bpe_tokenizer}

BPE builds an \emph{ordered} sequence of merge rules by recursively combining frequent adjacent token pairs. This merge order dictates token priority during segmentation, tying the tokenizer to its merge table and making it non-trivial to combine independently trained monolingual BPE models. A common, albeit suboptimal, approach is to concatenate their merge tables by appending non-overlapping tokens from one vocabulary to the other. However, early merges from one language can prevent tokens specific to the other from ever being formed, resulting in inefficiencies 
(Fig.~\ref{fig:inefficient_bpe_merge}).

Inspired by \citet{purason-etal-2026-teaching}, we propose a \textbf{sequential training} strategy (Algorithm~\ref{alg:sequential_multilingual_bpe}). Given an ordered list of languages $\{\mathcal{L}_1, \dots, \mathcal{L}_N\}$ with corpora $\{\mathcal{D}_1, \dots, \mathcal{D}_N\}$ and a per-language vocabulary budget $k$, it incrementally builds a multilingual tokenizer $\mathcal{T}$ while producing a monolingual subtokenizer $\mathcal{T}_i$ for each language. A BPE tokenizer of size $k$ is first trained on $\mathcal{D}_1$, initializing the global tokenizer  $\mathcal{T} = \mathcal{T}_1$. For each subsequent language $\mathcal{L}_i$, the corpus $\mathcal{D}_i$ is tokenized with $\mathcal{T}$, and a modified BPE procedure simultaneously trains $\mathcal{T}_i$ and extends $\mathcal{T}$. At each iteration, we identify the most frequent unigram $u_m$ and bigram $b_m = (x, y)$ in $\mathcal{T}(\mathcal{D}_i)$, with respective counts $f_u$ and $f_b$. If $f_u \geq f_b$, we add $u_m$ to $\mathcal{T}_i$ along with all ancestor units and merges required to recompose it. Otherwise, since $x$ and $y$ are more frequent than $(x, y)$ in $\mathcal{D}_i$ and thus already belong to $\mathcal{T}_i$, we add the merged unit $(x, y)$ to both $\mathcal{T}_i$ and $\mathcal{T}$. As in standard BPE, occurrences of $(x, y)$ are replaced with $b_m$ in $\mathcal{T}(\mathcal{D}_i)$ and the frequencies are updated accordingly. This continues until $|\mathcal{V}_i| = k$. Crucially, the sequential merge order guarantees that every token in $\mathcal{T}_i$ can be formed (proof in \ref{appendix:Properness}).

Tabs.~\ref{tab:nsl_lowres} and ~\ref{tab:more_nsl_table_bpe} show that both $\mathcal{T}$ and each subtokenizer $\mathcal{T}_i$ achieve compression comparable to independently trained monolingual BPE models. Moreover, language ordering has negligible impact on compression (Tab.~\ref{tab:small_bpe_puplets}), so we use alphabetical order by default. For any subset of languages, a unified subtokenizer is formed by taking the union of their vocabularies and merges, preserving their original order in $\mathcal{T}$. Therefore, all tokens remain constructible (proof in~\ref{appendix:Properness}), unlike with naive concatenation, ensuring competitive compression across all selected languages.

\section{Modular Tokenizers Resolve Multilingual Tokenizers' Inefficiencies}

We now show that modular tokenizers resolve both inefficiencies documented in \textsection\ref{sec:inefficiency}. As these hit low-resource languages hardest, we retrain, for this section only, monolingual, sequential, and joint BPE tokenizers on 21 languages where Estonian, Finnish, Lithuanian, Latvian, Slovak, and Swedish are replaced by Croatian, Nepali, Sinhala, Somali, Swahili, and Telugu (three new scripts; using FineWeb-2~\citep{penedo2025fineweb2} data, same construction as before: 24k tokens per language, alphabetical order).

\subsection{Multilingual Fairness}
\label{sec:solving_fairness}

\begin{table}[t]
\footnotesize
\renewcommand{\arraystretch}{1.2}
\setlength{\tabcolsep}{3.5pt}
\centering
\begin{tabular}{l c cccccc}
\toprule
 & & \multicolumn{6}{c}{NSL ($\downarrow$)} \\
Model & Size & HR & NE & SI & SO & SW & TE \\
\midrule
TinyAya & 261k & 1.27 & 1.41 & 9.08 & 1.61 & 1.39 & 1.68 \\
NLLB    & 256k & 1.24 & 1.30 & 1.50 & 1.29 & 1.26 & 1.40 \\
Gemma3  & 262k & 1.34 & 1.45 & 1.79 & 1.54 & 1.57 & 1.61 \\
Llama3  & 128k & 1.62 & 2.43 & 6.68 & 1.74 & 1.84 & 7.40 \\
Qwen3   & 152k & 1.66 & 4.15 & 5.38 & 1.74 & 1.84 & 6.33 \\
\midrule
Multi 21 & 384k & 0.92 & 0.99 & 0.96 & 0.97 & 0.99 & 0.95 \\
Seq.\ 24k & 387k & 0.95 & 0.97 & 0.95 & 0.96 & 0.98 & 0.97 \\
Extracted 24k & 24k & 1.04 & 1.01 & 1.00 & 1.02 & 1.03 & 1.01 \\
\bottomrule
\end{tabular}
\caption{NSL relative to 24k-token monolingual BPE references on low-resource languages. \emph{Multi 21}: a standard BPE jointly trained on the same corpora; \emph{Seq.\ 24k}: our sequential multilingual BPE (24k tokens per language); \emph{Extracted 24k}: its per-language subtokenizers.}
\vspace{-0.5em}
\label{tab:nsl_lowres}
\end{table}

Our modular tokenizer yields subtokenizers on par with independently trained monolingual ones: every extracted 24k subtokenizer stays within 4\% of its monolingual counterpart, on the low-resource languages (Tab.~\ref{tab:nsl_lowres}) as on the conservative set (Tab.~\ref{tab:more_nsl_table_bpe}). The full sequential tokenizer is equally fair (NSL 0.95--0.99), matching standard BPE trained on the same corpora at a comparable size — modularity costs nothing in compression, yet every language can be served by its own subtokenizer. The contrast is sharpest on the low-resource languages: Open-weight tokenizers instead inflate sequences even on languages they report as supported — up to $9.1\times$ for TinyAya on Sinhala ~\citep{salamanca2026tinyayabridgingscale}. In absolute terms, ours cost a uniform 0.21--0.27 tokens per character across the six languages, against 0.31--0.48 for Gemma-3 and 0.29--2.44 for TinyAya (\ref{appendix:lowres_fairness}): no language pays a disproportionate price for being under-represented.

\subsection{Inference and Memory Efficiency}
\label{sec:solving_efficiency}

The fairness gains of \textsection\ref{sec:solving_fairness} translate into efficiency: inference time and memory follow the two quantities a tokenizer controls — the active vocabulary size, whose impact fades with model size, and the sequence length it produces, whose impact persists. The vocabulary costs compute at the output layer: running with the full vocabulary (an illustrative extreme — inference would restrict to the target languages) costs $1.58\times$ the monolingual reference at 0.5B but $1.07\times$ at 30B (Tab.~\ref{tab:exec_scale}). It also costs memory in its parameters: using a 24k sub-vocabulary instead of a conventional 256k removes 47\%, 35\%, and 10\% of \emph{all} parameters at 1B, 2.5B, and 30B for untied embeddings (31\%, 21\%, and 5\% when tied). Compression, in contrast, is a property of the tokenizer alone: each token costs a forward pass and a KV-cache entry, so time and cache shrink with the sequence. In time, Gemma-3 runs $2.5\times$ slower than the monolingual reference on Telugu at 0.5B and still $1.7\times$ at 30B, while our extracted subtokenizers stay within $0.99$--$1.04\times$ at every scale and language (full grid in Appendix~\ref{appendix:scale_efficiency}). In memory, Telugu sequences are $1.61\times$ the monolingual length under Gemma-3 vs.\ $1.01\times$ with our subtokenizer (Tab.~\ref{tab:nsl_lowres}), which thus needs 37\% less cache, a cut over Gemma-3 of 22\% (Croatian) to 44\% (Sinhala), and ${\sim}30\%$ even on well-covered Bulgarian.

\begin{table}[t]
\footnotesize
\renewcommand{\arraystretch}{1.2}
\setlength{\tabcolsep}{5pt}
\centering
\begin{tabular}{l c c c c c}
\toprule
 & & \multicolumn{4}{c}{Normalized inference time  ($\downarrow$)}\\
Tokenizer & Lang & 0.5B & 2.5B & 7B & 30B \\
\midrule

\multirow{2}{*}{Gemma3} 
& EN & 1.49	& 1.26 & 1.12 & 1.05 \\
& TE & 2.46 &	2.09 & 1.84	& 1.73 \\

\midrule
\multirow{2}{*}{Qwen3} 
& EN & 1.24 & 1.11 & 1.03 & 1.00 \\
& TE & 8.18 & 7.34 & 6.86 & 6.56 \\
\midrule
\multirow{2}{*}{Seq 24k}
& EN & 1.58 & 1.28 & 1.12 & 1.07 \\
& TE & 1.58 & 1.27 & 1.12 & 1.07 \\
\midrule
Extracted & (all  langs) & \multicolumn{4}{c}{0.99--1.04} \\
\bottomrule
\end{tabular}
\caption{Forward time on FLORES relative to a dedicated 24k monolingual tokenizer, across model scales. Full grid in App.~\ref{appendix:scale_efficiency} and architecture details in Tab.~\ref{tab:mem_params_app}.}
\label{tab:exec_scale}
\end{table}

\section{Training Multilingual Models with Modular Vocabularies}
\label{sec:multilingual_training}

\begin{figure*}[t]
    \centering
    \includegraphics[width=\linewidth, trim = {1em 3em 1em 15em}, clip]{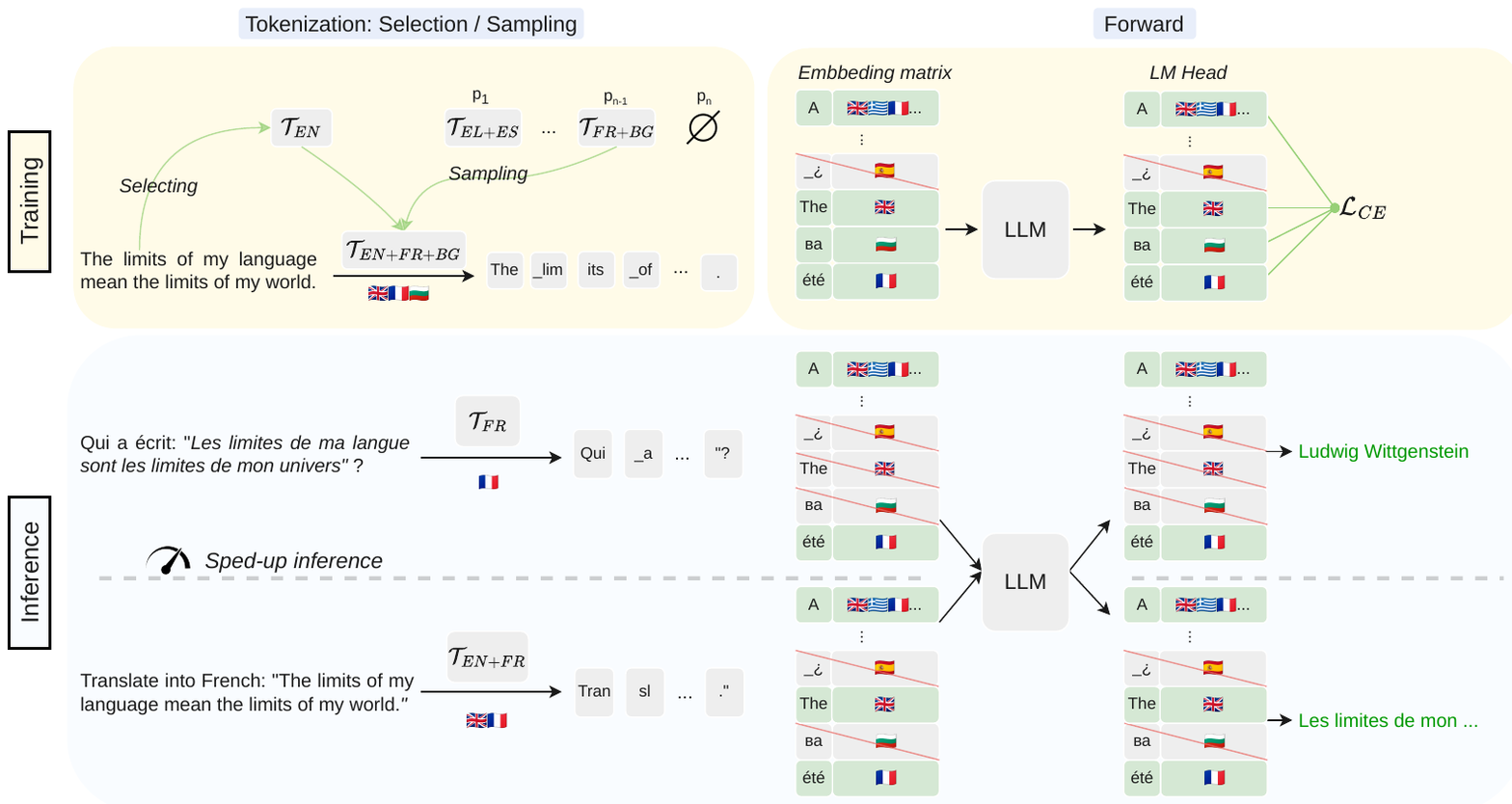}
    \caption{Overview of our multilingual training method and modular inference. \textbf{Top}: At training, monolingual batches are formed using either strictly monolingual or combined with $n$ other languages (here, $n=2$), restricting logit computation to that specific sub-vocabulary. \textbf{Bottom}: At inference, applying task-specific tokenizers (e.g., French-only for QA in French or English+French for translation tasks) shrinks the active vocabulary. This specialized approach reduces both the output token count and FLOPs, improving inference efficiency.}
    \label{fig:main_fig_multitraining}
\end{figure*}

\subsection{Experimental Settings}
\paragraph{Training Data.} For all 21 languages considered (Appendix~\ref{appendix:list_of_languages}), we use CommonCrawl data processed as follows: text is extracted from HTML using \texttt{resiliparse},\footnote{\url{https://resiliparse.chatnoir.eu}} language-identified with \texttt{fastText}\footnote{\url{https://fasttext.cc}} to build a dedicated corpus for each language, deduplicated at the paragraph level via hashing, and quality-filtered using a \texttt{fastText} classifier trained to distinguish high-quality documents (Wikipedia, textbooks, news, scientific articles) from random web pages. We use a mixture of 70\% English and 30\% other languages for training.

\paragraph{Models.} We use 2.5B non-embedding-parameter decoder-only Transformers \citep{Vaswani2017attention} with a hidden dimension of 3072 (hyperparameters in Appendix~\ref{appendix:model_hyperparameters}), trained on roughly 80B tokens with a context size of 4,096. Monolingual subtokenizers have a vocabulary budget of 24k tokens. As a baseline, standard training uses a single multilingual tokenizer with 384k tokens covering all languages. Merging monolingual Unigram tokenizers yields a vocabulary of 376,375 tokens, while sequential BPE yields 366,208 tokens. Since both approaches lead to the same conclusions (Tabs.~\ref{Tab: main_result_unigram} and~\ref{Tab: main_result_bpe}), the merged Unigram tokenizer is our default modular tokenizer unless stated otherwise.

\paragraph{Evaluation.} When using a subtokenizer, only the corresponding tokens are retained in the output matrix unless stated otherwise. For English, we evaluate on ARC Easy and Challenge \citep{clark2018thinksolvedquestionanswering}, HellaSwag \citep{zellers-etal-2019-hellaswag}, CommonSenseQA \citep{talmor-etal-2019-commonsenseqa}, BeleBele \citep{bandarkar-etal-2024-belebele}, PIQA \citep{bisk2020piqa}, SIQA \citep{sap-etal-2019-social}, and MMLU \citep{hendrycks2021measuring}, following the cloze form setting and normalization recommendations of \citet{gu-etal-2025-olmes}. For other languages, we use translated versions of ARC, HellaSwag, and MMLU from \citet{thelmann-etal-2024-towards}, as well as BeleBele \citep{bandarkar-etal-2024-belebele}. We refer to the average of these benchmarks as MCQ (Multi-Choice Questioning). For generation, we evaluate summarization on WikiLingua \citep{ladhak-etal-2020-wikilingua} in the monolingual setting, and translation on FLORES-200 \citep{costajussa-etal-2024-scaling} (English to other languages) in the multilingual setting. All evaluations are 5-shot, except HellaSwag and BeleBele which are 0-shot.

\subsection{Preliminary Analysis}
\label{sec:preliminary_analysis}

For simplicity, our experiments use the combination of all subtokenizers (referred to as \emph{full tokenizer} $\mathcal{T}$ ) as a reference to analyze model behavior when multiple subtokenizers are merged, even though in practice, only a subset would typically be used. Additional experiments with alternative subtokenizer combinations are presented in ~\ref{subsec:combos_subtokenizer}.

\begin{table}[t]

\footnotesize
\renewcommand{\arraystretch}{1.2}
\centering
\setlength{\tabcolsep}{2.8pt}
\begin{tabular}{l c ccccccc}
\toprule
\textbf{Model} & \textbf{Infer.} & \textbf{CS} & \textbf{EN} &  \textbf{FR} & \textbf{IT} & \textbf{NL} &  \textbf{PT} \\
\midrule

Baseline 
& Full & 11.9 & 19.2 & 15.3 & 13.4 & 18.0 & 17.9 \\
\midrule

\multirow{3}{*}{\shortstack[l]{\Nosampling}} 
& Lang & 12.2 & \textbf{21.8} &  14.9 & 13.6 & \textbf{19.5} & \textbf{19.4} \\
& Full & 4.2 & 20.4 &  7.2 & 3.8 & 4.6 & 5.1 \\
& Full - Lang & 11.9 & 20.9 &  14.1 & 12.9 & 18.1 & 18.5 \\

\midrule

\multirow{2}{*}{\Full} 
& Lang & 12.6 & 19.3 & 14.8 & 14.2 & 17.8 & 17.8 \\
& Full & 12.9 & 20.1 & 13.6 & \textbf{14.4} & 18.2 & 17.2 \\
\midrule

\multirow{2}{*}{\Data{2}}
& Lang & 12.3 & 20.1 & \textbf{16.1} & 13.2 & 18.9 & 18.5 \\
& Full & \textbf{13.0} & 20.2 & 15.9 & 12.7 & 18.8 & 17.5 \\

\bottomrule
\end{tabular}

\captionof{table}{Summarization across pretraining setups (Unigram). \emph{Infer.} is the inference tokenizer. \emph{Lang} is the monolingual tokenizer for the target language; \emph{Full} is the full tokenizer. \emph{Full-Lang}: inputs tokenized with \emph{Full}, predictions restricted to \emph{Lang} vocabulary.}
\vspace{-1em}
\label{Tab:sum_result_unigram}
\end{table}

\begin{table}[t]
\footnotesize
\setlength{\tabcolsep}{2.5pt}
\renewcommand{\arraystretch}{1.22}
\centering
\begin{tabular}{l ccccc}
\toprule
 & \textbf{BG} & \textbf{DA} & \textbf{EL} & \textbf{FR} & \textbf{IT} \\
\midrule

\textbf{\Nosampling} (EN+Lang) & 1.0 & 2.8 & 5.3 & 15.0 & 6.9 \\
\textbf{\Nosampling} (EN$|$Lang - Lang) & 1.6 & 1.5 & 12.3 & 26.0 & 12.7 \\
\textbf{Baseline} (Full) & \textbf{26.1} & \textbf{35.3} & \textbf{17.1} & \textbf{32.4} & \textbf{20.1} \\
\bottomrule
\end{tabular}
\captionof{table}{Translation results with the settings of Tab.~\ref{Tab:sum_result_unigram}. \emph{EN+Lang}: inference with the union of English and Lang subtokenizers. \emph{EN$|$Lang}: few-shot examples concatenate source tokenized with English subtokenizer and target with Lang subtokenizer; prediction restricted to Lang vocabulary.}

\label{Tab:translation_tok_results}

\end{table}

To study the effects of training with different tokenizers, we train a model referred to as \Nosampling\ on monolingual batches, where each language $\mathcal{L}_i$ is tokenized with $\mathcal{T}_i$ and the loss is computed only over $\mathcal{V}_i$ via the corresponding output layer activations. The baseline follows the same \emph{monolingual batching scheme} but with a single shared multilingual tokenizer. $\Nosampling$ matches the baseline on monolingual tasks including MCQ (Tab.~\ref{Tab: main_result_unigram}) and summarization (Tab.~\ref{Tab:sum_result_unigram}),  when each language is evaluated with its corresponding subtokenizer. However, combining subtokenizers at inference degrades performance, particularly for non-English generation, where outputs often mix tokens from different languages. This is expected: since token prediction is restricted to monolingual vocabularies during training, merging them at inference introduces a distribution shift in the prediction space.
To verify this, we evaluate the \emph{Full-Lang} setting on summarization (Tab.~\ref{Tab:sum_result_unigram}), where inputs are tokenized with $\mathcal{T}$ but generation is constrained to $\mathcal{V}_i$. Recovering baseline performance confirms that the distribution shift is the cause of degradation. English is less affected, as its dominance in training naturally biases \Nosampling\ toward English tokens.

\begin{table*}[t]
\centering
\footnotesize 
\setlength{\tabcolsep}{3pt} 
\renewcommand{\arraystretch}{1.3}

\begin{tabular}{l c l c c cc c cc c cc c cc c cc c cc c cc}
\toprule
\multirow{2}{*}{\textbf{Model}} & & \multirow{2}{*}{\textbf{Infer.}} & &
\textbf{EN} && 
\multicolumn{2}{c}{\textbf{BG}} && 
\multicolumn{2}{c}{\textbf{DA}} && 
\multicolumn{2}{c}{\textbf{EL}} && 
\multicolumn{2}{c}{\textbf{FR}} && 
\multicolumn{2}{c}{\textbf{IT}} && 
\multicolumn{2}{c}{\textbf{SK}} \\

\cmidrule(lr){5-5}
\cmidrule(lr){7-8}
\cmidrule(lr){10-11}
\cmidrule(lr){13-14}
\cmidrule(lr){16-17}
\cmidrule(lr){19-20}
\cmidrule(lr){22-23}

& & & &
MCQ && 
MCQ & MT && 
MCQ & MT &&
MCQ & MT && 
MCQ & MT && 
MCQ & MT && 
MCQ & MT \\

\midrule

Baseline 
& & Full & & 52.9 && 37.3 & 26.1 && 40.4 & \textbf{35.3} && 37.3 & 17.1 && 39.5 & 32.4 && 38.8 & 20.1 && 38.1 & 21.0 \\
\midrule

\multirow{2}{*}{\Nosampling}
& & Lang & & 52.6 && 38.3 & 1.0 && 40.5 & 2.8 && 37.9 & 5.3 && 40.3 & 15.0 && 38.9 & 6.9 && 38.9 & 1.6 \\
& & Full & & 49.6 && 36.3 & 0.9 && 38.3 & 2.4 && 37.1 & 2.5 && 37.8 & 7.4 && 37.2 & 3.0 && 36.1 & 1.4 \\
\midrule

\multirow{2}{*}{\Full}
& & Lang & & \textbf{53.2} && 38.1 & \textbf{27.9} && \textbf{41.6} & 35.1 && \textbf{38.0} & 16.9 && 39.3 & 32.5 && \textbf{39.1} & 19.8 && 38.3 & 20.3 \\
& & Full & & 53.0 && 38.0 & 27.7 && 41.5 & 35.2 && 37.9 & 16.8 && 39.4 & \textbf{32.6} && 38.8 & \textbf{20.2} && 38.2 & \textbf{21.1} \\
\midrule

\multirow{2}{*}{\Data{2}}
& & Lang & & 52.5 && \textbf{38.9} & 27.1 && 40.8 & 35.0 && 37.8 & \textbf{17.5} && \textbf{40.4} & 32.2 && 38.7 & 19.8 && \textbf{39.3} & 20.2 \\
& & Full & & 52.1 && 38.8 & 26.3 && 40.2 & 34.7 && 37.4 & 17.3 && 39.3 & 32.1 && 37.6 & 19.7 && 37.3 & 19.8 \\

\bottomrule
\end{tabular}
\caption{Multiple-Choice Questioning (MCQ) and Machine Translation (MT) performance under different pretraining sampling schemes for Unigram Tokenizers. \emph{Infer.} denotes the tokenizer used at inference; \emph{Full} corresponds to the full tokenizer; \emph{Lang} to the monolingual tokenizer for MCQ and the combined English–target tokenizer for MT.}
\label{Tab: main_result_unigram}
\end{table*}

This bias toward English is more pronounced when languages are mixed in context. We evaluate this in a 5-shot translation setting using the combination of the English and target language subtokenizers (\emph{EN+Lang} in Tab.~\ref{Tab:translation_tok_results}). Here, performance drops substantially compared to the baseline. The distribution shift in the prediction space partly explains this degradation, but does not fully account for it. To investigate further, we modify the evaluation protocol: source sentences are tokenized with the English subtokenizer and target translations with the target-language one, concatenated for each few-shot example, with generation restricted to the target vocabulary (\emph{EN $|$ Lang} in Tab.~\ref{Tab:translation_tok_results}). This mitigates the distribution shift and improves performance for some languages, but results remain far from the baseline. Despite these constraints, the model frequently produces English or mixed-language outputs, even for languages such as Bulgarian, which uses the Cyrillic script and shares few tokens with English. This suggests the issue goes beyond the prediction space distribution shift: during training, each sequence contains tokens from a single language only, so sequences mixing tokens from multiple languages at inference create out-of-distribution patterns that hinder proper language selection during translation.

\subsection{Training with Sampled Subtokenizers}

To mitigate this, we design $\Full$, a middle-ground training regime between using a single large multilingual tokenizer and monolingual subtokenizers. The former enables proper language selection but is costly; the latter is efficient but tends to produce language mixtures at inference. For a batch in language $\mathcal{L}_i$, we tokenize using either $\mathcal{T}$ or $\mathcal{T}_i$. Using $\mathcal{T}$ introduces a small fraction of tokens from other languages (0.5--5\%, depending on the language; Appendix~\ref{appendix:inserted_tokens}). We hypothesize that exposing the model to such mixed-token sequences acts as regularization, improving its ability to generate coherent text in $\mathcal{L}_i$ even when foreign tokens appear in context. As a result, the model assigns more probability mass to target-language tokens during translation, enabling proper language switching and recovering baseline performance. 

As the number of languages grows, the \emph{full tokenizer} becomes very large (Appendix~\ref{appendix:scaling_languages}), making training expensive and requiring approximate softmax methods \citep{joulin2017efficient, baevski2018adaptive}. To improve scalability, we propose \Data{n} (Fig.~\ref{fig:main_fig_multitraining}): text in language $\mathcal{L}_i$ is tokenized either with $\mathcal{T}_i$ alone or combined with $n$ additional subtokenizers (instead of $\mathcal{T}$ as in $\Full$ method). We decided to sample the additional subtokenizers proportionally to the frequencies of their corresponding languages in the training data. This sampling strategy ensures two key properties that we found to be important to match the baseline performance. First, it calibrates token probabilities across languages: tokens from overrepresented languages tend to have larger logits, so including their subtokenizers more often as "negatives" mitigates this bias. Second,  when tokenizing texts from different languages with $\mathcal{T}$, English contributes the largest share of foreign tokens due to its prevalence in training data (Appendix~\ref{appendix:inserted_tokens}). So subtokenizers introducing the most out-of-language tokens should be selected more frequently to reproduce the regularizing effect of the $\Full$ method.

To simplify training, we retain the \emph{monolingual batching scheme}: each batch uses a sampled subtokenizer (either $\mathcal{T}_i$ or a combination), ensuring all tokens belong to the same vocabulary subset and allowing the loss to be computed over that subset only. This avoids the training overhead caused by near-linear vocabulary growth, without requiring approximate softmax methods. Unless stated otherwise, we sample the monolingual tokenizer 50\% of the time for $\Full$ and \Data{n}. These strategies recover baseline performance across MCQ, translation (Tab.~\ref{Tab: main_result_unigram}), and summarization (Tab.~\ref{Tab:sum_result_unigram}). Crucially, \Data{n} maintains strong performance when using the full tokenizer at inference despite never seeing it during training, demonstrating that the model generalizes to any combination of subtokenizers without sacrificing performance.

\begin{table}[t]
\setlength{\tabcolsep}{3pt}
\centering
\begin{tabular}{l c c c c}
\toprule
$n$ lang. & 1 & 2 & 3 & 4\\
\midrule
Seq. & $1.01_{\pm .01}$ & $1.00_{\pm .01}$ & $0.99_{\pm .01}$ & $0.99_{\pm .01}$ \\
Concat. & $1.00_{\pm .00}$ & $1.22_{\pm .27}$ & $1.30_{\pm .27}$ & $1.33_{\pm .25}$ \\
\bottomrule
\end{tabular}

\caption{Mean NSL $\pm$ std across languages for $n$-language subtokenizers: concatenated monolingual tokenizers vs.\ extraction from sequential BPEs trained under different language orders (details in~\ref{appendix:p_uplet_violin_sequential}).}

\label{tab:small_bpe_puplets}
\end{table}

\subsection{Memory Footprint at Many Languages}
\label{sec:memory_scaling}

Scaling to more languages grows the vocabulary near-linearly (Appendix~\ref{appendix:scaling_languages}), and with it the memory-resident vocabulary state: embeddings, output matrix, optimizer states. CPU offloading lifts this bottleneck by keeping this state in CPU RAM \citep{miao_2021,ren2021zerooffloaddemocratizingbillionscalemodel}, each step touching only the sampled subtokenizers' rows. Updates are pipelined following the delayed scheme of \citet{ren2021zerooffloaddemocratizingbillionscalemodel}: while the GPU computes step $t$, the CPU updates the rows active at step $t{-}1$ with their gradients; the transfers (gradients of $t{-}1$ to the CPU, rows for step $t{+}1$ to the GPU) run asynchronously from pinned buffers and hide behind compute as well. Every per-step cost therefore depends only on the languages sampled, not on the total vocabulary: we measure identical step times whether CPU RAM holds 21, 100, or 250 languages. To introduce no overhead, this off-GPU work must complete within one GPU step, which we verify on a naive prototype (Appendix~\ref{appendix:offloading}), separate from our training runs: at batch 128, on \Data{5} batches, the full off-GPU work  takes 1.2\,s vs.\ a 1.5\,s GPU step at 1B, 2.6\,s vs.\ 3.1\,s at 2.5B, and 4.7\,s vs.\ 7.4\,s at 7B, and monolingual batches stay far below. These timings can be further reduced using the fused optimizer of \citet{ren2021zerooffloaddemocratizingbillionscalemodel}, which significantly speeds up the CPU update that dominates the off-GPU work.

\section{Ablations}\label{sec:ablations}

\begin{table}[t]
\centering
\footnotesize 
\setlength{\tabcolsep}{5pt} 
\renewcommand{\arraystretch}{1.2}
\begin{tabular}{l c cc c cc}
\toprule
\textbf{Model} & 
\textbf{BG} & \textbf{DA} & \textbf{EL} &  \textbf{FR} & \textbf{IT} &  \textbf{SK} \\
\midrule
Baseline
  & 26.1 & 35.3 & 17.1 & 32.4 & 20.1 & 21.0 \\
\midrule
\Data{1}
  & 21.4 & 26.4 & 14.7 & 27.4 & 16.7 & 13.6 \\
\midrule
\Data{2}
  & 26.3 & 34.7 & \textbf{17.3} & 32.1 & 19.7 & 19.8 \\
\midrule
\Data{5}
 & \textbf{28.0} & \textbf{36.1} & 17.2 & 32.4 & \textbf{20.7} & \textbf{21.2} \\
\midrule
\Full
 &27.7 & 35.2 & 16.8 & \textbf{32.6} & 20.2 & 21.1\\

\bottomrule
\end{tabular}
\caption{Translation performance when varing n in \Data{n} using the Full tokenizer (Unigram).}
\label{Tab:n_abalation_result_unigram}
\end{table}

\paragraph{Sequential BPE is order-robust and preserves compression.}
\label{sec: bpe_order_invariant}
Tab.~\ref{tab:small_bpe_puplets} compares subtokenizers covering $n$ languages built via (i) sequential BPE training followed by subtokenizer extraction and merging as in \textsection~\ref{sec:bpe_tokenizer}, evaluated across multiple training orders, and (ii) concatenation of independently trained monolingual tokenizers. For $n=1$, subtokenizers extracted from sequential BPE achieve compression comparable to independently trained monolingual tokenizers. For larger $n$, concatenation becomes sensitive to both language selection and order, leading to unstable and degraded compression. In contrast, sequential BPE is insensitive to both training and combination orders: fixed merge priorities avoid ordering ambiguity, preserving strong and consistent compression across all languages in the considered set.

\paragraph{Sampling more subtokenizers improves performance.}

\Data{n} interpolates between \Nosampling\ $\sim$\Data{0} and \Full\ $\sim$\Data{n-1} for $n$ languages. Increasing $n$ improves the model's ability to handle mixed-language contexts, as in translation (Tab.~\ref{Tab:n_abalation_result_unigram}). Even small $n$ achieves near-baseline performance, while larger $n$ is beneficial when scaling to more languages. The main constraint is memory, as larger $n$ increases the output matrix size. Since no performance degradation was observed, $n$ can be as large as memory permits.
\section{Conclusions}

In this work, we highlight a key limitation of multilingual LLMs: their reliance on a single large shared tokenizer causes uneven compression across languages and memory and compute inefficiencies, especially when models are deployed on a subset of the supported languages. To address this, we introduce modular multilingual tokenizers for both BPE and Unigram, decomposable into fair and tailorable subtokenizers. We further propose a pretraining approach that samples subtokenizers per training batch, computing logits only over the relevant vocabulary subset, curbing the cost of large vocabularies. This enables flexible subtokenizer selection at inference. Our results show that this yields more efficient, fair, and modular multilingual LLMs — better cross-lingual balance, lower memory usage, and faster inference than shared-tokenizer models — at competitive performance.

\section*{Limitations}
While our approach improves compression fairness and inference efficiency, some limitations remain. First, our low-resource evaluation is tokenizer-level: compression and fairness are measured on six genuinely low-resource languages, but reliable downstream validation requires stronger pretraining for these languages — our preliminary runs were too noisy to be conclusive — and remains future work. Second, subtokenizer selection at inference requires knowing the target language(s) in advance; language identification or a broader subtokenizer union can cover uncertain inputs, but we do not evaluate such pipelines.

\section*{Acknowledgments}
François Yvon has been partly funded by the French National Funding Agency (ANR) under the France 2030 program (ref.\ ANR-23-IACL-0007).

\bibliography{custom}

\appendix

\clearpage

\section{Tokenizer Study}
\subsection{Tokenizer construction}

\subsubsection{Multilingual Unigram Tokenizer}
\label{appendix:unig_tokenization}
\paragraph{Forward–Backward Re-estimation over Merged Vocabularies.}
To establish a unified probability space for the merged vocabulary $V = \bigcup V_i$, parameter estimation is performed over the aggregate multilingual corpus $\mathcal{D} = \bigcup \mathcal{D}_i$. As detailed in Algorithm \ref{algo:unigram_em}, the Expectation-Maximization (EM) procedure reestimates token probabilities $p(v)$ using the forward-backward algorithm \citep{forwardbackward}.\\

For a sentence $S \in \mathcal{D}$ of length $n$, the forward and backward variables, $\alpha(i)$ and $\beta(i)$, are computed to represent the marginal probabilities of all segmentations for the prefix $S[0:i]$ and suffix $S[i:n]$, respectively. 

Under the Unigram independence assumption, the total sentence likelihood is given by $P(S) = \alpha(n) = \beta(0)$. The E-step utilizes these variables to calculate the expected counts $E[c(v)]$ for each token $v \in V$ by marginalizing over all valid segmentation paths in the corpus. In the subsequent M-step, the parameters are updated via the normalization:
\[
p(v) = \frac{E[c(v)]}{\sum_{v' \in V} E[c(v')]}
\]
This iterative process ensures that the probabilities for the merged vocabulary are normalized and remain consistent with the multilingual data distribution.

\subsubsection{Multilingual BPE Tokenizer}

\label{appendix:bpe_tokenizer}

Each BPE tokenizer $\mathcal{T}$ consists of an ordered list of merges $\mathcal{T}.\text{merges}$ and of tokens $\mathcal{T}.\text{vocab}$.

\paragraph{Proof of Properness.}
\label{appendix:Properness}
To prove the correctness of Algorithm~\ref{alg:sequential_multilingual_bpe}, we show that it preserves the \emph{properness} of the merge sequence \cite{berglund-vandermerwe-2023-formalizing}. This property ensures that any compound token is introduced only after all tokens required to construct it have already been included.

Formally, let $\mu = [\mu_0, \mu_1, \dots, \mu_N]$ be the sequence of merges, where each $\mu_i = (x_i, y_i) \rightarrow z_i$. The sequence $\mu$ is \emph{proper} if for all $i$:
\begin{itemize}
    \item $x_i \in \Sigma$ or $\exists j < i$ such that $x_i = z_j$,
    \item $y_i \in \Sigma$ or $\exists j < i$ such that $y_i = z_j$.
\end{itemize}
That is, both constituents of any merge must either belong to the base alphabet $\Sigma$ or have been produced by a previous merge. Note that any prefix of a proper merge sequence is also proper. \\

\emph{Global tokenizer $\mathcal{T}$:}
The global tokenizer $\mathcal{T}$ is initialized using standard BPE on $\mathcal{D}_1$, which is known to produce a proper merge sequence. At each subsequent step, Algorithm~\ref{alg:sequential_multilingual_bpe} only appends a merge $(x,y)$ to $\mathcal{T}$ when both $x$ and $y$ already belong to $\mathcal{T}$. Since $(x,y)$ is selected from $\mathcal{T}(\mathcal{D}_i)$, both $x$ and $y$ are already representable using $\mathcal{T}$. Therefore, any newly added merge satisfies the properness condition, and $\mathcal{T}$ remains proper by closure under extension. \\

\emph{Language-specific tokenizers $\mathcal{T}_i$:}
Each $\mathcal{T}_i$ is constructed by adding tokens together with all their ancestors in $\mathcal{T}$, in the order induced by $\mathcal{T}$. Since $\mathcal{T}$ is proper, every merge depends only on tokens that appear earlier in this order. Thus, when a merge $(x,y) \rightarrow z$ is included in $\mathcal{M}_i$, both $x$ and $y$ have already been added to $\mathcal{V}_i$. Moreover, the merge sequence of $\mathcal{T}_i$ is an order-preserving subsequence of that of $\mathcal{T}$ that is closed under ancestry up to the vocabulary budget $k$. Although some tokens may be omitted due to truncation, no merge is included without its dependencies. Therefore, the merge sequence of $\mathcal{T}_i$ is proper.\\

\emph{Conclusion.}
Both the global tokenizer $\mathcal{T}$ and each $\mathcal{T}_i$ maintain proper merge sequences, ensuring that all constructed tokenizers are valid BPE tokenizers.

\paragraph{Proof of Properness of Combined Subtokenizers.}
\label{appendix:Properness_combination}

For any subset of languages $\{\mathcal{L}_1,\dots,\mathcal{L}_k\}$, let $\mathcal{T}_{\text{comb}}$ be the tokenizer obtained by taking the union of their vocabularies and merges and reordering them according to the global tokenizer $\mathcal{T}$. Since each language-specific tokenizer $\mathcal{T}_i$ is proper, every token it contains is associated with a valid ancestral merge chain in the global sequence $\mathcal{T}$. As $\mathcal{T}_{\text{comb}}$ is formed by taking the union of such tokens and preserving their global ordering, it remains closed under these ancestral dependencies. Therefore, every token in $\mathcal{T}_{\text{comb}}$ is constructible from $\Sigma$, and $\mathcal{T}_{\text{comb}}$ is proper.

In contrast, arbitrary concatenation of independently trained tokenizers may introduce tokens whose merge histories are incompatible or incomplete with respect to $\mathcal{T}$, leading to tokens that are not jointly constructible within a shared merge graph.

\begin{figure*}[t]
  \begin{minipage}{\textwidth}

\begin{algorithm}[H]
\caption{Unigram Probabilities Estimation via Expectation-Maximization}
\label{algo:unigram_em}
\begin{algorithmic}[1]
\Require Multilingual Corpus $\mathcal{D}$, Merged Vocabulary $V$
\Statex
\State \textbf{Initialization:} Set $p(v) = \frac{1}{|V|}$ for all $v \in V$
\Statex
\Repeat
    \State $E[c(v)] \gets 0, \quad \forall v \in V$ \Comment{Reset expected counts}
    \Statex
    
    \ForAll{sentences $S \in \mathcal{D}$}
        \State $n \gets |S|$
        \State $\alpha(0) \gets 1, \quad \beta(n) \gets 1$
        \Statex
        
        \For{$j = 1$ \textbf{to} $n$} \Comment{ E-step: Forward Pass}
            \State $\alpha(j) \gets \sum_{i: 0 \le i < j, S[i:j] \in V} \alpha(i) \cdot p(S[i:j])$
        \EndFor
        \Statex
        
        \For{$i = n-1$ \textbf{down to} $0$} \Comment{E-step: Backward Pass}
            \State $\beta(i) \gets \sum_{j: i < j \le n, S[i:j] \in V} p(S[i:j]) \cdot \beta(j)$
        \EndFor
        \Statex
        
        \For{$i = 0$ \textbf{to} $n-1$} \Comment{E-step: Expectation Accumulation}
            \For{$j = i+1$ \textbf{to} $n$}
                \If{$S[i:j] \in V$}
                    \State $v \gets S[i:j]$
                    \State $E[c(v)] \gets E[c(v)] + \frac{\alpha(i) \cdot p(v) \cdot \beta(j)}{\alpha(n)}$
                \EndIf
            \EndFor
        \EndFor
    \Statex
    \EndFor
    \\
    
    \State $Z \gets \sum_{v' \in V} E[c(v')]$ \Comment{ M-step: Probabilities Update}
    \ForAll{$v \in V$}
        \State $p(v) \gets \frac{E[c(v)]}{Z}$
    \EndFor
\Statex
\Until{Log-likelihood $\sum_{S \in \mathcal{D}} \log \alpha(n)$ converges}
\end{algorithmic}
\end{algorithm}
  \end{minipage}
\end{figure*}
\clearpage

\begin{figure*}[!t]
  \begin{minipage}{\textwidth}
\begin{algorithm}[H]
\caption{Sequential Multilingual BPE with Language-Specific Extraction}
\label{alg:sequential_multilingual_bpe}
\begin{algorithmic}[1]

\Statex \textbf{Inputs:} Ordered languages $\{\mathcal{L}_1, \dots, \mathcal{L}_N\}$; corresponding corpora $\{\mathcal{D}_1, \dots, \mathcal{D}_N\}$; vocabulary budget $k$ per language
\Statex \textbf{Outputs:} Global multilingual tokenizer $\mathcal{T}$; language-specific tokenizers $\{\mathcal{T}_i\}_{i=1}^N$, each preserving the merge order of $\mathcal{T}$ and containing at most $k$ tokens
\Statex
\State $\mathcal{T}_1 \gets$ Train BPE with $|\mathcal{V}_1| = k$ on $\mathcal{D}_1$
\State $\mathcal{T} \gets \mathcal{T}_1$
\Statex
\For{$i = 2$ to $N$}

    \State $\mathcal{D}_i^{tok} \gets \mathcal{T}(\mathcal{D}_i)$
    \State Compute unigram and bigram frequencies in $\mathcal{D}_i^{tok}$

    \State Initialize $\mathcal{V}_i \gets \emptyset$, \quad $\mathcal{M}_i \gets \emptyset$

    \Statex

    \While{$|\mathcal{V}_i| < k$}

        \State $u_m \gets$ most frequent unigram with frequency $f_u$
        \State $b_m = (x,y) \gets$ most frequent bigram with frequency $f_b$

        \If{$f_u \geq f_b$}
            \State $t_{\text{new}} \gets u_m$
        \Else
            \State $t_{\text{new}} \gets \textsc{Merge}(x,y)$
            \State Replace all occurrences of $(x,y)$ with $t_{\text{new}}$ in $\mathcal{D}_i^{tok}$
            \State Update unigram and bigram frequencies
            \State Append $(x,y)$ to $\mathcal{T}.\text{merges}$
            \State Append $t_{\text{new}}$ to $\mathcal{T}.\text{vocab}$
        \EndIf
        \Statex

        \State $(\mathcal{A}, \mathcal{P}) \gets$ ancestors of $t_{\text{new}}$ in $\mathcal{T}$ and corresponding merges (ordered as in $\mathcal{T}$)

        \For{$(t, merge) \in (\mathcal{A}, \mathcal{P})$}
            \If{$|\mathcal{V}_i| < k$}
                \State $\mathcal{V}_i \gets \mathcal{V}_i \cup \{t\}$
                \If{$merge \neq \text{None}$} \Comment{None for original bytes/characters}
                    \State $\mathcal{M}_i \gets \mathcal{M}_i \cup \{merge\}$
                \EndIf
            \Else
                \State \textbf{break}
            \EndIf
        \EndFor
    \Statex

    \EndWhile

    \State Order $\mathcal{V}_i$ and $\mathcal{M}_i$ according to $\mathcal{T}$
    \State Define $\mathcal{T}_i$ from $(\mathcal{V}_i, \mathcal{M}_i)$

\EndFor

\State \Return $\mathcal{T}$, $\{\mathcal{T}_i\}_{i=1}^N$

\end{algorithmic}
\end{algorithm}
\end{minipage}
\end{figure*}

\clearpage

\twocolumn[{%
\noindent\begin{minipage}{\textwidth}
\centering
\scriptsize
\setlength{\tabcolsep}{3pt}
\renewcommand{\arraystretch}{1.2}

\begin{tabular}{l ccccccccccccccccccccc}
\toprule
 \textbf{Model}
& \textbf{BG} & \textbf{CS} & \textbf{DA} & \textbf{DE} & \textbf{EL} & \textbf{EN}
& \textbf{ES} & \textbf{FR} & \textbf{HR} & \textbf{HU} & \textbf{IT}
& \textbf{NE} & \textbf{NL} & \textbf{PL} & \textbf{PT} & \textbf{RO}
& \textbf{SI} & \textbf{SL} & \textbf{SO} & \textbf{SW} & \textbf{TE} \\
\midrule
TinyAya & 0.32 & 0.29 & 0.27 & 0.23 & 0.31 & \textbf{0.21} & 0.23 & 0.24 & 0.30 & 0.31 & 0.23 & 0.33 & 0.25 & 0.29 & 0.24 & 0.28 & 2.44 & 0.30 & 0.37 & 0.29 & 0.39 \\
\midrule
NLLB & 0.30 & 0.32 & 0.26 & 0.27 & 0.33 & 0.25 & 0.25 & 0.28 & 0.29 & 0.29 & 0.26 & 0.30 & 0.26 & 0.31 & 0.26 & 0.29 & 0.40 & 0.29 & 0.29 & 0.26 & 0.32 \\
\midrule
Gemma3 & 0.33 & 0.33 & 0.29 & 0.24 & 0.38 & \textbf{0.21} & 0.23 & 0.25 & 0.31 & 0.33 & 0.25 & 0.34 & 0.26 & 0.30 & 0.25 & 0.29 & 0.48 & 0.33 & 0.35 & 0.33 & 0.37 \\
\midrule
Llama3 & 0.41 & 0.33 & 0.33 & 0.28 & 0.39 & \textbf{0.21} & 0.27 & 0.28 & 0.38 & 0.42 & 0.29 & 0.56 & 0.30 & 0.37 & 0.29 & 0.35 & 1.79 & 0.38 & 0.40 & 0.39 & 1.70 \\
\midrule
Qwen3 & 0.44 & 0.45 & 0.33 & 0.28 & 0.86 & \textbf{0.21} & 0.27 & 0.27 & 0.39 & 0.42 & 0.29 & 0.96 & 0.30 & 0.35 & 0.28 & 0.35 & 1.45 & 0.39 & 0.39 & 0.39 & 1.46 \\
\midrule
BPE mono 24k & \textbf{0.23} & 0.24 & 0.23 & \textbf{0.21} & \textbf{0.22} & 0.22 & 0.22 & 0.23 & 0.23 & 0.23 & \textbf{0.21} & \textbf{0.23} & \textbf{0.21} & \textbf{0.22} & 0.22 & 0.23 & 0.27 & 0.23 & 0.23 & \textbf{0.21} & 0.23 \\
\midrule
Extracted 24k & \textbf{0.23} & 0.24 & 0.23 & 0.22 & \textbf{0.22} & 0.22 & 0.22 & 0.23 & 0.24 & 0.24 & 0.22 & \textbf{0.23} & 0.22 & 0.23 & 0.22 & 0.23 & 0.27 & 0.24 & 0.23 & 0.22 & 0.23 \\
\midrule
Seq. 24k & \textbf{0.23} & \textbf{0.23} & \textbf{0.22} & \textbf{0.21} & \textbf{0.22} & \textbf{0.21} & \textbf{0.21} & \textbf{0.22} & \textbf{0.22} & \textbf{0.22} & \textbf{0.21} & \textbf{0.23} & \textbf{0.21} & \textbf{0.22} & \textbf{0.21} & \textbf{0.22} & \textbf{0.26} & \textbf{0.22} & \textbf{0.22} & \textbf{0.21} & \textbf{0.22} \\
\bottomrule
\end{tabular}
\captionof{table}{\textbf{Tokens per character} ($\downarrow$) on FLORES-200; lower and more uniform is fairer.}
\label{tab:token_per_char}

\label{tab:token_per_char_lowres}
\vspace{1em}

\end{minipage}
}]

\subsection{Language choice}
\label{appendix:list_of_languages}
The selected dataset comprises twenty-one official working languages of the European Union, strategically chosen to provide a rigorous and representative linguistic testbed. We exclude Croatian, Irish and Maltese because of the poor coverage of these languages in our training data. A primary motivation for utilizing this specific European cohort is its coverage of three distinct scripts: Latin, Cyrillic, and Greek. By evaluating our method across these diverse writing systems, we can effectively demonstrate the framework's modularity and robust handling of script-level variation. The chosen languages are named using the ISO 639-1 language codes:\footnote{\url{https://en.wikipedia.org/wiki/List_of_ISO_639_language_codes}}
\begin{itemize}
\item \textbf{Latin Script (19 languages)}: Czech (cs), Danish (da), German (de), English (en), Spanish (es), Estonian (et), Finnish (fi), French (fr), Hungarian (hu), Italian (it), Lithuanian (lt), Latvian (lv), Dutch (nl), Polish (pl), Portuguese (pt), Romanian (ro), Slovak (sk), Slovenian (sl), Swedish (sv);
\item \textbf{Cyrillic Script (1 language)}: Bulgarian (bg);
\item \textbf{Greek Script (1 language)}: Greek (el).
\end{itemize}

\subsection{Tokenizer Compression Efficiency}
\label{appendix:compression_efficiency}

In these experiments, we evaluate the compression of tokenizers using tokens per character (Tab.~\ref{tab:token_per_char}). We then compare our modular multilingual tokenizers—Sequential BPE (Tab.~\ref{tab:more_nsl_table_bpe}) and merged monolingual Unigram (Tab.~\ref{tab:more_nsl_table_unigram})—showing compression comparable to both monolingual baselines and a standard multilingual tokenizer of similar size. Sequential BPE is trained with an alphabetical ordering of languages. We also report compression for frequency-based subtokenizers (Tab.~\ref{tab:more_nsl_table_freq}).

\subsubsection{Tokens per Character ($\downarrow$)}
\label{appendix:lowres_fairness}
Tab.~\ref{tab:token_per_char_lowres} reports tokens per character across all 21 languages. Our tokenizers consistently achieve the lowest and most uniform values (0.21--0.27), while open-weight tokenizers are higher and more variable, especially on the low-resource languages.

\subsubsection{BPE Tokenizer}
\label{appendix:bpe_tokenizer}

Tab.~\ref{tab:more_nsl_table_bpe} reports NSL scores for BPE tokenizers across all 21 languages. Open-weight tokenizers show consistently high NSL. For similar vocabulary budgets, sequential BPE matches standard multilingual BPE (Multi 21), showing that modular construction does not hurt compression, while enabling extraction of monolingual subtokenizers that match independently trained monolingual tokenizers at a fraction of the global vocabulary size.

\begin{table*}

\footnotesize
\setlength{\tabcolsep}{8.5pt}
\renewcommand{\arraystretch}{1.2}
\centering

\begin{tabular}{llccccccccccc} 
\toprule
\rowcolor{blue!8}
\multicolumn{2}{c}{} & \multicolumn{11}{c}{NSL ($\downarrow$)} \\ 
\rowcolor{blue!8}
Model & Size & BG & CS & DA & DE & EL & EN & ES & ET & FI & FR & HU \\

\midrule
TinyAya & 261k & 1.38 & 1.22 & 1.18 & 1.11 & 1.42 & 0.98 & 1.09 & 1.34 & 1.35 & 1.06 & 1.39 \\
NLLB    & 256k & 1.33 & 1.32 & 1.16 & 1.28 & 1.51 & 1.14 & 1.19 & 1.27 & 1.30 & 1.21 & 1.31 \\
Gemma3  & 262k & 1.42 & 1.38 & 1.29 & 1.13 & 1.73 & 0.98 & 1.07 & 1.48 & 1.54 & 1.09 & 1.49 \\
Llama3  & 128k & 1.78 & 1.36 & 1.45 & 1.35 & 1.78 & 0.98 & 1.27 & 1.75 & 1.83 & 1.24 & 1.87 \\
Qwen3   & 152k & 1.92 & 1.86 & 1.44 & 1.32 & 3.90 & 0.96 & 1.25 & 1.74 & 1.82 & 1.22 & 1.87 \\
\midrule
\multirow{4}{*}{Multi 21} 
& 128k & 1.16 & 1.07 & 1.07 & 1.10 & 1.17 & 1.06 & 1.08 & 1.11 & 1.14 & 1.09 & 1.12 \\
& 256k & 1.04 & 0.96 & 0.99 & 1.00 & 1.04 & 0.98 & 1.00 & 1.00 & 1.02 & 1.01 & 1.01 \\
& 384k & 0.99 & 0.91 & 0.95 & 0.95 & 0.98 & 0.96 & 0.97 & 0.94 & 0.95 & 0.97 & 0.96 \\
& 512k & 0.96 & 0.88 & 0.93 & 0.93 & 0.94 & 0.94 & 0.95 & 0.91 & 0.91 & 0.95 & 0.92 \\
\midrule
Seq. 8k  & 116k & 1.18 & 1.13 & 1.10 & 1.13 & 1.19 & 1.10 & 1.12 & 1.13 & 1.16 & 1.10 & 1.16 \\
Seq. 16k & 241k & 1.05 & 1.01 & 1.00 & 1.02 & 1.05 & 1.01 & 1.02 & 1.02 & 1.03 & 1.01 & 1.04 \\
Seq. 24k & 366k & 1.00 & 0.95 & 0.96 & 0.97 & 0.98 & 0.97 & 0.97 & 0.96 & 0.97 & 0.97 & 0.98 \\
\midrule
Extracted 24k & 24k & 1.00 & 1.00 & 1.01 & 1.02 & 0.99 & 1.01 & 1.01 & 1.02 & 1.05 & 1.01 & 1.04 \\
\midrule
\rowcolor{blue!8}
Model & Size  & IT & LT & LV & NL & PL & PT & RO & SK & SL & SV &\\
 \midrule
TinyAya & 261k & 1.09 & 1.40 & 1.45 & 1.19 & 1.32 & 1.10 & 1.25 & 1.28 & 1.30 & 1.21 \\
NLLB    & 256k & 1.22 & 1.30 & 1.27 & 1.23 & 1.40 & 1.21 & 1.29 & 1.23 & 1.23 & 1.22 \\
Gemma3  & 262k & 1.16 & 1.59 & 1.72 & 1.24 & 1.33 & 1.12 & 1.30 & 1.43 & 1.42 & 1.29 \\
Llama3  & 128k & 1.38 & 2.03 & 2.10 & 1.43 & 1.67 & 1.31 & 1.53 & 1.62 & 1.62 & 1.46 \\
Qwen3   & 152k & 1.36 & 1.94 & 2.08 & 1.41 & 1.55 & 1.28 & 1.52 & 1.82 & 1.68 & 1.45 \\
\midrule
\multirow{4}{*}{Multi 21} 

 & 128k & 1.09 & 1.11 & 1.11 & 1.09 & 1.11 & 1.08 & 1.08 & 1.06 & 1.09 & 1.07 \\
 & 256k & 1.00 & 1.00 & 1.01 & 1.00 & 1.00 & 0.99 & 1.00 & 0.96 & 0.99 & 0.99 \\
 & 384k & 0.97 & 0.94 & 0.95 & 0.96 & 0.94 & 0.96 & 0.95 & 0.91 & 0.94 & 0.94 \\
 & 512k & 0.95 & 0.90 & 0.92 & 0.94 & 0.90 & 0.94 & 0.93 & 0.88 & 0.91 & 0.92 \\

\midrule

Seq. 8k  & 116k & 1.12 & 1.16 & 1.15 & 1.10 & 1.15 & 1.11 & 1.11 & 1.10 & 1.11 & 1.09 \\
Seq. 16k & 241k & 1.02 & 1.03 & 1.03 & 1.01 & 1.03 & 1.00 & 1.01 & 0.99 & 1.01 & 0.99 \\
Seq. 24k & 366k & 0.97 & 0.97 & 0.97 & 0.97 & 0.96 & 0.96 & 0.96 & 0.94 & 0.96 & 0.95 \\
\midrule
Extracted 24k & 24k & 1.02 & 1.04 & 1.03 & 1.03 & 1.03 & 1.01 & 1.02 & 1.04 & 1.04 & 1.03 \\
\bottomrule
\end{tabular}

\caption{\textbf{BPE NSL} per language, relative to a 24k-token monolingual BPE tokenizer. \emph{Multi 21}: standard BPE trained jointly on all 21 languages. \emph{Seq.\ k}: the global sequential tokenizer with a budget of $k$ tokens per language (Algorithm~\ref{alg:sequential_multilingual_bpe}). \emph{Extracted 24k}: per-language subtokenizers extracted from \emph{Seq.\ 24k} (\textsection~\ref{sec:bpe_tokenizer}).}

\label{tab:more_nsl_table_bpe}
\end{table*}

\subsubsection{Unigram Tokenizer}
\label{appendix:unigram_nsl}
Tab.~\ref{tab:more_nsl_table_unigram} reports NSL scores for Unigram tokenizers. Results are consistent with BPE: for similar vocabulary budgets, merged monolingual Unigram tokenizers achieve compression comparable to standard multilingual Unigram (Multi 21), while the extracted monolingual subtokenizers match independently trained monolingual tokenizers across all languages.

\begin{table*}

\footnotesize
\setlength{\tabcolsep}{8.5pt}
\renewcommand{\arraystretch}{1.2}
\centering

\begin{tabular}{llccccccccccc} 
\toprule
\rowcolor{blue!8}
\multicolumn{2}{c}{} & \multicolumn{11}{c}{NSL ($\downarrow$)} \\ 
\rowcolor{blue!8}
Model & Size & BG & CS & DA & DE & EL & EN & ES & ET & FI & FR & HU \\
\midrule
TinyAya & 261k & 1.38 & 1.22 & 1.18 & 1.11 & 1.42 & 0.98 & 1.09 & 1.34 & 1.35 & 1.06 & 1.39 \\
NLLB    & 256k & 1.33 & 1.32 & 1.16 & 1.28 & 1.51 & 1.14 & 1.19 & 1.27 & 1.30 & 1.21 & 1.31 \\
Gemma3  & 262k & 1.42 & 1.38 & 1.29 & 1.13 & 1.73 & 0.98 & 1.07 & 1.48 & 1.54 & 1.09 & 1.49 \\
Llama3  & 128k & 1.78 & 1.36 & 1.45 & 1.35 & 1.78 & 0.98 & 1.27 & 1.75 & 1.83 & 1.24 & 1.87 \\
Qwen3   & 152k & 1.92 & 1.86 & 1.44 & 1.32 & 3.90 & 0.96 & 1.25 & 1.74 & 1.82 & 1.22 & 1.87 \\
\midrule
\multirow{4}{*}{Multi 21} 
& 128k & 1.18 & 1.04 & 1.09 & 1.07 & 1.15 & 1.05 & 1.06 & 1.20 & 1.13 & 1.07 & 1.09 \\
& 256k & 1.10 & 0.97 & 1.02 & 1.00 & 1.07 & 1.00 & 1.00 & 1.09 & 1.02 & 1.01 & 1.01 \\
& 384k & 1.08 & 0.96 & 1.01 & 0.98 & 1.06 & 1.00 & 1.00 & 1.06 & 1.00 & 1.00 & 1.00 \\
& 512k & 1.08 & 0.96 & 1.01 & 0.98 & 1.06 & 1.00 & 1.00 & 1.06 & 1.00 & 1.00 & 1.00 \\

\midrule
Merged 8k  & 120k & 1.18 & 1.12 & 1.09 & 1.12 & 1.20 & 1.10 & 1.10 & 1.12 & 1.13 & 1.10 & 1.13 \\
Merged 16k & 249k & 1.06 & 1.00 & 1.00 & 1.01 & 1.06 & 1.01 & 1.01 & 1.01 & 1.01 & 1.01 & 1.01 \\
Merged 24k & 376k & 1.00 & 0.94 & 0.96 & 0.97 & 0.99 & 0.98 & 0.97 & 0.96 & 0.95 & 0.98 & 0.97 \\

\midrule
Extracted 24k & 24k & 1.00 & 1.00 & 1.00 & 1.00 & 1.00 & 1.00 & 1.00 & 1.00 & 1.00 & 1.00 & 1.00 \\
\bottomrule
\end{tabular}
\end{table*}

\begin{table*}

\footnotesize
\setlength{\tabcolsep}{8.5pt}
\renewcommand{\arraystretch}{1.2}
\centering

\begin{tabular}{llccccccccccc} 
\toprule
\rowcolor{blue!8}
\multicolumn{2}{c}{} & \multicolumn{11}{c}{NSL ($\downarrow$)} \\ 
\rowcolor{blue!8}
Model & Size  & IT & LT & LV & NL & PL & PT & RO & SK & SL & SV &\\
 \midrule
TinyAya & 261k & 1.09 & 1.40 & 1.45 & 1.19 & 1.32 & 1.10 & 1.25 & 1.28 & 1.30 & 1.21 \\
NLLB    & 256k & 1.22 & 1.30 & 1.27 & 1.23 & 1.40 & 1.21 & 1.29 & 1.23 & 1.23 & 1.22 \\
Gemma3  & 262k & 1.16 & 1.59 & 1.72 & 1.24 & 1.33 & 1.12 & 1.30 & 1.43 & 1.42 & 1.29 \\
Llama3  & 128k & 1.38 & 2.03 & 2.10 & 1.43 & 1.67 & 1.31 & 1.53 & 1.62 & 1.62 & 1.46 \\
Qwen3   & 152k & 1.36 & 1.94 & 2.08 & 1.41 & 1.55 & 1.28 & 1.52 & 1.82 & 1.68 & 1.45 \\
\midrule
\multirow{4}{*}{Multi 21} 

& 128k & 1.07 & 1.17 & 1.26 & 1.08 & 1.06 & 1.06 & 1.09 & 1.08 & 1.14 & 1.08 \\
& 256k & 1.00 & 1.07 & 1.14 & 1.01 & 0.98 & 1.00 & 1.02 & 1.01 & 1.06 & 1.00 \\
& 384k & 1.00 & 1.06 & 1.12 & 1.00 & 0.97 & 0.99 & 1.01 & 1.00 & 1.04 & 0.99 \\
& 512k & 0.99 & 1.06 & 1.11 & 1.00 & 0.96 & 0.99 & 1.01 & 1.00 & 1.04 & 0.99 \\

\midrule

Merged 8k  & 120k & 1.09 & 1.14 & 1.14 & 1.10 & 1.15 & 1.10 & 1.11 & 1.11 & 1.11 & 1.10 \\
Merged 16k & 249k & 1.00 & 1.01 & 1.02 & 1.01 & 1.02 & 1.00 & 1.01 & 1.00 & 1.00 & 1.00 \\
Merged 24k & 376k & 0.97 & 0.95 & 0.96 & 0.97 & 0.96 & 0.97 & 0.96 & 0.94 & 0.95 & 0.96 \\

\midrule
Extracted 24k & 24k & 1.00 & 1.00 & 1.00 & 1.00 & 1.00 & 1.00 & 1.00 & 1.00 & 1.00 & 1.00 \\
\bottomrule
\end{tabular}

\caption{\textbf{Unigram NSL} per language, relative to a 24k-token monolingual Unigram tokenizer. \emph{Multi 21}: standard Unigram trained jointly on all 21 languages. \emph{Merged k}: concatenation of the 21 monolingual tokenizers of vocabulary size $k$, with scores re-estimated as in Algorithm~\ref{algo:unigram_em}. \emph{Extracted 24k}: per-language subtokenizers extracted from \emph{Merged 24k} (\textsection~\ref{sec:unigram}); all values round to 1.00 (true range 0.9999--1.0027).}

\label{tab:more_nsl_table_unigram}
\end{table*}

\subsubsection{Frequency-based Subtokenizers}
\label{appendix:frequency_based}

Tab.~\ref{tab:more_nsl_table_freq} reports NSL scores for frequency-based subtokenizers extracted from multilingual Unigram tokenizers of two sizes (256k and 384k). As $k$ increases, compression approaches the full tokenizer, with $k=24$k already achieving comparable compression. However, compression is inherently bounded by the base tokenizer — languages poorly compressed by it remain so regardless of $k$. This limitation motivates our focus on the merged monolingual approach, which matches independently trained monolingual tokenizers.

\begin{table*}
\footnotesize
\setlength{\tabcolsep}{8.5pt}
\renewcommand{\arraystretch}{1.2}
\centering

\begin{tabular}{lcccccccccccc} 
\toprule
\rowcolor{blue!8}
\multicolumn{2}{c}{} & \multicolumn{11}{c}{NSL ($\downarrow$)} \\ 
\rowcolor{blue!8}
Base Size & k & BG & CS & DA & DE & EL & EN & ES & ET & FI & FR & HU \\
\midrule
\multirow{5}{*}{256k} 

& 8k & 1.19 & 1.26 & 1.19 & 1.21 & 1.21 & 1.17 & 1.18 & 1.21 & 1.23 & 1.17 & 1.20\\
& 16k & 1.10 & 1.08 & 1.07 & 1.07 & 1.09 & 1.05 & 1.06 & 1.12 & 1.09 & 1.06 & 1.07\\
& 24k & 1.10 & 1.02 & 1.04 & 1.03 & 1.08 & 1.02 & 1.03 & 1.10 & 1.05 & 1.03 & 1.03\\
& 32k & 1.10 & 1.00 & 1.03 & 1.01 & 1.07 & 1.01 & 1.01 & 1.09 & 1.03 & 1.02 & 1.02\\
& Full & 1.10 & 0.97 & 1.02 & 1.00 & 1.07 & 1.00 & 1.00 & 1.09 & 1.02 & 1.01 & 1.01\\
\midrule
\multirow{5}{*}{384k} 
& 8k & 1.19 & 1.28 & 1.19 & 1.22 & 1.22 & 1.18 & 1.19 & 1.22 & 1.24 & 1.17 & 1.21\\
& 16k & 1.10 & 1.08 & 1.08 & 1.07 & 1.09 & 1.06 & 1.06 & 1.12 & 1.09 & 1.06 & 1.07\\
& 24k & 1.09 & 1.02 & 1.04 & 1.03 & 1.07 & 1.02 & 1.03 & 1.09 & 1.05 & 1.03 & 1.03\\
& 32k & 1.08 & 1.00 & 1.03 & 1.01 & 1.06 & 1.01 & 1.01 & 1.08 & 1.03 & 1.02 & 1.02\\
& Full & 1.08 & 0.96 & 1.01 & 0.98 & 1.06 & 1.00 & 1.00 & 1.06 & 1.00 & 1.00 & 1.00\\
\midrule
\rowcolor{blue!8}
Base Size & k  & IT & LT & LV & NL & PL & PT & RO & SK & SL & SV &\\
\midrule
\multirow{5}{*}{256k} 
& 8k & 1.20 & 1.21 & 1.22 & 1.19 & 1.26 & 1.20 & 1.19 & 1.23 & 1.19 & 1.20\\
& 16k & 1.07 & 1.11 & 1.16 & 1.07 & 1.08 & 1.06 & 1.08 & 1.09 & 1.10 & 1.07\\
& 24k & 1.03 & 1.09 & 1.14 & 1.03 & 1.02 & 1.03 & 1.05 & 1.04 & 1.07 & 1.03\\
& 32k & 1.02 & 1.08 & 1.14 & 1.02 & 1.00 & 1.01 & 1.03 & 1.03 & 1.06 & 1.02\\
& Full & 1.00 & 1.07 & 1.14 & 1.01 & 0.98 & 1.00 & 1.02 & 1.01 & 1.06 & 1.00\\
\midrule
\multirow{5}{*}{384k} 
& 8k & 1.21 & 1.22 & 1.23 & 1.20 & 1.29 & 1.21 & 1.20 & 1.24 & 1.20 & 1.21\\
& 16k & 1.07 & 1.11 & 1.15 & 1.07 & 1.09 & 1.07 & 1.08 & 1.09 & 1.10 & 1.07\\
& 24k & 1.03 & 1.08 & 1.13 & 1.03 & 1.02 & 1.03 & 1.04 & 1.04 & 1.07 & 1.03\\
& 32k & 1.02 & 1.07 & 1.12 & 1.02 & 1.00 & 1.01 & 1.03 & 1.02 & 1.06 & 1.01\\
& Full & 1.00 & 1.06 & 1.12 & 1.00 & 0.97 & 0.99 & 1.01 & 1.00 & 1.04 & 0.99\\

\bottomrule
\end{tabular}

\caption{\textbf{Frequency-based subtokenizers NSL score.} NSL per language, computed relative to a 24k tokens monolingual Unigram tokenizer. From multilingual Unigram tokenizers of various sizes, we extract the $k$ most frequent tokens, as described in \textsection~\ref{sec:unigram}, to build a language-specific tokenizer. We then compute the NSL scores for both the multilingual tokenizer and the language-specific one. We observe that when $k = 24\text{k}$, the language-specific subtokenizer achieves a compression performance similar to that of the original multilingual tokenizer (\emph{Full}).}
\label{tab:more_nsl_table_freq}

\end{table*}

\subsection{Sequential BPE Robustness}
\subsubsection{Robustness to Language Order in Training}

\begin{figure*}[!h]

    \centering
    \includegraphics[width=0.8\linewidth, trim=0 0 0 2.5em, clip]
    {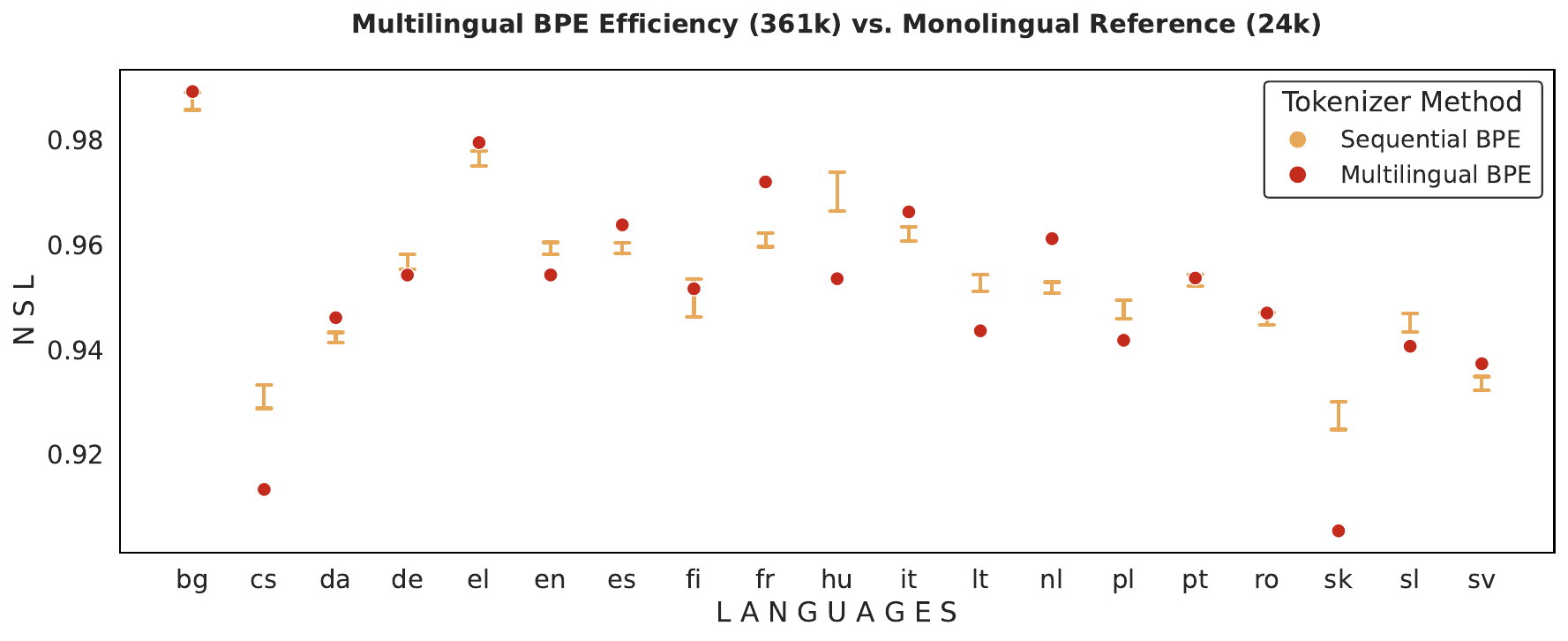}

    \caption{Per-language NLS comparison between a standard multilingual BPE tokenizer and a sequential one across language orders. In the sequential setting, each language is assigned a 24k token budget, yielding an average final vocabulary size of ~361k tokens across orders, which matches the size of the standard BPE used for comparison. Horizontal bars show the min–max range across orders.}
    \label{fig:placeholder}
\end{figure*}

To study the sensitivity of the resulting tokenizer to the order of languages in sequential BPE training, we train the tokenizer using multiple language orderings. Fig.~\ref{fig:placeholder} shows that the NLS of the resulting tokenizer varies only slightly across different training orders for all considered languages. This supports our choice of using alphabetical ordering for training in all experiments, as it provides a simple and consistent strategy without significantly impacting results.

\subsubsection{Robustness against Naive Concatenation in Subtokenizer Construction}
\label{appendix:p_uplet_violin_sequential}

\begin{figure*}[!h]
    \centering
    \vspace{-0.3em}
    \begin{subfigure}[b]{0.48\textwidth}
    \includegraphics[width=\linewidth]{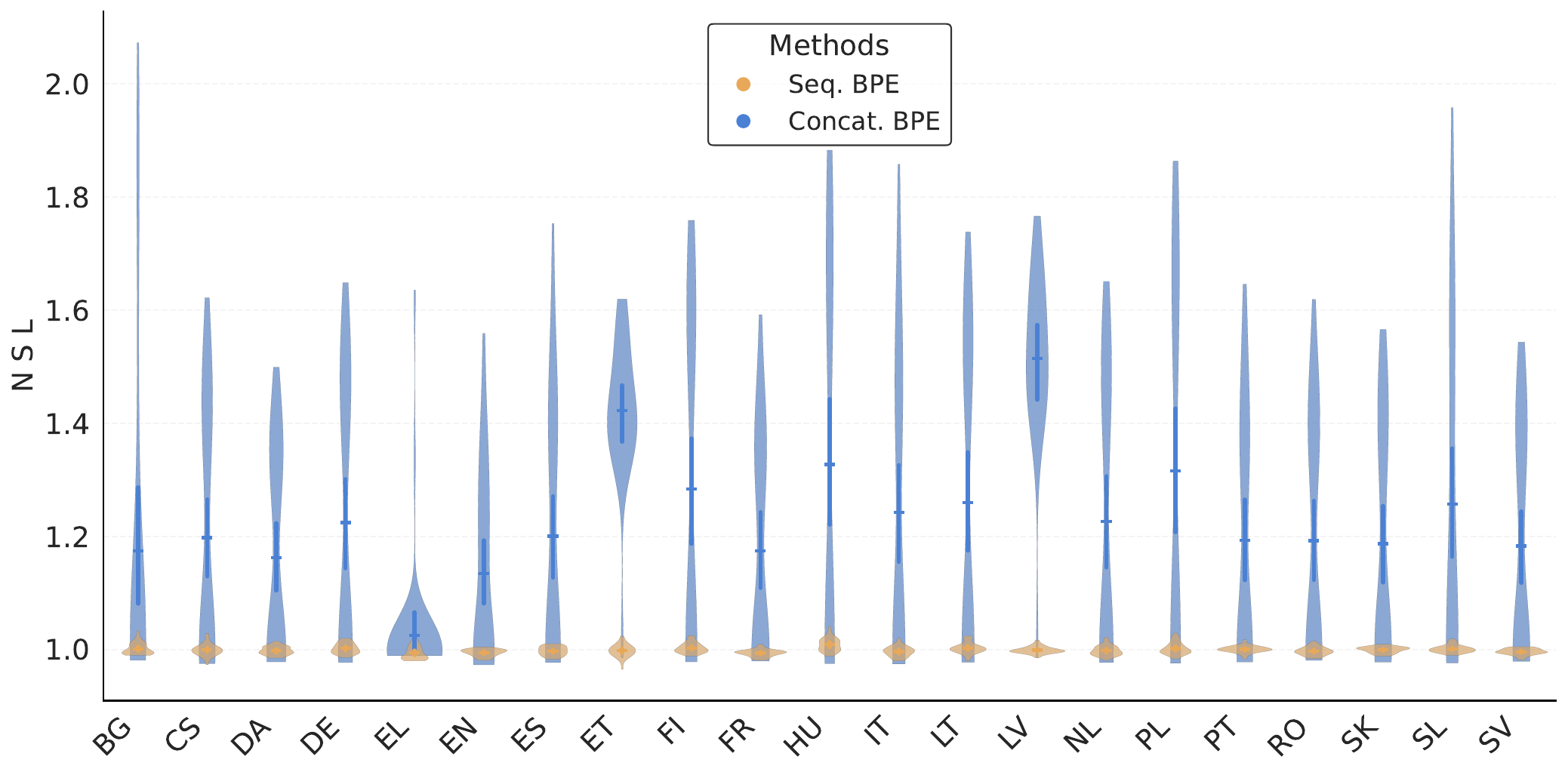}
    \caption{$2$-uplet of languages}
    \label{fig:left_image}
     \end{subfigure}
    \begin{subfigure}[b]{0.48\textwidth}
    \includegraphics[width=\linewidth]{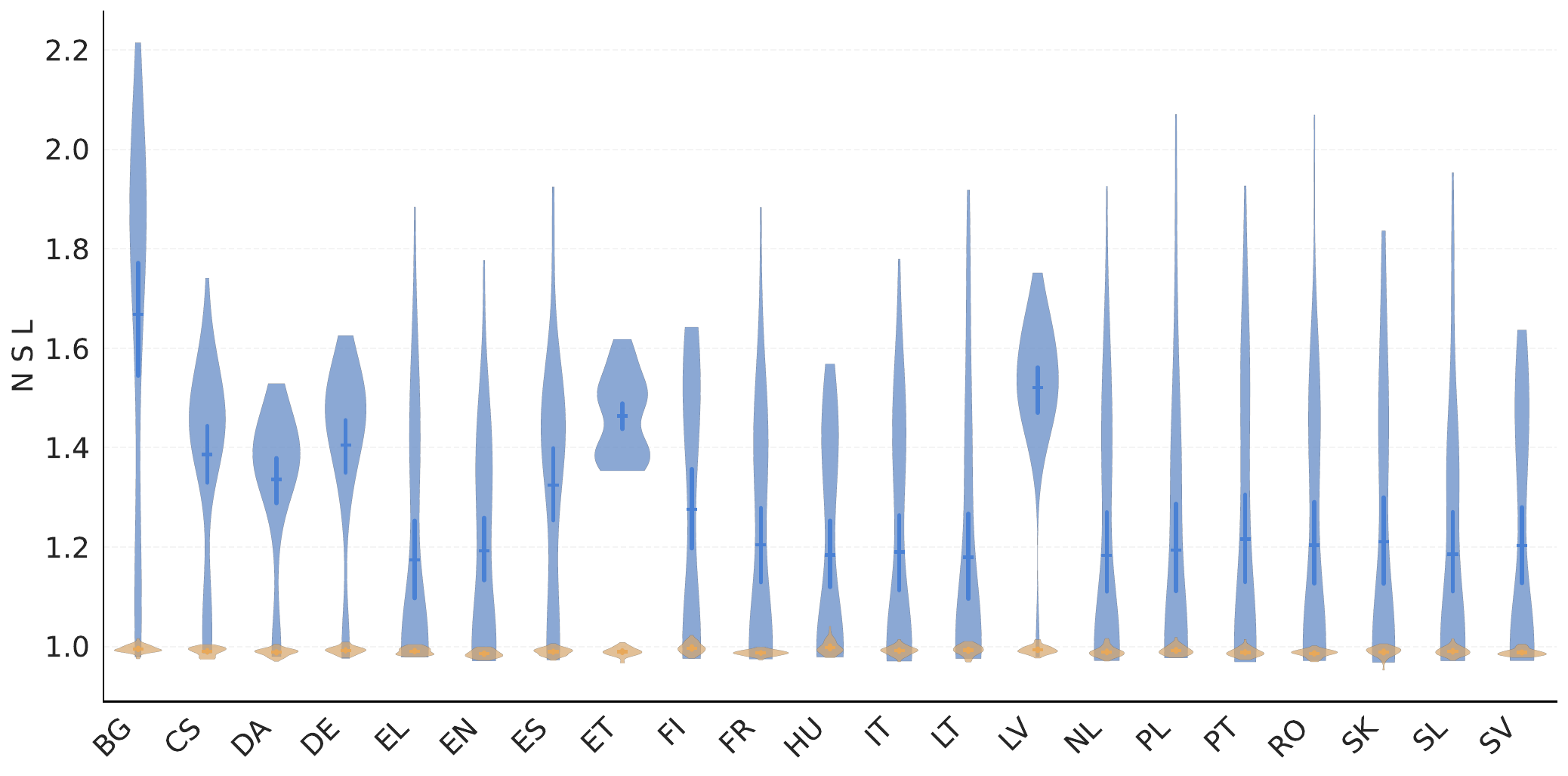}
        \caption{$3$-uplet of languages}
    \label{fig:mid_image}
     \end{subfigure}
     \begin{subfigure}[b]{0.48\textwidth}
        \includegraphics[width=\linewidth]{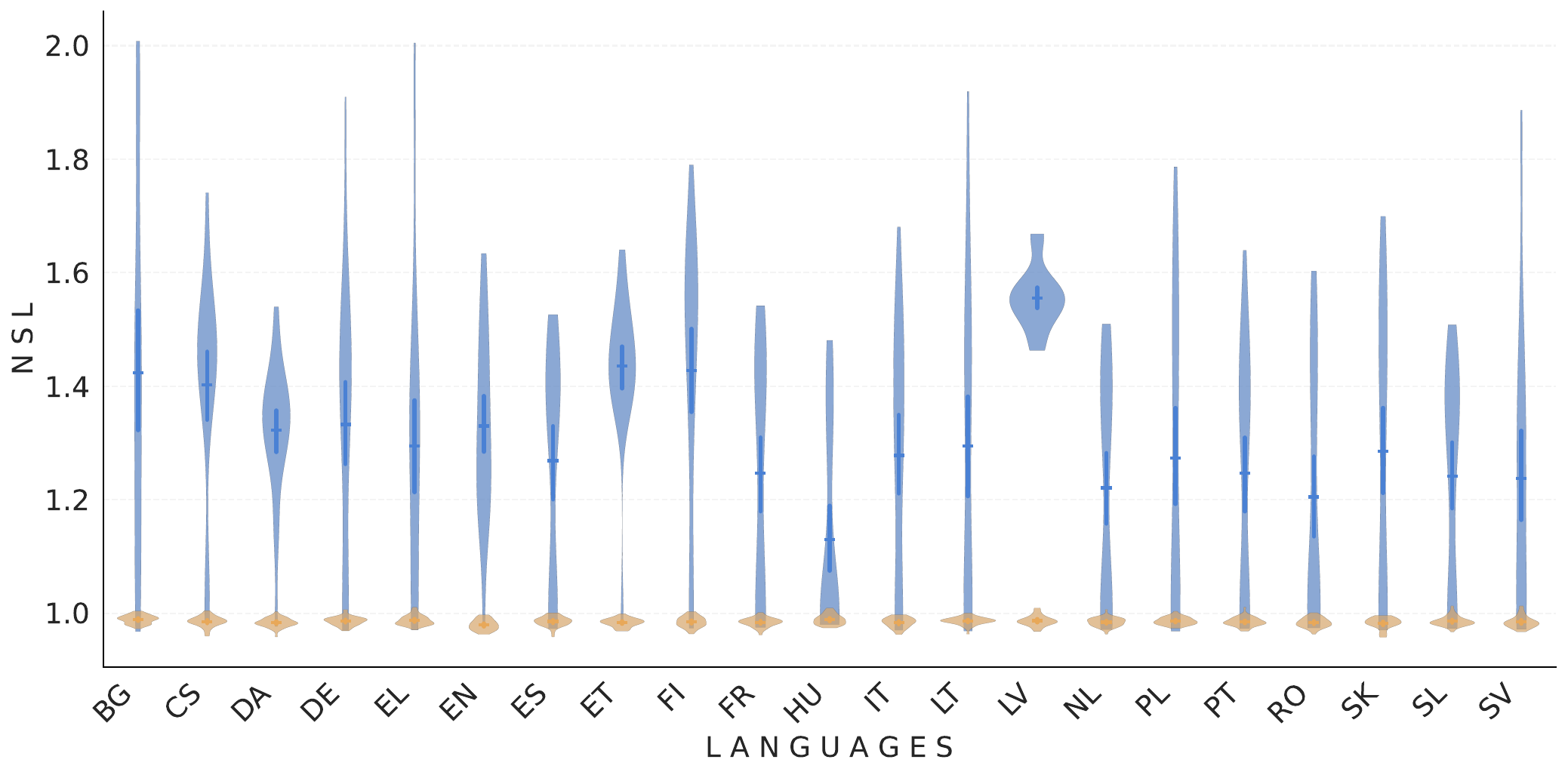}
         \caption{$4$-uplet of languages}
    \label{fig:right_image}
     \end{subfigure}
     
    \vspace{-0.3em}
    \caption{Comparison per language of NLS for combinations of BPE tokenizers over $p$ languages under different language orders, either via trivial concatenation (blue) or by combining the corresponding subtokenizers extracted from a sequential BPE (orange). The density (horizontal thickness) indicates the number of tokenizer combinations falling within each NLS range. It has been truncated where the density approaches zero for visual clarity.}

    \label{fig:p_uplet_violin_sequential}
\end{figure*}

The experiment in Fig.~\ref{fig:p_uplet_violin_sequential} extends \textsection~\ref{sec: bpe_order_invariant}. We compare two approaches for constructing subtokenizers for a subset of $n$ languages: (i) combining, as described in \textsection~\ref{sec:bpe_tokenizer}, the corresponding subtokenizers extracted from sequential BPE trained under different language orders, and (ii) a concatenation baseline that appends non-overlapping tokens and merges from independently trained monolingual tokenizers, where the resulting compression ratio varies significantly with the order of concatenation. Results in Fig.~\ref{fig:p_uplet_violin_sequential} show that sequential training yields more robust subtokenizer composition than naive concatenation.

\begin{figure*}[!h]
    \centering
    \vspace{-0.3em}
    \includegraphics[width=0.5\linewidth]{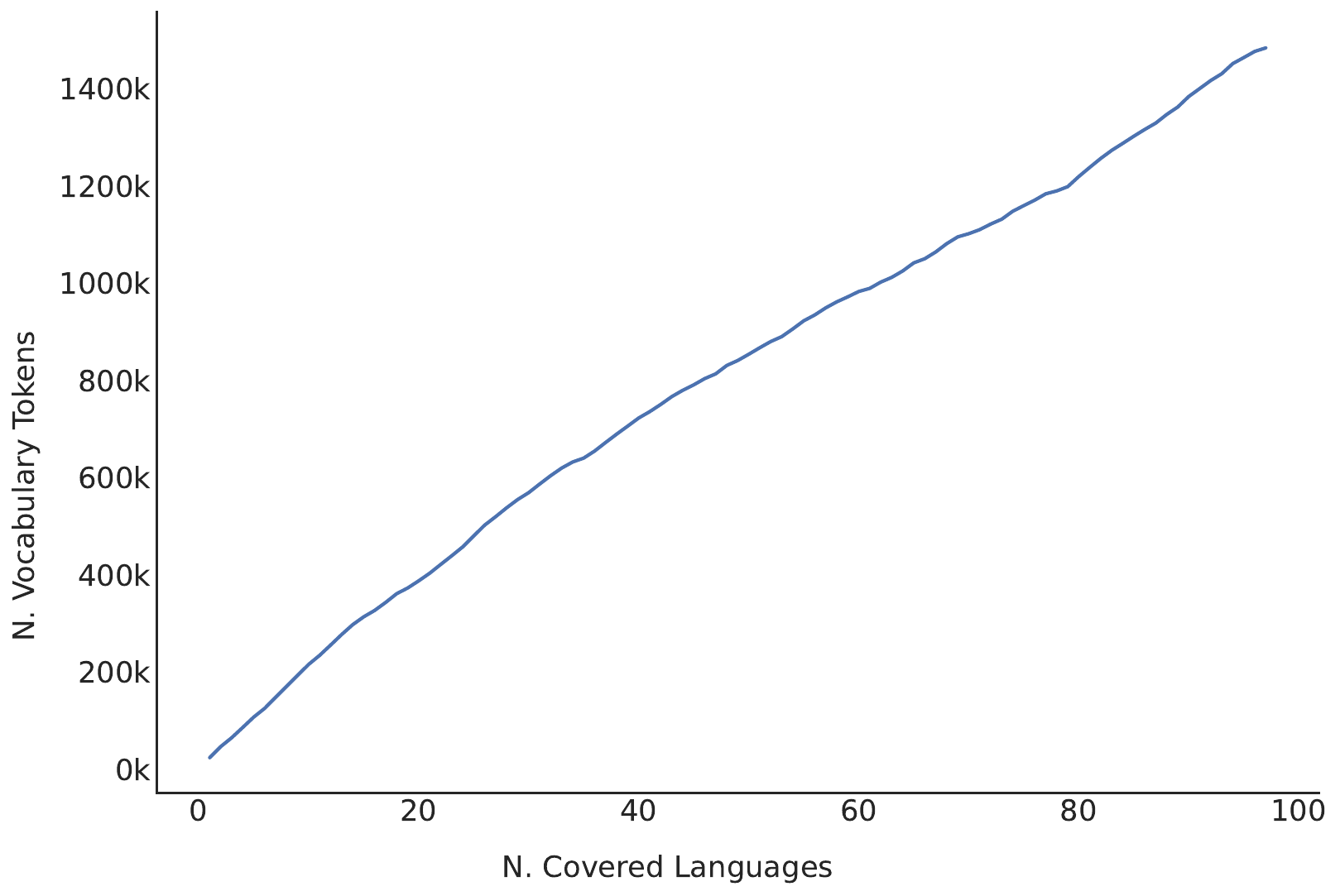}
    \vspace{-0.8em}
    \caption{\textbf{Vocabulary size scaling} of a sequential multilingual BPE (24k tokens/language): despite shared tokens, the vocabulary grows near-linearly with the number of languages.}

    \label{fig:scaling_more_languages_BPE}
\end{figure*}

\subsection{Scaling of Vocabulary Size in Sequential Multilingual BPE with the Number of Languages}
\label{appendix:scaling_languages}
While ensuring fairness across languages, this sequential approach scales poorly as the language set expands. As illustrated in Fig.~\ref{fig:scaling_more_languages_BPE}, the vocabulary size increases linearly with the number of languages. This trend underscores the need for training methods capable of integrating subtokenizers to maintain equitable compression ratios across a diverse linguistic range while staying computationally feasible. Our selection encompasses a highly diverse range of linguistic families, including major groups such as Indo-European, Afro-Asiatic, and Niger-Congo, as well as Dravidian, Turkic, Austronesian, Sino-Tibetan, and Tai-Kadai systems. This ensures our evaluation covers a broad spectrum of morphological structures.

\subsection{Inference Time Comparison for Unigram Tokenizers}
\label{appendix:inference_time_comparison_unigram}

\begin{figure*}[!t]
    \centering
    \includegraphics[width=0.7\linewidth]{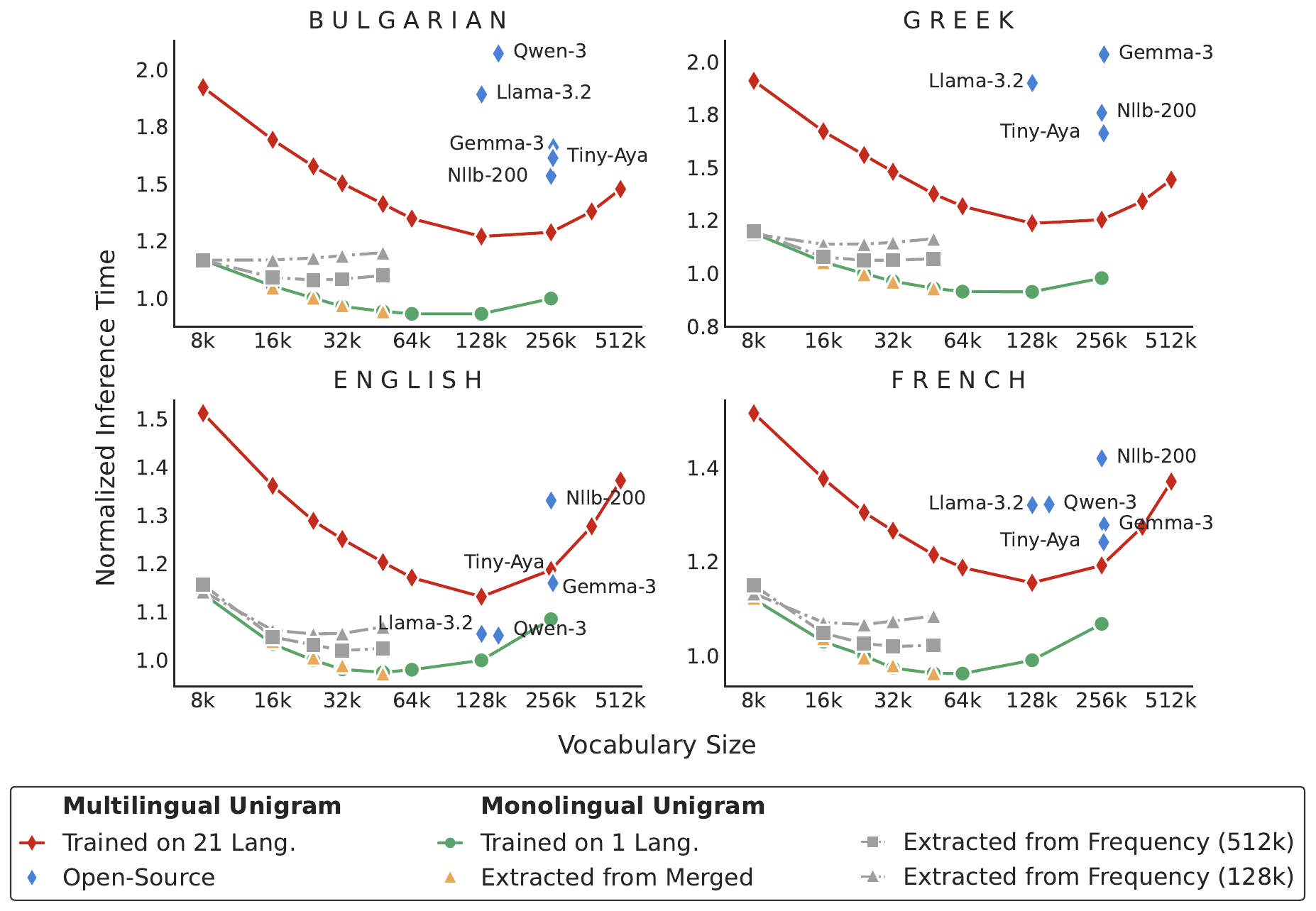}
    \caption{\textbf{Normalized inference time for Unigram Tokenizer}: We evaluate the inference time of a 2.5B parameter model (excluding embedding parameters) on FLORES-200 using both monolingual and multilingual Unigram tokenizers of varying vocabulary sizes. Monolingual tokenizers with a 24k vocabulary serve as the reference. We consider two strategies for deriving monolingual subtokenizers from multilingual tokenizers. First, \emph{Extracted from Merged} corresponds to subtokenizers obtained from a multilingual Unigram built by merging independently trained monolingual tokenizers, followed by probability renormalization, as described in \textsection~\ref{sec:unigram}. Second, \emph{Extracted from Frequency $v$} constructs subtokenizers of size $k$ by selecting the $k$ most frequent tokens from a multilingual tokenizer of size $v$, as described in \textsection~\ref{sec:unigram}. The curves for $v=256$k and $v=512$k overlap (not shown for clarity), showing diminishing returns when increasing the multilingual tokenizer size from $v=256$k to $v=512$k, and that subtokenizer compression remains bounded by the global tokenizer. }
    \label{fig:inference_time_comparison_unigram}
\end{figure*}

Fig.~\ref{fig:inference_time_comparison_unigram} reports normalized inference time for Unigram tokenizers. Extracted subtokenizers from the merged multilingual tokenizer achieve inference speeds comparable to independently trained monolingual tokenizers. Frequency-based subtokenizers also reduce inference time, but remain bounded by the compression of the base multilingual tokenizer, with diminishing returns beyond $v=256$k.

\section{Experimental Settings  }
\label{appendix:model_hyperparameters}

\begin{table}[h!]
\begin{minipage}{\columnwidth}
\renewcommand{\arraystretch}{1.1}
\setlength{\tabcolsep}{5pt}

\begin{tabular}{lc}
\toprule
\multicolumn{2}{c}{\textbf{Model Architecture}} \\
\midrule
\makecell[l]{Number of non-embedding \\ parameters} & 2.5B \\
Layers & 24 \\
Hidden dimension & 3072 \\
FFN latent dimension & 8448 \\
Number of heads & 24 \\
GQA & 2 \\
RoPE $\theta$ & 100,000 \\
Context length & 4096 \\
Activation function & SiGLU \\
Normalization & RMS pre-norm \\
\bottomrule
\end{tabular}

\captionof{table}{Model architecture details.}
\label{tab:model_architecture}
\end{minipage}

\hfill

\vspace{1em}

\begin{minipage}{\columnwidth}
\renewcommand{\arraystretch}{1.1}
\setlength{\tabcolsep}{5pt}

\begin{tabular}{l c}
\toprule
\textbf{Hyperparameters} & \textbf{Pre-training Settings} \\
\midrule

\multicolumn{2}{c}{\textit{Optimizer}} \\
Optimizer Type & AdamW \\
Beta 1 & 0.9 \\
Beta 2 & 0.95 \\
$\epsilon$ & $1\mathrm{e}{-8}$ \\
Weight Decay & 0.1 \\
Gradient Clipping & 1.0 \\

\midrule

\multicolumn{2}{c}{\textit{Scheduler}} \\
LR Scheduler & Cosine \\
Max LR & $3\mathrm{e}{-4}$ \\
Final LR & $3\mathrm{e}{-5}$ \\
Warmup Steps & 2000 \\
Number of Steps & 300k \\

\midrule

Batch Size & 64 \\
Training Tokens & 78B \\

\bottomrule
\end{tabular}

\captionof{table}{Pre-training hyperparameters.}
\label{tab:hyperparameters_pt}
\end{minipage}
\end{table}

\section{Multilingual Performance}

\subsection{Subtokenizer robustness at inference time.}
\label{subsec:combos_subtokenizer}

\begin{figure*}
\vspace{2em}
    \centering
    \begin{subfigure}[b]{0.48\textwidth}
    \includegraphics[width=\linewidth, clip]{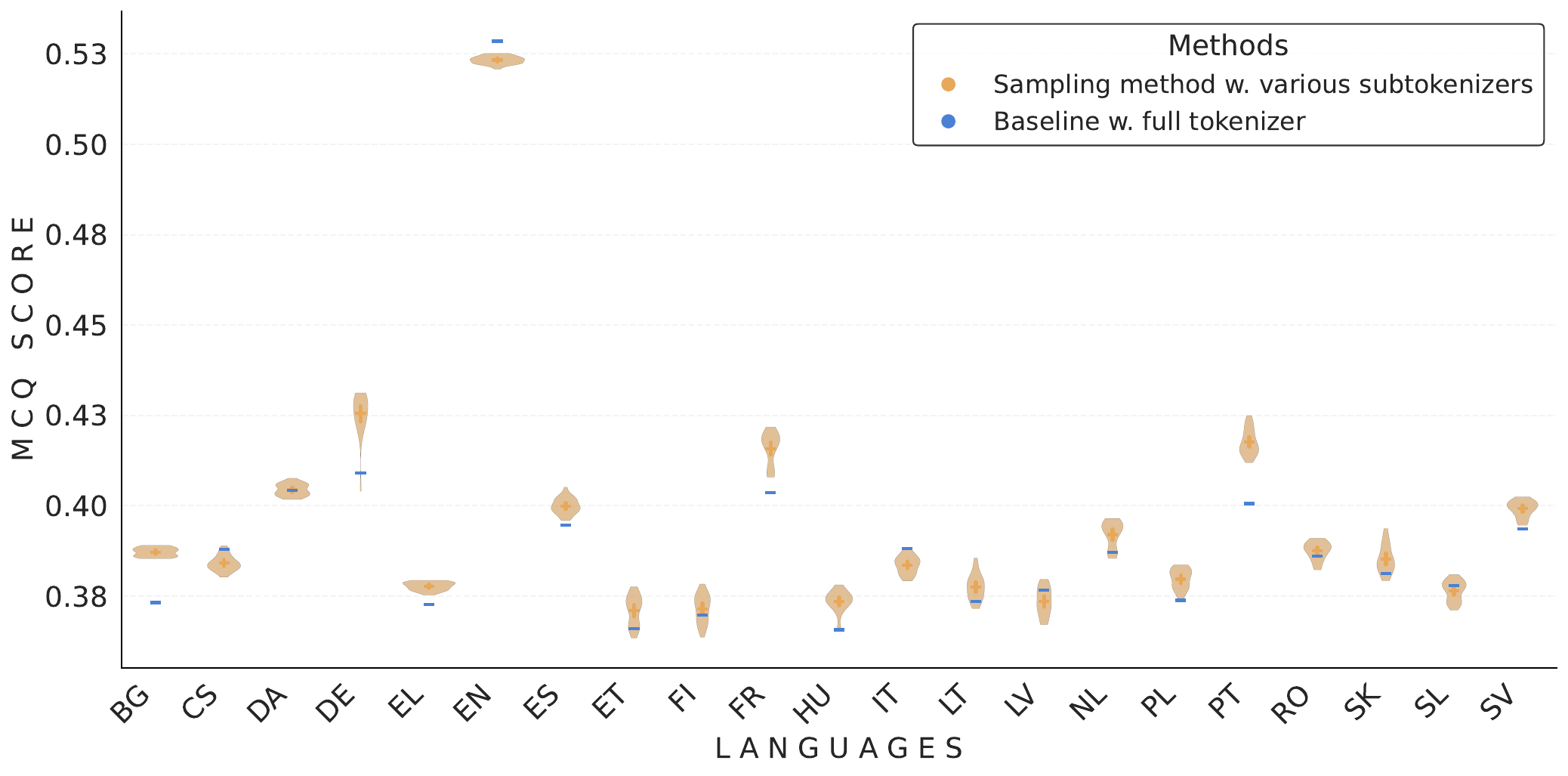}
    \caption{Multi-Choice Questioning}
    \label{fig:left_image}
    \end{subfigure}
    \hfill
    \begin{subfigure}[b]{0.48\textwidth}
    \includegraphics[width=\linewidth]{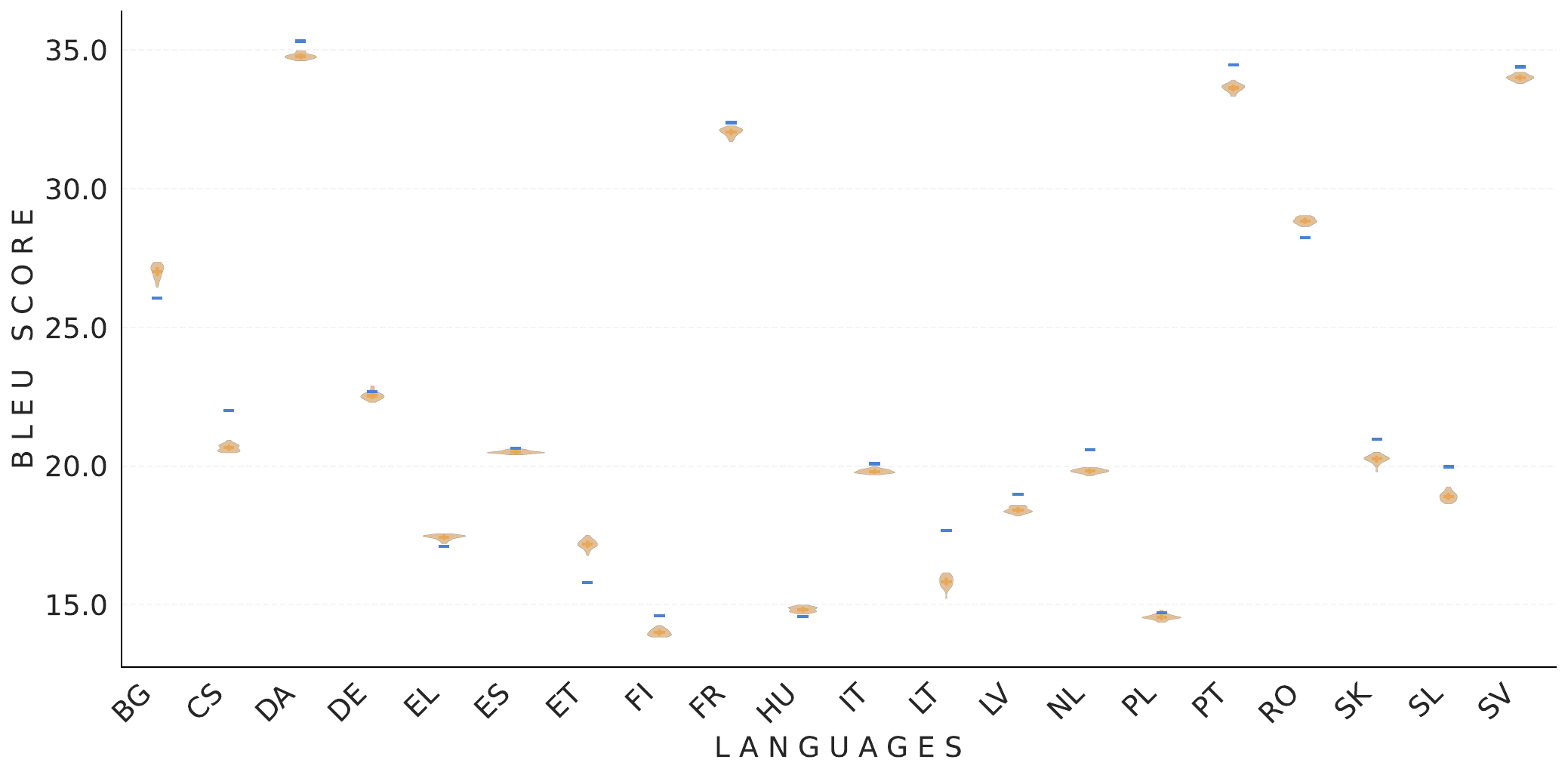}
    \caption{Machine Translation}
    \label{fig:mid_image}
    \end{subfigure}

    \caption{Comparison of the baseline and our sampling training at inference. Both are trained using Unigram tokenizers. We use the \Data{2} method at training time but evaluated it using subtokenizers covering between 2 and 10 languages to underline the robustness at inference time of this method.}
    \label{fig:results_w_combos_languages_atinference}
\vspace{1em}
\end{figure*}

The \Data{n} training method is designed to support monolingual subtokenizers, enhancing both fairness and efficiency across diverse languages. This approach also accommodates subtokenizers covering specific language groups, which is particularly beneficial for tasks like translation. In Fig.~\ref{fig:results_w_combos_languages_atinference}, we highlight that training with \Data{2} (in Unigram setting) enables the use of various subtokenizers at inference time covering between 2 and 10 languages with minimal variations on downstream performance. Notably, the score variance remains low, and average performance closely tracks the baseline that utilizes the full tokenizer for both training and inference.

\subsection{BPE Experiments}
\label{appendix:bpe_perf_results}
Tab.~\ref{Tab: main_result_bpe} presents MCQ and MT results for BPE tokenizers following the same setting as Tab.~\ref{Tab: main_result_unigram}, confirming that our modular tokenizer approach generalizes across both tokenization schemes. Detailed per-language results are reported in Tab.~\ref{tab:detailed_results_bpe}.

\begin{table*}[!h]
\centering
\footnotesize 
\setlength{\tabcolsep}{3pt} 
\renewcommand{\arraystretch}{1.3}

\begin{tabular}{l c l c c cc c cc c cc c cc c cc c cc c cc}
\toprule
\multirow{2}{*}{\textbf{Model}} & & \multirow{2}{*}{\textbf{Infer.}} & &
\textbf{EN} && 
\multicolumn{2}{c}{\textbf{BG}} && 
\multicolumn{2}{c}{\textbf{DA}} && 
\multicolumn{2}{c}{\textbf{EL}} && 
\multicolumn{2}{c}{\textbf{FR}} && 
\multicolumn{2}{c}{\textbf{IT}} && 
\multicolumn{2}{c}{\textbf{SK}} \\

\cmidrule(lr){5-5}
\cmidrule(lr){7-8}
\cmidrule(lr){10-11}
\cmidrule(lr){13-14}
\cmidrule(lr){16-17}
\cmidrule(lr){19-20}
\cmidrule(lr){22-23}

& & & &
MCQ && 
MCQ & MT && 
MCQ & MT &&
MCQ & MT && 
MCQ & MT && 
MCQ & MT && 
MCQ & MT \\

\midrule

Baseline 
& & Full & & \textbf{52.6} && 37.2 & 28.0 && 39.4 & \textbf{36.1} && 36.9 & \textbf{18.5} && 39.2 & \textbf{32.8} && 38.7 & \textbf{20.8} && 38.7 & \textbf{22.6 }\\
\midrule

\multirow{2}{*}{\shortstack[l]{\Nosampling} }
& & Lang & & 52.4 && \textbf{39.6} & 0.8 && 40.1 & 2.5 && \textbf{38.4} & 1.5 && 39.6 & 9.6 && 39.2 & 4.8 && 37.7 & 0.9 \\
& & Full & & 50.1 && 37.4 & 0.7 && 38.8 & 2.4 && 36.7 & 1.0 && 37.1 & 6.3 && 37.4 & 2.5 && 35.2 & 0.9 \\
\midrule

\multirow{2}{*}{\shortstack[l]{\Full}}
& & Lang & & 52.3 && 38.3 & 27.9 && 39.7 & 35.3 && 38.3 & 18.1 && \textbf{40.1} & 31.9 && 39.2 & 20.0 && 38.7 & 20.2 \\
& & Full & & 52.5 && 38.1 & 27.9 && 39.7 & 35.7 && 37.6 & 18.2 && 39.6 & 31.7 && 39.1 & 20.1 && 38.5 & 20.8 \\
\midrule

\multirow{2}{*}{\shortstack[l]{\Data{3}}}
& & Lang & & 52.5 && 38.6 & \textbf{28.1} && 39.9 & 36.0 && 38.0 & 18.3 && 39.8 & 32.5 && \textbf{39.4} & 20.3 && \textbf{39.0} & 21.4 \\
& & Full & & 52.4 && 38.2 & 27.7 && \textbf{40.2} & \textbf{36.1} && 37.8 & 18.3 && 39.4 & 32.3 && 38.8 & 20.6 && 38.5 & 21.7 \\
                            
\bottomrule
\end{tabular}
\caption{MCQ and MT performance under different pretraining sampling schemes for BPE tokenizers, following the same setting as Tab.~\ref{Tab: main_result_unigram}. \emph{Infer.} denotes the tokenizer used at inference; \emph{Full} corresponds to the full tokenizer; \emph{Lang} to the monolingual tokenizer for MCQ and the combined English--target tokenizer for MT.}
\label{Tab: main_result_bpe}
\vspace{1em}
\end{table*}

\begin{table*}[!h]
\scriptsize
\setlength{\tabcolsep}{5pt}
\renewcommand{\arraystretch}{1.2}
\centering
\begin{tabular}{l l c c ccc c ccc}
\toprule
& & & & \multicolumn{3}{c}{Lang.}& &  \multicolumn{3}{c}{Full}\\
\textbf{Lang.} & \textbf{Tasks} & Baseline & & \Data{3} & \Data{5} & \Full & & \Data{3} & \Data{5} & \Full  \\
\midrule

\multirow{7}{*}{\textbf{BG}}
& ARC-E & 51.3 & & \textbf{55.0} & 54.5 & 54.0 & & 53.1 & 54.9 & 54.0 \\
& ARC-C & 31.9 & & \textbf{32.6} & 32.3 & 32.1 & & 31.2 & 31.6 & 32.0 \\
& HellaSwag & 47.1 & & \textbf{48.3} & 47.5 & 48.0 & & 47.3 & 48.2 & 47.6 \\
& MMLU & 28.5 & & 28.9 & 29.0 & \textbf{29.7} & & 29.1 & 29.1 & 29.1 \\
& BeleBele & 27.1 & & 28.3 & 27.7 & \textbf{28.6} & & 27.9 & 27.8 & 27.7 \\
& Flores & 28.0 & & 28.1 & 27.7 & 28.1 & & \textbf{28.1} & 27.9 & 27.9 \\
& Avg. & 35.7 & & \textbf{36.9} & 36.4 & 36.7 & & 36.1 & 36.6 & 36.4 \\
\midrule
\multirow{7}{*}{\textbf{CS}}
& ARC-E & 55.1 & & 55.4 & 54.9 & \textbf{55.9} & & 54.3 & 54.5 & 54.9 \\
& ARC-C & 33.2 & & 31.5 & 30.7 & 30.5 & & 29.3 & \textbf{34.0} & 32.0 \\
& HellaSwag & 47.9 & & 48.0 & 47.3 & 48.3 & & 47.9 & \textbf{48.7} & 48.2 \\
& MMLU & \textbf{29.8} & & 29.4 & 29.0 & 29.4 & & 29.0 & 29.0 & 29.7 \\
& BeleBele & 29.0 & & \textbf{31.8} & 30.3 & 30.4 & & 29.7 & 30.9 & 28.3 \\
& Flores & \textbf{22.3} & & 21.5 & 21.2 & 22.3 & & 21.9 & 20.9 & 20.8 \\
& Avg. & 36.2 & & 36.3 & 35.6 & 36.1 & & 35.4 & \textbf{36.3} & 35.7 \\
\midrule
\multirow{7}{*}{\textbf{DA}}
& ARC-E & 56.2 & & 56.9 & \textbf{57.1} & 55.3 & & 56.2 & 56.0 & 56.6 \\
& ARC-C & 32.2 & & 32.1 & \textbf{33.3} & 32.3 & & 31.8 & 32.8 & 33.0 \\
& HellaSwag & 50.7 & & \textbf{51.5} & 50.6 & 50.1 & & 49.5 & 50.6 & 50.2 \\
& MMLU & 29.5 & & 29.1 & \textbf{29.8} & 29.1 & & 29.2 & 28.9 & 29.3 \\
& BeleBele & 28.6 & & 30.0 & \textbf{30.2} & 30.0 & & 30.0 & \textbf{30.2} & 29.6 \\
& Flores & \textbf{36.1} & & 36.0 & 36.1 & 35.7 & & 36.0 & 35.3 & 35.7 \\
& Avg. & 38.9 & & 39.3 & \textbf{39.5} & 38.7 & & 38.8 & 39.0 & 39.1 \\

\bottomrule
\end{tabular}
\end{table*}

\begin{table*}

\scriptsize
\setlength{\tabcolsep}{5pt}
\renewcommand{\arraystretch}{1.2}
\centering
\begin{tabular}{l l c c ccc c ccc}
\toprule
& & & & \multicolumn{3}{c}{Lang.}& &  \multicolumn{3}{c}{Full}\\
\textbf{Lang.} & \textbf{Tasks} & Baseline & & \Data{3} & \Data{5} & \Full & & \Data{3} & \Data{5} & \Full  \\

\midrule

\multirow{7}{*}{\textbf{DE}}
& ARC-E & 53.3 & & \textbf{53.7} & 53.3 & 52.7 & & 52.1 & 53.0 & 52.4 \\
& ARC-C & 30.9 & & 30.9 & 31.7 & 31.8 & & 31.2 & \textbf{34.8} & 34.0 \\
& HellaSwag & 45.8 & & \textbf{45.9} & 45.0 & 45.3 & & 45.0 & 45.8 & 45.7 \\
& MMLU & 29.3 & & 29.5 & \textbf{30.3} & 28.9 & & 29.2 & 29.0 & 29.1 \\
& BeleBele & 28.6 & & 28.2 & 28.0 & \textbf{29.7} & & 29.2 & 28.6 & 27.8 \\
& Flores & 22.6 & & 22.5 & 22.3 & 23.0 & & \textbf{23.2} & 21.6 & 21.7 \\
& Avg. & 35.1 & & 35.1 & 35.1 & 35.2 & & 35.0 & \textbf{35.4} & 35.1 \\
\midrule
\multirow{7}{*}{\textbf{EL}}
& ARC-E & 49.7 & & 51.3 & 50.7 & 50.9 & & 50.2 & 52.6 & \textbf{52.7} \\
& ARC-C & 31.7 & & 32.1 & 33.1 & \textbf{33.8} & & 32.3 & 33.5 & 31.1 \\
& HellaSwag & 46.0 & & \textbf{47.1} & 46.6 & 46.8 & & 46.5 & 46.5 & 46.2 \\
& MMLU & 28.1 & & 29.3 & \textbf{29.5} & 29.1 & & 28.6 & 28.8 & 29.3 \\
& BeleBele & 28.9 & & 30.1 & 28.9 & 27.9 & & 28.1 & \textbf{30.2} & 28.9 \\
& Flores & 18.5 & & 18.3 & 18.3 & \textbf{19.0} & & 19.0 & 18.1 & 18.2 \\
& Avg. & 33.8 & & 34.7 & 34.5 & 34.6 & & 34.1 & \textbf{35.0} & 34.4 \\
\midrule
\multirow{9}{*}{\textbf{EN}}
& ARC-E & 66.6 & & 67.7 & \textbf{68.0} & 67.3 & & 67.4 & 66.3 & 67.3 \\
& ARC-C & 39.2 & & 38.3 & 38.3 & 40.3 & & 39.3 & 40.2 & \textbf{40.9} \\
& CSQA & \textbf{61.5} & & 60.1 & 60.0 & 60.6 & & 60.5 & 60.4 & 59.1 \\
& HellaSwag & 62.6 & & \textbf{62.8} & 62.3 & 62.1 & & 61.7 & 62.1 & 62.1 \\
& PIQA & 75.1 & & 75.8 & 75.5 & \textbf{76.4} & & 76.0 & 74.7 & 75.4 \\
& PIQA & 49.0 & & 50.1 & 49.4 & 49.4 & & 49.6 & 50.5 & \textbf{50.7} \\
& MMLU & 34.1 & & \textbf{34.3} & 34.2 & 33.3 & & 33.1 & 33.7 & 33.4 \\
& BeleBele & \textbf{32.8} & & 31.2 & 31.4 & 32.3 & & 32.2 & 30.6 & 31.0 \\
& Avg. & 52.6 & & 52.5 & 52.4 & \textbf{52.7} & & 52.5 & 52.3 & 52.5 \\
\midrule
\multirow{7}{*}{\textbf{ES}}
& ARC-E & 55.7 & & 56.6 & 56.1 & 55.9 & & 55.6 & \textbf{57.0} & 56.7 \\
& ARC-C & 33.7 & & 34.3 & 33.9 & 33.3 & & 32.0 & 33.4 & \textbf{34.6} \\
& HellaSwag & 48.8 & & \textbf{49.3} & 48.5 & 49.1 & & 48.1 & 49.1 & 48.8 \\
& MMLU & 29.8 & & \textbf{30.4} & 30.2 & 29.8 & & 30.4 & 30.0 & 29.4 \\
& BeleBele & 29.0 & & \textbf{32.2} & 29.7 & 30.0 & & 30.3 & 30.7 & 30.1 \\
& Flores & \textbf{21.6} & & 20.8 & 20.7 & 21.2 & & 21.2 & 20.5 & 20.8 \\
& Avg. & 36.4 & & \textbf{37.3} & 36.5 & 36.5 & & 36.3 & 36.8 & 36.7 \\
\midrule
\multirow{7}{*}{\textbf{ET}}
& ARC-E & 49.6 & & 50.1 & 48.4 & 49.6 & & 48.8 & \textbf{50.9} & 50.7 \\
& ARC-C & 31.2 & & 30.5 & \textbf{31.5} & 30.5 & & 28.6 & 30.6 & 30.6 \\
& HellaSwag & 45.2 & & 45.8 & 45.1 & \textbf{46.0} & & 44.9 & 45.7 & 45.5 \\
& MMLU & 28.8 & & \textbf{29.0} & 28.9 & 28.5 & & 28.4 & 28.9 & 28.4 \\
& BeleBele & 27.6 & & 27.7 & 28.3 & 27.2 & & 27.7 & 28.4 & \textbf{29.8} \\
& Flores & 17.2 & & 17.4 & \textbf{17.4} & 16.9 & & 17.3 & 16.5 & 16.4 \\
& Avg. & 33.2 & & 33.4 & 33.3 & 33.1 & & 32.6 & 33.5 & \textbf{33.6} \\
\midrule
\multirow{7}{*}{\textbf{FI}}
& ARC-E & 51.3 & & \textbf{51.5} & 49.2 & 50.8 & & 49.3 & 51.3 & \textbf{51.5} \\
& ARC-C & 29.4 & & 30.4 & \textbf{30.8} & 29.5 & & 29.8 & 30.4 & 30.2 \\
& HellaSwag & 46.1 & & 46.9 & 45.3 & \textbf{46.9} & & 45.8 & 46.5 & 46.1 \\
& MMLU & 28.5 & & \textbf{29.1} & 28.3 & 28.6 & & 28.9 & 28.7 & 28.8 \\
& BeleBele & 28.3 & & 28.9 & 27.9 & \textbf{29.3} & & 28.2 & 29.0 & 29.0 \\
& Flores & 15.5 & & 15.3 & \textbf{15.6} & 15.5 & & 15.4 & 15.0 & 14.4 \\
& Avg. & 33.2 & & \textbf{33.7} & 32.8 & 33.4 & & 32.9 & 33.5 & 33.3 \\
\midrule
\multirow{7}{*}{\textbf{FR}}
& ARC-E & 54.8 & & 55.6 & 55.5 & 54.5 & & 54.0 & \textbf{56.1} & 55.6 \\
& ARC-C & 32.2 & & 33.0 & 32.5 & 33.6 & & 32.5 & \textbf{34.2} & 32.5 \\
& HellaSwag & \textbf{49.6} & & 49.5 & 48.6 & 49.4 & & 48.9 & 49.2 & 48.9 \\
& MMLU & \textbf{30.0} & & 29.8 & 29.7 & 29.2 & & 29.5 & 30.0 & 29.5 \\
& BeleBele & 29.6 & & 31.3 & 30.4 & 31.6 & & 30.6 & 31.1 & \textbf{31.7} \\
& Flores & 32.8 & & 32.5 & 32.3 & \textbf{33.0} & & 33.0 & 31.9 & 31.7 \\
& Avg. & 38.1 & & 38.6 & 38.2 & 38.6 & & 38.1 & \textbf{38.7} & 38.3 \\
\midrule
\multirow{7}{*}{\textbf{HU}}
& ARC-E & 51.0 & & 49.9 & 49.5 & \textbf{51.3} & & 50.3 & 50.5 & 50.6 \\
& ARC-C & 30.4 & & 29.3 & 30.3 & \textbf{32.4} & & 30.3 & 31.8 & 30.6 \\
& HellaSwag & 44.0 & & 44.3 & 43.6 & 44.4 & & 43.5 & \textbf{44.5} & 44.1 \\
& MMLU & 29.1 & & 29.1 & 28.5 & 28.5 & & \textbf{29.2} & 29.2 & 28.7 \\
& BeleBele & 28.7 & & \textbf{30.0} & 29.7 & \textbf{30.0} & & 28.9 & 28.8 & 29.2 \\
& Flores & 15.0 & & 15.1 & 14.8 & 15.0 & & \textbf{15.4} & 14.2 & 14.5 \\
& Avg. & 33.0 & & 33.0 & 32.7 & \textbf{33.6} & & 32.9 & 33.2 & 33.0 \\
\midrule
\multirow{3}{*}{\textbf{IT}}
& ARC-E & 52.9 & & 54.8 & 54.0 & 54.4 & & 52.9 & 55.8 & \textbf{56.0} \\
& ARC-C & \textbf{35.0} & & 32.6 & 32.8 & 32.7 & & 33.4 & 32.8 & 32.2 \\
& HellaSwag & 47.8 & & \textbf{48.7} & 47.7 & 48.2 & & 47.7 & 48.3 & 48.2 \\
& MMLU & 30.0 & & \textbf{30.5} & 29.7 & 28.9 & & 28.8 & 29.9 & 29.3 \\
& BeleBele & 27.9 & & \textbf{30.4} & 29.8 & 30.0 & & \textbf{30.4} & 29.3 & 29.8 \\
& Flores & \textbf{20.8} & & 20.3 & 20.6 & 20.5 & & 20.7 & 20.0 & 20.1 \\
& Avg. & 35.7 & & \textbf{36.2} & 35.8 & 35.8 & & 35.7 & 36.0 & 35.9 \\

\bottomrule
\end{tabular}
\end{table*}

\begin{table*}
\vspace{-3em}
\scriptsize
\setlength{\tabcolsep}{5pt}
\renewcommand{\arraystretch}{1.2}
\centering
\begin{tabular}{l l c c ccc c ccc}
\toprule
& & & & \multicolumn{3}{c}{Lang.}& &  \multicolumn{3}{c}{Full}\\
\textbf{Lang.} & \textbf{Tasks} & Baseline & & \Data{3} & \Data{5} & \Full & & \Data{3} & \Data{5} & \Full  \\
\midrule

\multirow{7}{*}{\textbf{LT}}
& ARC-E & 52.9 & & 53.3 & 52.6 & 53.5 & & 53.5 & \textbf{54.5} & 54.2 \\
& ARC-C & 31.9 & & 29.9 & 31.7 & 30.0 & & 30.4 & \textbf{32.8} & 31.7 \\
& HellaSwag & 46.5 & & 46.4 & 45.6 & \textbf{46.8} & & 45.5 & 46.5 & 46.0 \\
& MMLU & 29.3 & & 29.2 & 28.9 & \textbf{29.3} & & 28.7 & 28.8 & 28.9 \\
& BeleBele & 27.7 & & 28.6 & 28.3 & 28.4 & & 27.2 & \textbf{29.0} & 28.1 \\
& Flores & 17.9 & & \textbf{18.4} & 18.3 & 17.6 & & 17.7 & 17.7 & 18.1 \\
& Avg. & 34.4 & & 34.3 & 34.2 & 34.3 & & 33.8 & \textbf{34.9} & 34.5 \\
\midrule
\multirow{7}{*}{\textbf{LV}}
& ARC-E & 53.2 & & 53.4 & 51.4 & 53.5 & & 51.3 & \textbf{53.5} & 53.1 \\
& ARC-C & 31.1 & & 33.3 & 31.4 & 31.7 & & 30.0 & \textbf{33.4} & 32.8 \\
& HellaSwag & 46.3 & & \textbf{46.9} & 46.0 & 46.5 & & 45.6 & 46.5 & 46.5 \\
& MMLU & 28.8 & & 29.1 & 28.5 & 29.1 & & 27.9 & \textbf{29.7} & 28.1 \\
& BeleBele & 27.9 & & 29.9 & \textbf{30.0} & 28.3 & & 28.1 & 27.2 & 29.0 \\
& Flores & 20.2 & & 19.9 & 19.8 & 20.4 & & \textbf{20.4} & 19.6 & 19.9 \\
& Avg. & 34.6 & & \textbf{35.4} & 34.5 & 34.9 & & 33.9 & 35.0 & 34.9 \\
\midrule
\multirow{7}{*}{\textbf{NL}}
& ARC-E & 53.2 & & 53.4 & 52.5 & 52.3 & & 51.4 & \textbf{54.6} & 53.7 \\
& ARC-C & 29.3 & & 31.0 & \textbf{32.8} & 31.1 & & 31.1 & 32.3 & 30.5 \\
& HellaSwag & 47.7 & & 47.7 & 47.1 & 48.2 & & 47.6 & \textbf{48.5} & 48.3 \\
& MMLU & 29.0 & & 28.9 & \textbf{29.5} & 28.5 & & 29.3 & 28.5 & 28.9 \\
& BeleBele & 30.0 & & 31.3 & 30.4 & \textbf{31.7} & & 30.8 & 31.1 & 30.3 \\
& Flores & \textbf{20.9} & & 20.5 & 20.3 & 20.3 & & 20.2 & 20.2 & 20.1 \\
& Avg. & 35.0 & & 35.5 & 35.4 & 35.3 & & 35.1 & \textbf{35.9} & 35.3 \\
\midrule
\multirow{7}{*}{\textbf{PL}}
& ARC-E & \textbf{53.9} & & 51.6 & 50.8 & 51.1 & & 49.8 & 52.9 & 53.0 \\
& ARC-C & 32.5 & & 30.8 & \textbf{33.5} & 32.3 & & 32.3 & 31.7 & 33.2 \\
& HellaSwag & 46.6 & & 46.5 & 45.9 & \textbf{47.0} & & 46.1 & 46.8 & 46.6 \\
& MMLU & 28.6 & & 28.6 & 28.3 & 28.6 & & 28.1 & \textbf{29.0} & 28.3 \\
& BeleBele & 30.3 & & 30.7 & 27.4 & 30.1 & & 29.7 & 30.6 & \textbf{31.2} \\
& Flores & \textbf{15.2} & & 14.5 & 14.4 & 14.4 & & 14.7 & 14.0 & 14.3 \\
& Avg. & \textbf{34.5} & & 33.8 & 33.4 & 33.9 & & 33.4 & 34.2 & 34.4 \\
\midrule
\multirow{7}{*}{\textbf{PT}}
& ARC-E & 53.9 & & \textbf{54.3} & 54.1 & 53.9 & & 53.4 & 53.6 & 53.5 \\
& ARC-C & 33.8 & & 31.8 & 33.0 & \textbf{34.2} & & 33.0 & 32.6 & 31.9 \\
& HellaSwag & 47.6 & & \textbf{48.3} & 47.0 & 48.2 & & 46.9 & 48.0 & 48.0 \\
& MMLU & 29.6 & & 29.8 & 29.7 & \textbf{29.9} & & 29.8 & 29.8 & 29.8 \\
& BeleBele & 27.9 & & 29.3 & 29.0 & \textbf{30.0} & & 29.7 & 29.0 & 29.2 \\
& Flores & \textbf{35.5} & & 34.3 & 34.3 & 34.6 & & 34.9 & 33.5 & 34.2 \\
& Avg. & 38.0 & & 38.0 & 37.9 & \textbf{38.5} & & 37.9 & 37.8 & 37.8 \\
\midrule
\multirow{7}{*}{\textbf{RO}}
& ARC-E & 54.2 & & 54.5 & 53.3 & 54.0 & & \textbf{54.8} & 53.4 & 54.4 \\
& ARC-C & 30.5 & & 30.7 & 31.5 & \textbf{32.1} & & 31.7 & 31.2 & 30.2 \\
& HellaSwag & 47.6 & & \textbf{48.2} & 47.0 & 47.8 & & 46.6 & 48.1 & 47.8 \\
& MMLU & 29.7 & & \textbf{30.2} & 30.2 & 29.4 & & 28.9 & 29.9 & 29.8 \\
& BeleBele & 28.4 & & \textbf{29.3} & 28.1 & 29.2 & & 28.2 & 28.3 & 28.0 \\
& Flores & \textbf{29.8} & & 28.6 & 28.7 & 28.9 & & 29.0 & 28.3 & 28.8 \\
& Avg. & 36.7 & & \textbf{36.9} & 36.4 & 36.9 & & 36.5 & 36.5 & 36.5 \\
\midrule
\multirow{7}{*}{\textbf{SK}}
& ARC-E & 53.4 & & \textbf{54.8} & 53.7 & 53.8 & & 53.4 & 53.9 & 53.8 \\
& ARC-C & \textbf{32.7} & & 31.9 & \textbf{32.7} & 32.3 & & 31.4 & 31.3 & 30.3 \\
& HellaSwag & 47.5 & & 47.6 & 46.9 & 47.5 & & 47.0 & \textbf{47.7} & 47.6 \\
& MMLU & 29.3 & & 29.0 & 28.6 & \textbf{30.1} & & 29.4 & 29.5 & 29.5 \\
& BeleBele & 30.9 & & \textbf{31.6} & 30.6 & 31.1 & & 29.9 & 31.0 & 31.4 \\
& Flores & \textbf{22.6} & & 21.4 & 21.7 & 21.7 & & 21.9 & 20.2 & 20.8 \\
& Avg. & 36.1 & & 36.0 & 35.7 & \textbf{36.1} & & 35.5 & 35.6 & 35.6 \\
\midrule
\multirow{7}{*}{\textbf{SL}}
& ARC-E & 53.5 & & \textbf{54.1} & 52.7 & 52.9 & & 52.6 & 53.7 & 53.5 \\
& ARC-C & 32.3 & & 31.2 & 31.8 & 30.6 & & 30.5 & \textbf{32.9} & 31.7 \\
& HellaSwag & 46.7 & & \textbf{47.1} & 46.0 & 46.2 & & 45.7 & 46.8 & 46.3 \\
& MMLU & 28.5 & & 29.2 & 28.8 & 28.4 & & 28.3 & 28.9 & \textbf{29.5} \\
& BeleBele & 29.9 & & 29.9 & 27.6 & 29.9 & & \textbf{30.1} & 29.1 & 29.4 \\
& Flores & 19.7 & & 19.7 & 19.8 & 19.6 & & \textbf{19.9} & 18.8 & 19.2 \\
& Avg. & 35.1 & & \textbf{35.2} & 34.4 & 34.6 & & 34.5 & 35.0 & 34.9 \\
\midrule
\multirow{7}{*}{\textbf{SV}}
& ARC-E & 54.9 & & \textbf{56.4} & 54.7 & 53.7 & & 53.5 & 54.5 & 54.9 \\
& ARC-C & \textbf{34.6} & & 32.0 & 33.4 & 32.7 & & 30.8 & 32.2 & 31.8 \\
& HellaSwag & 49.4 & & \textbf{49.7} & 49.0 & 48.9 & & 48.1 & 49.2 & 49.7 \\
& MMLU & 29.3 & & \textbf{30.0} & 29.9 & 28.8 & & 29.2 & 29.6 & 29.4 \\
& BeleBele & 30.2 & & 29.8 & 29.3 & 29.1 & & 29.0 & \textbf{30.4} & 30.3 \\
& Flores & 34.2 & & 34.2 & 34.0 & 34.5 & & \textbf{34.6} & 33.6 & 33.9 \\
& Avg. & \textbf{38.8} & & 38.7 & 38.4 & 38.0 & & 37.5 & 38.2 & 38.3 \\

\bottomrule

\end{tabular}
\caption{Detailed results for BPE experiments}
\label{tab:detailed_results_bpe}
\end{table*}

\newpage

\subsection{Translation Performance}

\begin{table*}
\centering
\scriptsize
\setlength{\tabcolsep}{2.5pt}
\renewcommand{\arraystretch}{1.2}
\centering

\begin{tabular}{c l ccccccccccccccccccccc}
\toprule
\textbf{$\rightarrow$ Lang.} & \textbf{Model} 
& \textbf{BG} & \textbf{CS} & \textbf{DA} & \textbf{DE} & \textbf{EL} & \textbf{EN}
& \textbf{ES} & \textbf{ET} & \textbf{FI} & \textbf{FR} & \textbf{HU} 
& \textbf{IT} & \textbf{LT} & \textbf{LV} & \textbf{NL} & \textbf{PL} 
& \textbf{PT} & \textbf{RO} & \textbf{SK} & \textbf{SL} & \textbf{SV} \\
\midrule
\multirow{2}{*}{BG} &
Baseline & - & \textbf{21.5} & \textbf{21.8} & \textbf{21.2} & \textbf{15.8} & \textbf{29.9} & \textbf{17.5} & \textbf{17.3} & \textbf{16.3} & \textbf{21.6} & \textbf{13.7} & \textbf{16.6} & \textbf{17.0} & \textbf{18.4} & \textbf{16.1} & \textbf{18.0} & \textbf{22.6} & \textbf{21.5} & \textbf{22.2} & \textbf{18.8} & \textbf{21.1} \\
& \Data{5}
& - & 19.3 & 1.8 & 19.2 & 0.5 & 29.6 & 16.6 & 14.2 & 13.8 & 15.9 & 5.2 & 11.8 & 16.9 & 18.0 & 10.8 & 13.9 & 19.5 & 8.3 & 20.9 & 1.2 & 8.4 \\

\midrule
\multirow{2}{*}{CS} &
Baseline
& \textbf{16.0} & - & \textbf{18.9} & \textbf{19.1} & \textbf{14.7} & 23.4 & \textbf{13.2} & \textbf{14.4} & 10.3 & \textbf{17.0} & \textbf{13.1} & \textbf{13.9} & \textbf{14.6} & \textbf{15.0} & \textbf{13.4} & \textbf{16.0} & \textbf{16.5} & \textbf{17.5} & \textbf{26.5} & 9.3 & \textbf{17.4} \\
& \Data{5}
& 4.2 & - & 7.5 & 18.3 & 12.8 & \textbf{23.6} & 11.8 & 11.7 & \textbf{11.0} & 12.6 & 9.0 & 11.0 & 12.6 & 13.5 & 12.1 & 13.6 & 14.7 & 15.2 & 23.7 & \textbf{14.6} & 11.0 \\

\midrule
\multirow{2}{*}{DA} &
Baseline
& 21.6 & 23.3 & - & \textbf{25.2} & 19.5 & \textbf{38.1} & 18.5 & \textbf{19.6} & 18.5 & \textbf{24.1} & 16.2 & \textbf{19.5} & 17.7 & 19.1 & 18.5 & 17.6 & 25.2 & 22.9 & 22.6 & \textbf{20.4} & \textbf{33.5} \\
& \Data{5}
& \textbf{23.2} & \textbf{23.4} & - & 25.1 & \textbf{20.2} & 37.4 & \textbf{18.8} & 18.3 & \textbf{19.2} & 22.8 & \textbf{17.9} & 18.6 & \textbf{18.1} & \textbf{19.2} & \textbf{19.2} & \textbf{17.6} & \textbf{26.4} & \textbf{24.4} & \textbf{23.1} & 19.9 & 32.5 \\
\midrule
\multirow{2}{*}{DE} &
Baseline
& \textbf{17.3} & \textbf{19.8} & \textbf{21.2} & - & \textbf{15.3} & \textbf{25.1} & \textbf{14.5} & \textbf{15.8} & \textbf{13.9} & \textbf{18.9} & 13.1 & \textbf{15.8} & \textbf{14.1} & \textbf{15.1} & \textbf{17.0} & \textbf{14.9} & \textbf{19.7} & \textbf{18.8} & \textbf{19.1} & \textbf{16.3} & \textbf{20.4} \\
& \Data{5}
& 16.0 & 16.8 & 6.6 & - & 14.5 & 24.6 & 13.3 & 12.1 & 12.7 & 16.5 & \textbf{13.9} & 14.5 & 13.0 & 13.7 & 13.8 & 13.6 & 16.7 & 17.8 & 17.2 & 14.8 & 18.9 \\
\midrule
\multirow{2}{*}{EL} &
Baseline
& \textbf{11.0} & \textbf{13.4} & \textbf{14.9} & \textbf{14.4} & - & 19.3 & \textbf{13.3} & \textbf{10.0} & \textbf{9.4} & \textbf{16.0} & \textbf{7.2} & \textbf{13.1} & \textbf{8.5} & \textbf{10.5} & \textbf{10.3} & \textbf{9.7} & \textbf{15.4} & \textbf{13.0} & \textbf{12.9} & \textbf{11.6} & \textbf{14.1} \\

& \Data{5}
& 5.9 & 11.2 & 13.9 & 12.1 & - & \textbf{19.6} & 10.8 & 7.8 & 7.6 & 11.9 & 5.8 & 7.0 & 7.7 & 9.4 & 9.2 & 8.1 & 13.6 & 11.5 & 11.3 & 8.3 & 11.4 \\
\midrule
\multirow{2}{*}{EN} &
Baseline
& \textbf{35.4} & \textbf{34.1} & \textbf{44.7} & \textbf{37.5} & \textbf{30.1} & - & \textbf{27.5} & \textbf{29.1} & \textbf{27.8} & \textbf{39.0} & \textbf{26.7} & \textbf{29.8} & \textbf{27.2} & \textbf{30.0} & \textbf{28.8} & \textbf{25.3} & \textbf{42.3} & \textbf{37.6} & \textbf{33.5} & \textbf{29.7} & \textbf{44.0} \\
& \Data{5}
& 34.7 & 32.4 & 43.2 & 34.5 & 29.8 & - & 27.1 & 26.9 & 25.8 & 36.4 & 25.4 & 29.1 & 25.4 & 28.3 & 28.4 & 23.8 & 41.9 & 35.7 & 31.7 & 27.0 & 42.5 \\
\midrule
\multirow{2}{*}{ES} &
Baseline
& \textbf{16.2} & \textbf{16.9} & \textbf{17.4} & \textbf{17.6} & \textbf{16.1} & \textbf{22.2} & - & \textbf{14.0} & \textbf{14.1} & \textbf{21.0} & 13.2 & \textbf{19.1} & 12.4 & \textbf{13.1} & \textbf{14.1} & 15.5 & \textbf{19.7} & 19.7 & \textbf{16.9} & \textbf{14.9} & \textbf{17.1} \\
& \Data{5}
& 16.0 & 15.5 & 9.3 & 16.8 & 15.2 & 21.7 & - & 11.1 & 13.2 & 20.3 & \textbf{14.4} & 18.0 & \textbf{13.5} & 12.7 & 13.4 & \textbf{15.5} & 17.8 & \textbf{19.9} & 16.2 & 13.2 & 15.2 \\
\midrule
\multirow{2}{*}{ET} &
Baseline
& \textbf{9.6} & 12.1 & \textbf{12.6} & \textbf{12.5} & 5.5 & \textbf{18.9} & \textbf{8.2} & - & 10.7 & \textbf{11.2} & 3.5 & 6.4 & 6.1 & 8.6 & \textbf{8.2} & 8.5 & \textbf{9.2} & 8.5 & 9.2 & \textbf{9.4} & \textbf{12.6} \\
& \Data{5}
& 6.3 & \textbf{12.9} & 5.6 & 11.3 & \textbf{7.6} & 17.7 & 7.4 & - & \textbf{12.2} & 4.8 & \textbf{7.5} & \textbf{7.7} & \textbf{12.7} & \textbf{9.4} & 8.1 & \textbf{10.6} & 5.2 & \textbf{9.3} & \textbf{10.3} & 9.1 & 10.9 \\
\midrule
\multirow{2}{*}{FI} &
Baseline
& \textbf{9.0} & \textbf{12.2} & \textbf{13.7} & \textbf{13.2} & \textbf{7.9} & \textbf{17.1} & \textbf{9.6} & \textbf{13.2} & - & \textbf{12.2} & \textbf{7.2} & \textbf{9.9} & \textbf{10.3} & \textbf{11.8} & \textbf{9.6} & \textbf{9.1} & \textbf{12.1} & \textbf{11.2} & \textbf{12.3} & \textbf{10.6} & \textbf{13.6} \\
& \Data{5}
& 3.7 & 11.1 & 10.8 & 11.8 & 6.4 & 16.7 & 8.9 & 4.5 & - & 7.8 & 5.2 & 7.3 & 8.6 & 10.8 & 7.5 & 8.6 & 7.8 & 9.5 & 10.5 & 5.7 & 6.8 \\
\midrule
\multirow{2}{*}{FR} &
Baseline
& \textbf{20.7} & 20.9 & \textbf{21.0} & 20.1 & \textbf{21.3} & \textbf{35.1} & 18.5 & \textbf{15.4} & 15.2 & - & 12.6 & 18.6 & 13.3 & 14.4 & \textbf{18.0} & 15.7 & 20.6 & 27.0 & \textbf{22.1} & \textbf{19.1} & \textbf{24.0} \\
& \Data{5}
& 20.6 & \textbf{20.9} & 15.6 & \textbf{22.2} & 19.2 & 34.5 & \textbf{20.2} & 11.8 & \textbf{17.1} & - & \textbf{17.0} & \textbf{23.5} & \textbf{17.5} & \textbf{14.9} & 16.8 & \textbf{19.2} & \textbf{29.7} & \textbf{28.9} & 20.4 & 16.4 & 20.7 \\
\midrule
\multirow{2}{*}{HU} &
Baseline
& \textbf{10.6} & \textbf{13.3} & \textbf{13.7} & \textbf{14.9} & \textbf{8.7} & \textbf{16.6} & \textbf{11.1} & \textbf{11.1} & \textbf{8.6} & \textbf{12.7} & - & \textbf{10.9} & \textbf{8.2} & \textbf{9.9} & \textbf{10.8} & \textbf{8.8} & \textbf{13.3} & \textbf{12.0} & \textbf{12.0} & \textbf{12.2} & \textbf{13.6} \\
& \Data{5}
& 2.3 & 9.6 & 10.6 & 12.5 & 0.7 & 16.1 & 8.5 & 4.2 & 6.8 & 7.4 & - & 6.0 & 8.1 & 7.7 & 8.2 & 8.3 & 6.5 & 7.1 & 9.2 & 9.7 & 7.6 \\
\midrule
\multirow{2}{*}{IT} &
Baseline
& \textbf{14.8} & \textbf{17.3} & \textbf{17.7} & \textbf{17.8} & \textbf{15.0} & \textbf{22.5} & \textbf{18.7} & \textbf{14.0} & \textbf{12.1} & \textbf{23.0} & \textbf{12.8} & - & \textbf{13.0} & \textbf{13.6} & \textbf{13.9} & \textbf{14.6} & \textbf{20.8} & \textbf{20.7} & \textbf{17.1} & \textbf{14.4} & \textbf{15.7} \\
& \Data{5}
& 8.8 & 15.0 & 10.8 & 14.6 & 5.2 & 20.5 & 15.8 & 7.1 & 11.2 & 19.2 & 10.9 & - & 11.2 & 10.5 & 11.9 & 11.9 & 19.2 & 18.8 & 14.6 & 9.8 & 11.9 \\
\midrule
\multirow{2}{*}{LT} &
Baseline
& \textbf{8.3} & \textbf{12.4} & \textbf{13.4} & \textbf{14.6} & \textbf{7.7} & \textbf{19.6} & \textbf{11.2} & \textbf{12.1} & \textbf{10.8} & \textbf{14.4} & \textbf{2.7} & \textbf{7.6} & - & \textbf{14.7} & \textbf{9.1} & \textbf{10.4} & \textbf{8.7} & \textbf{10.2} & \textbf{9.4} & \textbf{7.1} & \textbf{12.7} \\
& \Data{5}
& 0.7 & 12.2 & 11.9 & 13.2 & 0.4 & 19.2 & 8.5 & 1.3 & 9.3 & 6.7 & 1.1 & 1.9 & - & 8.1 & 7.7 & 6.1 & 4.3 & 9.3 & 6.2 & 2.2 & 9.9 \\
\midrule
\multirow{2}{*}{LV} &
Baseline
& \textbf{12.3} & 10.9 & \textbf{14.6} & \textbf{15.2} & \textbf{8.7} & \textbf{21.3} & \textbf{11.3} & \textbf{13.2} & \textbf{12.1} & \textbf{13.9} & 4.9 & \textbf{11.4} & \textbf{13.8} & - & 9.4 & 9.8 & \textbf{12.3} & 11.3 & 11.4 & \textbf{11.2} & \textbf{11.8} \\
& \Data{5}
& 2.6 & \textbf{14.0} & 7.1 & 14.0 & 3.6 & 20.0 & 9.0 & 2.3 & 10.3 & 6.1 & \textbf{8.2} & 7.5 & 10.5 & - & \textbf{9.7} & \textbf{10.0} & 11.4 & \textbf{11.7} & \textbf{13.0} & 7.6 & 11.5 \\
\midrule
\multirow{2}{*}{NL} &
Baseline
& \textbf{14.7} & \textbf{16.2} & \textbf{18.5} & 19.0 & \textbf{13.7} & \textbf{21.6} & \textbf{13.6} & \textbf{13.1} & \textbf{13.3} & \textbf{16.6} & \textbf{12.3} & \textbf{14.5} & \textbf{12.3} & \textbf{13.0} & - & \textbf{13.7} & \textbf{16.4} & \textbf{16.0} & \textbf{16.2} & \textbf{14.1} & \textbf{17.9} \\
& \Data{5}
& 11.2 & 14.7 & 7.8 & \textbf{19.1} & 11.3 & 21.3 & 12.6 & 10.6 & 11.1 & 13.5 & 11.0 & 13.6 & 11.7 & 12.0 & - & 12.9 & 15.5 & 14.6 & 14.8 & 9.2 & 14.8 \\
\midrule
\multirow{2}{*}{PL} &
Baseline
& \textbf{11.6} & \textbf{14.0} & \textbf{12.7} & \textbf{14.0} & \textbf{8.5} & \textbf{15.9} & \textbf{11.6} & \textbf{9.7} & \textbf{8.9} & \textbf{14.6} & \textbf{8.2} & \textbf{11.8} & \textbf{9.8} & \textbf{10.7} & \textbf{10.5} & - & \textbf{12.4} & \textbf{12.3} & \textbf{13.7} & \textbf{9.5} & \textbf{11.9} \\
& \Data{5}
& 3.2 & 6.0 & 5.3 & 11.8 & 3.1 & 15.4 & 6.9 & 2.0 & 4.3 & 6.0 & 3.3 & 1.5 & 8.8 & 3.5 & 6.7 & - & 4.2 & 6.1 & 5.8 & 4.1 & 2.6 \\
\midrule
\multirow{2}{*}{PT} &
Baseline
& \textbf{23.6} & \textbf{22.9} & \textbf{26.4} & \textbf{24.8} & \textbf{21.2} & \textbf{37.6} & \textbf{22.9} & \textbf{20.1} & \textbf{18.4} & \textbf{31.2} & \textbf{17.7} & \textbf{24.8} & \textbf{18.1} & \textbf{19.2} & \textbf{18.9} & \textbf{18.3} & - & \textbf{28.1} & \textbf{23.0} & \textbf{20.1} & \textbf{25.5} \\
& \Data{5}
& 16.9 & 20.5 & 18.2 & 23.7 & 13.4 & 36.7 & 21.3 & 15.3 & 17.1 & 28.6 & 15.3 & 23.0 & 15.5 & 15.7 & 16.9 & 16.4 & - & 27.6 & 19.8 & 17.3 & 21.2 \\
\midrule
\multirow{2}{*}{RO} &
Baseline
& \textbf{20.2} & \textbf{21.7} & \textbf{24.2} & \textbf{23.7} & \textbf{18.4} & \textbf{31.8} & \textbf{21.2} & \textbf{15.4} & \textbf{15.4} & \textbf{27.9} & \textbf{14.2} & \textbf{22.1} & \textbf{14.7} & \textbf{16.0} & \textbf{16.7} & \textbf{16.6} & \textbf{27.6} & - & \textbf{21.2} & \textbf{19.0} & \textbf{22.7} \\
& \Data{5}
& 7.8 & 19.1 & 22.1 & 22.8 & 3.9 & 30.5 & 17.4 & 12.1 & 11.4 & 18.7 & 6.7 & 5.9 & 13.5 & 11.7 & 11.0 & 12.5 & 20.8 & - & 13.1 & 7.1 & 19.2 \\
\midrule
\multirow{2}{*}{SK} &
Baseline
& \textbf{17.6} & \textbf{27.1} & \textbf{19.4} & \textbf{20.3} & \textbf{14.0} & \textbf{24.4} & \textbf{14.2} & \textbf{14.2} & \textbf{12.6} & \textbf{18.6} & \textbf{12.1} & \textbf{15.3} & \textbf{14.2} & \textbf{15.1} & \textbf{14.2} & \textbf{15.3} & \textbf{18.2} & \textbf{16.9} & - & \textbf{16.9} & \textbf{19.3} \\
& \Data{5}
& 4.5 & 23.3 & 16.0 & 18.6 & 13.0 & 22.9 & 11.9 & 9.5 & 10.0 & 14.5 & 7.7 & 10.9 & 13.5 & 11.9 & 11.2 & 12.6 & 14.9 & 13.4 & - & 10.3 & 14.6 \\
\midrule
\multirow{2}{*}{SL} &
Baseline
& \textbf{15.8} & \textbf{18.7} & \textbf{17.1} & \textbf{16.5} & \textbf{13.6} & \textbf{21.1} & \textbf{12.6} & \textbf{14.7} & \textbf{12.7} & \textbf{15.8} & \textbf{13.0} & \textbf{12.5} & \textbf{13.0} & \textbf{15.1} & \textbf{12.9} & \textbf{14.2} & \textbf{14.6} & \textbf{15.6} & \textbf{18.2} & - & \textbf{16.5} \\
& \Data{5}
& 9.3 & 17.3 & 7.6 & 15.0 & 12.1 & 18.7 & 10.2 & 10.5 & 11.6 & 12.0 & 11.4 & 11.2 & 12.6 & 12.2 & 11.0 & 13.1 & 9.5 & 14.3 & 17.3 & - & 13.0 \\
\midrule
\multirow{2}{*}{SV} &
Baseline
& 17.5 & \textbf{22.0} & \textbf{32.2} & \textbf{24.2} & 14.5 & \textbf{37.2} & \textbf{16.5} & 16.0 & \textbf{18.5} & \textbf{22.8} & 13.8 & \textbf{17.3} & 15.4 & \textbf{16.5} & 16.6 & \textbf{14.8} & 22.8 & 21.2 & \textbf{20.9} & 17.4 & - \\
& \Data{5}
& \textbf{18.8} & 20.7 & 7.4 & 23.9 & \textbf{16.9} & 36.2 & 16.5 & \textbf{16.3} & 16.8 & 20.8 & \textbf{14.5} & 16.5 & \textbf{15.5} & 16.3 & \textbf{18.0} & 14.7 & \textbf{22.9} & \textbf{21.6} & 19.4 & \textbf{17.6} & - \\

\bottomrule
\end{tabular}
\caption{Machine translation results for all language-to-language directions. We compare the baseline against the sampling method. In both cases, we use the full tokenizer at inference.}
\label{tab:translation_tok_group}
\end{table*}

Tab.~\ref{tab:translation_tok_group} evaluates translation across more diverse language pairs to assess language switching. We modify the training distribution to 40\% English and 60\% other languages to improve non-English translation, keeping all other hyperparameters unchanged. We compare \Data{5} with the baseline, both using a Unigram tokenizer.

Both models show improved performance when translating directly between non-English languages. However, results vary depending on the language pair: \Data{5} sometimes matches the baseline, but in other cases performs worse. Analysis of generated outputs shows that even the baseline struggles to correctly switch languages, which is consistent with prior work relying on English as a pivot for such translations \citep{imamura-etal-2023-pivot}. These limitations are primarily due to insufficient training. Direct translation between non-English languages requires strong cross-lingual alignment and reliable language switching, which our models have not yet fully acquired. In practice, strong performance on this task typically depends on training with substantially more tokens.

\begin{figure*}[t]
    \centering
    \includegraphics[width=0.6\linewidth]{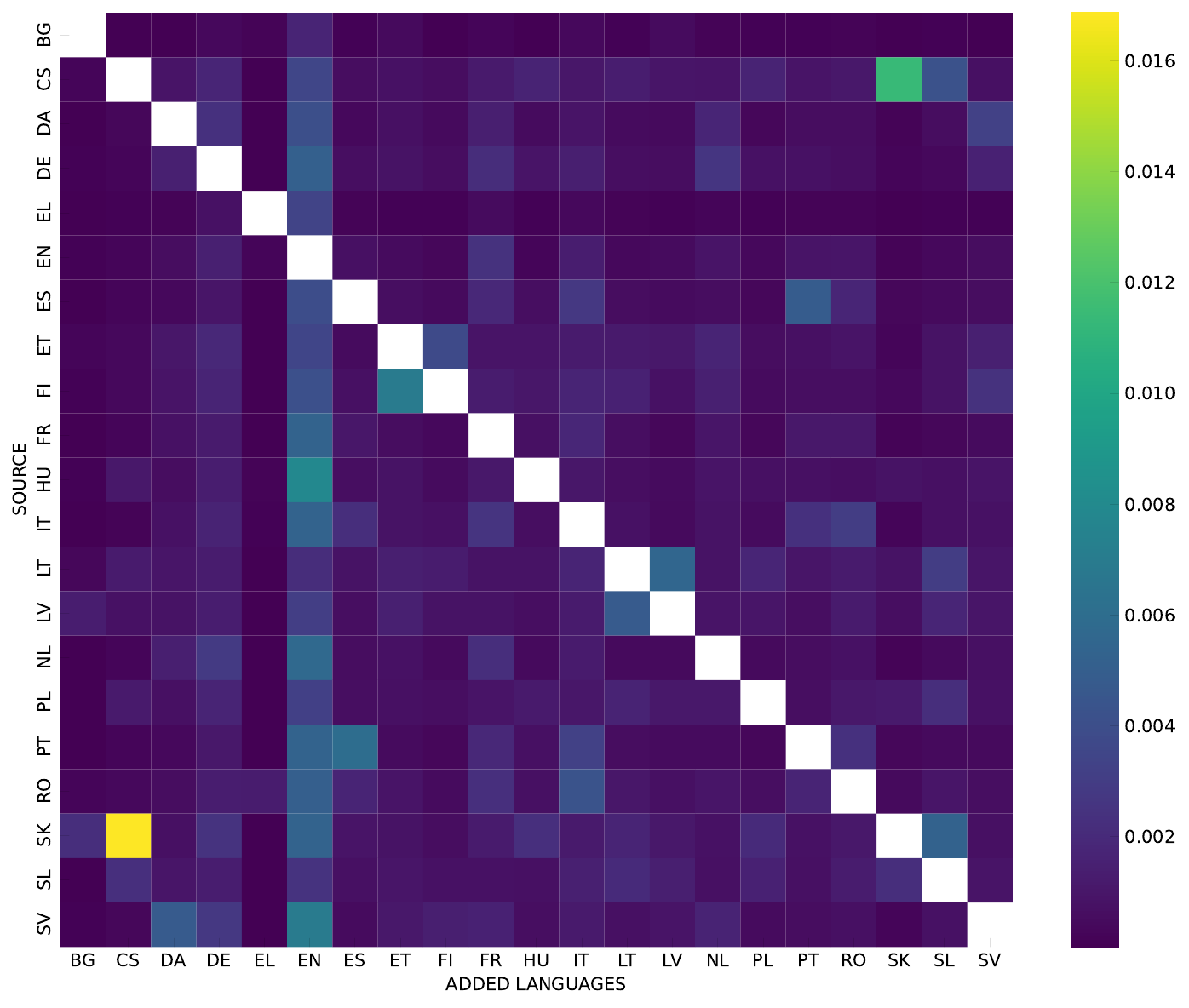}
    \caption{For each \emph{Source} language, we compute the proportion of tokens that are exclusively assigned to each \emph{Added Language} when tokenizing a text from the \emph{Source} language with the \emph{full tokenizer}. } 
    \label{fig:inserted_tokens}
\end{figure*}

\section{Supplementary Ablations}

\subsection{Language Origin of Tokens Produced by the Full Tokenizer}
\label{appendix:inserted_tokens}
To determine the language origin of the tokens introduced during tokenization when processing text in a \emph{Source} language with the \emph{full tokenizer}, we select a subset of the training data for each source language. In Fig.~\ref{fig:inserted_tokens}, for each \emph{Source} language, we compute the proportion of tokens that are exclusively assigned to each \emph{Added Language} when tokenizing the text using the \emph{full tokenizer}. We observe that English introduces the largest proportion of tokens across the considered languages, mainly due to its dominance in multilingual corpora.

\subsection{Impact of the Sampling Probability}

Tabs.~\ref{tab:mcq_p_sampling} and \ref{tab:translation_p_sampling} report the performance of $\Full$ across different values of $p$, the probability of sampling the \emph{full tokenizer} during training; the following conclusions also hold for \Data{n}.

As observed in \textsection~\ref{sec:preliminary_analysis}, $p=0$ matches the baseline performance on MCQ tasks across all languages when inference is performed using the corresponding monolingual subtokenizer. However, this setting struggles to handle combinations of subtokens for MCQ and performs even worse on translation.

The case $p=1$ corresponds to a setting similar to the baseline, where the model is trained exclusively using the \emph{full tokenizer}. We observe baseline-level performance on translation when using both the \emph{full tokenizer} and the combination (\emph{EN + LANG}). The modularity of our tokenizer allows us to select the most relevant tokens for each language, leading to strong performance. However, a performance gap remains on MCQ tasks between the \emph{full tokenizer} and the monolingual subtokenizers. This highlights the importance of sampling monolingual subtokenizers during training, in order to enable the model to handle arbitrary combinations of subtokenizers at inference time. In practice, training in this setting is computationally challenging due to the large vocabulary size, especially when scaling to many languages.

We also observe that lower values of $p$ favor better performance when using monolingual subtokenizers, while higher values improve performance when using the \emph{full tokenizer} (which serves as a proxy for inference with combined subtokenizers). We therefore selected $p=0.5$ in our experiments to balance these two effects. 

Additionally, lower values of $p$ can help reduce training costs, as they decrease the proportion of batches in which logits must be computed over combined vocabularies, particularly for large values of $n$, and instead restrict computation to smaller, language-specific vocabularies. However, this effect is dependent on model size and is more pronounced in smaller-scale models, where the cost of computing logits represents a larger fraction of the overall computation.

\begin{table*}[t]
\centering
\renewcommand{\arraystretch}{1.5} 
\setlength{\tabcolsep}{3pt} 
\footnotesize
\begin{tabular}{cccccccccccccccccccccc}
\midrule
\textbf{Lang}. & \textbf{BG} & \textbf{CS} & \textbf{DA} & \textbf{DE} & \textbf{EL} & \textbf{EN}
& \textbf{ES} & \textbf{ET} & \textbf{FI} & \textbf{FR} & \textbf{HU} 
& \textbf{IT} & \textbf{LT} & \textbf{LV} & \textbf{NL} & \textbf{PL} 
& \textbf{PT} & \textbf{RO} & \textbf{SK} & \textbf{SL} & \textbf{SV} \\
\midrule
$p$ (\%) & 0.5 & 3.5 & 1.9 & 2.2 & 0.7 & 1.3 & 2.2 & 2.3 & 2.8 & 1.7 & 2.2 
 & 2.4 & 2.8 & 2.5 & 2.0 & 2.2 & 2.5 & 2.6 & 4.8 & 2.3 & 2.7 \\

\bottomrule
\end{tabular}

\caption{Percentage of tokens originating from other languages when tokenizing text in each reported language using the full tokenizer.}
\label{tab:token_language_proportions}
\end{table*}

\begin{table*}[t]
\centering
\scriptsize
\setlength{\tabcolsep}{3.5pt}
\renewcommand{\arraystretch}{1.2}
\centering

\begin{tabular}{c c ccccccccccccccccccccc}
\toprule
 
& Infer. & \textbf{BG} & \textbf{CS} & \textbf{DA} & \textbf{DE} & \textbf{EL} & \textbf{EN}
& \textbf{ES} & \textbf{ET} & \textbf{FI} & \textbf{FR} & \textbf{HU} 
& \textbf{IT} & \textbf{LT} & \textbf{LV} & \textbf{NL} & \textbf{PL} 
& \textbf{PT} & \textbf{RO} & \textbf{SK} & \textbf{SL} & \textbf{SV} \\
\midrule

\multirow{2}{*}{Baseline}
& Lang & 37.3 & 38.8 & 40.4 & 37.9 & 37.3 & 52.9 & 39.5 & 36.6 & 37.0 & 39.5 & 36.6 & 38.8 & 37.4 & 37.7 & 38.7 & 37.4 & 38.3 & 38.6 & 38.1 & 37.8 & 39.4 \\
& Full & 37.3 & 38.8 & 40.4 & 37.9 & 37.3 & 52.9 & 39.5 & 36.6 & 37.0 & 39.5 & 36.6 & 38.8 & 37.4 & 37.7 & 38.7 & 37.4 & 38.3 & 38.6 & 38.1 & 37.8 & 39.4 \\
\midrule

\multirow{2}{*}{0}
& Lang & 38.3 & \textbf{38.8} & 40.5 & 37.7 & 37.9 & 52.6 & 39.9 & 37.3 & \textbf{37.9} & 40.3 & 37.0 & 38.9 & 37.4 & \textbf{38.2} & 39.4 & 37.9 & 38.4 & 38.5 & \textbf{38.9} & 37.7 & 39.7 \\
& Full & 36.3 & 36.6 & 38.3 & 36.3 & 37.1 & 49.6 & 38.0 & 34.4 & 34.2 & 37.8 & 35.3 & 37.2 & 34.6 & 35.1 & 37.3 & 35.1 & 37.7 & 36.2 & 36.1 & 36.0 & 38.2 \\
\midrule

\multirow{2}{*}{0.1}
& Lang & \textbf{38.7} & 38.4 & 41.3 & 38.2 & \textbf{38.1} & 52.4 & \textbf{40.0} & 37.0 & 37.7 & 39.3 & \textbf{37.5} & 38.8 & 37.5 & 37.6 & 39.4 & 37.7 & 38.4 & \textbf{39.1} & 38.3 & 37.8 & 39.6 \\
& Full & 38.2 & 38.0 & 40.6 & 37.5 & 37.5 & 52.1 & 39.0 & 35.8 & 36.5 & 38.9 & 36.6 & 38.1 & 35.9 & 36.1 & 38.6 & 36.6 & 38.0 & 37.8 & 37.8 & 36.9 & 38.9 \\
\midrule

\multirow{2}{*}{0.2}
& Lang & 38.0 & 37.6 & 39.7 & 37.4 & 37.7 & 52.2 & 39.4 & 36.5 & 36.9 & 39.2 & 36.5 & 38.2 & 37.3 & 37.4 & 38.7 & 37.5 & 38.2 & 37.7 & 38.0 & 37.5 & 39.3 \\
& Full & 37.5 & 37.9 & 39.1 & 37.3 & 36.6 & 51.8 & 38.1 & 35.8 & 35.9 & 38.8 & 35.9 & 37.5 & 36.7 & 36.4 & 37.8 & 36.8 & 37.2 & 37.6 & 37.8 & 36.9 & 38.8 \\
\midrule

\multirow{2}{*}{0.5}
& Lang & 38.1 & 38.4 & \textbf{41.6} & 38.4 & 38.0 & 53.2 & 39.8 & \textbf{37.6} & 37.6 & 39.3 & 36.9 & \textbf{39.1} & 38.1 & 38.2 & \textbf{39.7} & \textbf{38.1} & \textbf{38.8} & 38.5 & 38.3 & 38.2 & 40.1 \\
& Full & 38.0 & 38.2 & 41.5 & 38.1 & 37.9 & 53.0 & 39.7 & 36.7 & 37.6 & 39.4 & 36.8 & 38.8 & \textbf{38.1} & 37.6 & 39.3 & 37.8 & 38.2 & 38.2 & 38.2 & 37.8 & 39.7 \\
\midrule

\multirow{2}{*}{0.8}
& Lang & 38.5 & 38.1 & 40.3 & 37.9 & 37.5 & 52.9 & 39.5 & 36.7 & 37.2 & 39.4 & 36.7 & 38.7 & 37.4 & 37.6 & 38.9 & 36.9 & 38.4 & 38.2 & 38.5 & \textbf{38.2} & \textbf{40.4} \\
& Full & 38.6 & 38.6 & 40.8 & 38.0 & 36.9 & \textbf{53.5} & 39.7 & 36.2 & 37.3 & \textbf{41.6} & 36.3 & 38.7 & 37.7 & 37.9 & 39.6 & 37.5 & 38.4 & 38.5 & 38.7 & 38.0 & 40.0 \\
\midrule

\multirow{2}{*}{1}
& Lang & 37.4 & 35.5 & 38.8 & 36.6 & 37.3 & 50.3 & 37.0 & 34.5 & 34.7 & 37.6 & 35.2 & 36.2 & 34.8 & 35.6 & 37.1 & 35.2 & 35.9 & 36.2 & 34.9 & 35.3 & 37.7 \\
& Full & 37.4 & 38.8 & 40.8 & \textbf{38.5} & 37.9 & 52.5 & 39.8 & 36.5 & 37.3 & 39.5 & 36.6 & 38.3 & 37.1 & 37.9 & 38.9 & 37.1 & 38.5 & 38.7 & 38.1 & 37.7 & 39.9 \\

\bottomrule
\end{tabular}
\caption{MCQ Performance of $\Full$ for various probabilities of sampling the full tokenizer (Unigram).}
\label{tab:mcq_p_sampling}
\end{table*}

\begin{table*}[!t]
\centering
\scriptsize

\setlength{\tabcolsep}{3.5pt}
\renewcommand{\arraystretch}{1.2}
\centering

\begin{tabular}{c c ccccccccccccccccccccc}
\toprule
 
& Infer. & \textbf{BG} & \textbf{CS} & \textbf{DA} & \textbf{DE} & \textbf{EL} & \textbf{EN}
& \textbf{ES} & \textbf{ET} & \textbf{FI} & \textbf{FR} & \textbf{HU} 
& \textbf{IT} & \textbf{LT} & \textbf{LV} & \textbf{NL} & \textbf{PL} 
& \textbf{PT} & \textbf{RO} & \textbf{SK} & \textbf{SL} & \textbf{SV} \\
\midrule
\multirow{2}{*}{Baseline}
& Lang & 26.1 & 22.0 & 35.3 & 22.7 & 17.1 & - & 20.6 & 15.8 & 14.6 & 32.4 & 14.6 & 20.1 & 17.7 & 19.0 & 20.6 & 14.7 & 34.5 & 28.2 & 21.0 & 20.0 & 34.4 \\
& Full & 26.1 & 22.0 & 35.3 & 22.7 & 17.1 & - & 20.6 & 15.8 & 14.6 & 32.4 & 14.6 & 20.1 & 17.7 & 19.0 & 20.6 & 14.7 & 34.5 & 28.2 & 21.0 & 20.0 & 34.4 \\
\midrule

\multirow{2}{*}{0}
& Lang & 1.0 & 1.8 & 2.8 & 2.6 & 5.3 & - & 6.2 & 1.5 & 1.1 & 15.0 & 1.5 & 6.9 & 1.0 & 1.0 & 2.6 & 1.4 & 5.2 & 9.3 & 1.6 & 2.0 & 2.6 \\
& Full & 0.9 & 1.5 & 2.4 & 2.1 & 2.5 & - & 2.8 & 1.1 & 0.8 & 7.4 & 0.9 & 3.0 & 1.0 & 0.9 & 2.3 & 1.0 & 4.5 & 4.8 & 1.4 & 2.2 & 2.3 \\
\midrule

\multirow{2}{*}{0.1}
& Lang & 23.3 & 19.3 & 33.3 & 19.9 & 16.4 & - & 18.3 & 13.5 & 12.8 & 29.5 & 12.1 & 18.6 & 13.6 & 15.6 & 18.9 & 13.0 & 29.4 & 26.4 & 18.5 & 17.7 & 31.9 \\
& Full & 23.7 & 19.4 & 33.1 & 20.1 & 16.4 & - & 18.6 & 14.7 & 13.2 & 29.3 & 11.9 & 18.7 & 15.7 & 17.5 & 18.8 & 13.2 & 30.9 & 26.0 & 18.8 & 18.0 & 31.8 \\
\midrule

\multirow{2}{*}{0.2}
& Lang & 26.0 & 19.9 & 33.9 & 21.6 & 16.1 & - & 19.6 & 15.4 & 14.0 & 30.8 & 13.9 & 18.6 & 16.0 & 17.1 & 19.2 & 12.2 & 31.1 & 26.3 & 19.8 & 18.2 & 33.2 \\
& Full & 25.7 & 20.3 & 34.3 & 21.5 & 16.1 & - & 19.7 & 16.3 & 13.9 & 30.5 & 14.2 & 18.7 & 16.6 & 18.7 & 19.3 & 13.5 & 31.8 & 26.5 & 20.0 & 18.5 & 33.2 \\
\midrule

\multirow{2}{*}{0.5}
& Lang & \textbf{27.9} & 20.2 & 35.1 & 21.1 & 16.9 & - & 20.6 & 15.9 & 14.9 & 32.5 & 13.9 & 19.8 & 16.9 & 18.6 & 19.9 & 14.0 & 33.3 & 28.3 & 20.3 & 19.2 & 33.8 \\
& Full & 27.7 & 20.5 & 35.2 & 21.1 & 16.8 & - & 20.8 & 16.2 & 15.0 & 32.6 & 13.9 & 20.2 & 17.8 & 19.5 & 20.1 & 14.3 & 33.7 & 28.4 & 21.1 & 18.9 & 34.2 \\
\midrule

\multirow{2}{*}{0.8}
& Lang & 27.6 & 20.6 & 34.1 & 22.5 & 17.2 & - & 20.3 & 16.8 & 14.3 & 31.5 & 14.7 & 19.3 & 17.2 & 18.3 & 19.5 & 14.2 & 33.1 & 27.5 & 19.6 & 19.3 & 32.6 \\
& Full & 27.4 & 21.2 & 34.8 & 22.8 & 17.1 & - & 20.7 & 17.0 & 15.1 & 31.7 & 14.7 & 19.8 & 17.5 & 19.2 & 19.7 & 14.3 & 33.7 & 28.4 & 20.7 & 19.6 & 33.0 \\
\midrule

\multirow{2}{*}{1}
& Lang & 27.5 & 20.6 & 36.0 & 22.7 & 17.3 & - & 20.6 & 17.1 & 14.6 & 32.8 & 15.3 & 20.1 & 17.3 & 19.0 & 20.0 & 13.8 & 33.8 & 28.1 & 20.0 & 18.8 & 34.4 \\
& Full & 27.3 & \textbf{22.1} & \textbf{37.1} & \textbf{23.2} & \textbf{17.4} & - & \textbf{21.1} & \textbf{17.5} & \textbf{15.4} & \textbf{34.0} & \textbf{15.6} & \textbf{20.7} & \textbf{19.1} & \textbf{20.1} & \textbf{20.7} & \textbf{14.9} & \textbf{35.0} & \textbf{29.4} & \textbf{22.1} & \textbf{20.3} & \textbf{35.5} \\

\bottomrule
\end{tabular}
\caption{MT Performance of $\Full$ for various probabilities of sampling the full tokenizer (Unigram).}
\label{tab:translation_p_sampling}
\end{table*}

\begin{table*}[t]
\centering
\footnotesize 
\setlength{\tabcolsep}{7pt} 
\renewcommand{\arraystretch}{1.3}

\begin{tabular}{l c c c c c}
\toprule
Model (batch) & $K$ & fwd/bwd & off-GPU & xfer & step w/ offload \\
\midrule
1B (128)    & 1 & 1.37 & 0.32 & 0.05 & 1.43 \\
1B (128)    & 6 & 1.53--1.55 & 1.15--1.24 & 0.15 & 1.72--1.74 \\
\midrule
2.5B (64)   & 1 & 1.47 & 0.54 & 0.07 & 1.55 \\
2.5B (64)   & 6 & 1.59 & 1.63--1.88 & 0.23 & 2.03--2.38 \\
2.5B (128)  & 1 & 2.90 & 0.31--0.51 & 0.06 & 2.96--3.00 \\
2.5B (128)  & 6 & 3.13 & 2.20--2.65 & 0.21--0.26 & 3.76--3.89 \\
\midrule
7B (64)     & 1 & 3.62 & 0.65--0.83 & 0.10 & 3.73--3.76 \\
7B (64)     & 6 & 3.77--3.80 & 1.98--2.19 & 0.30 & 4.09--4.21 \\
7B (128)    & 1 & 7.11--7.14 & 0.65 & 0.10 & 7.23--7.24 \\
7B (128)    & 6 & 7.39--7.46 & 3.71--4.72 & 0.32--0.39 & 7.82--7.85 \\
\bottomrule

\end{tabular}
\captionof{table}{Offloading proof of concept (seconds/step, median of 12; ranges over repeat runs on different nodes). \emph{fwd/bwd}: step time with all state GPU-resident; \emph{off-GPU}: the vocabulary round trip alone; \emph{xfer}: its two row transfers; \emph{step w/ offload}: step time with the round trip pipelined in. $K{=}1$: monolingual batch; $K{=}6$: \Data{5}.}
\label{tab:offload_poc}

\end{table*}

\section{Offloading Vocabulary State to CPU: Proof of Concept}
\label{appendix:offloading}
\paragraph{Setup.} We benchmark our real training step — FSDP-sharded forward/backward with flash attention, activation rematerialization, and AdamW — on one node of 8$\times$H100, at 1B, 2.5B, and a 7B configuration ($d{=}4096$, 32 layers). The vocabulary state resides in CPU RAM: an fp16 table holding the embedding rows of all 21 languages, with fp32 Adam moments and master weights host-side. A step activates the batch's input rows (${\sim}$24k, one language) plus the output rows of the $K$ sampled subtokenizers; we measure $K{=}1$ (monolingual batches, half of training at $p{=}0.5$) and $K{=}6$ (\Data{5}). The host side is deliberately naive: unpinned memory, threaded-NumPy AdamW. We report batch 64 — the configuration our full-vocabulary training was memory-constrained to — and batch 128, which the freed vocabulary memory makes affordable.

\paragraph{Pipeline.} Following ZeRO-Offload's delayed parameter update \citep{ren2021zerooffloaddemocratizingbillionscalemodel}: while the GPU computes step $t$, the CPU gathers the active rows, applies AdamW using the gradients of step $t{-}1$, and writes them back to the host table; the vocabulary rows thus lag one gradient step, a scheme whose convergence neutrality is established by \citet{ren2021zerooffloaddemocratizingbillionscalemodel}. The backbone updates synchronously on GPU. In our prototype the two row transfers (gradients to the CPU, updated parameter rows back) run between steps, since copies issued while a step is in flight are stream-ordered behind its kernels; production implementations overlap them with compute through pinned buffers and dedicated copy streams \citep{ren2021zerooffloaddemocratizingbillionscalemodel,miao_2021}.

\paragraph{Results.} Tab.~\ref{tab:offload_poc} reports, per configuration, the step time, the full off-GPU round trip measured alone (gather, both transfers, CPU update, write-back), its transfer share, and the step time with the round trip pipelined in. The round trip fits within the step at every batch-128 configuration and every monolingual one; the sole overflow is \Data{5} at the memory-constrained 2.5B/batch-64 setting, where the naive CPU update alone (1.4--1.7\,s) exceeds the 1.6\,s step — the component a fused CPU optimizer accelerates several-fold. Transfers are 10--15\% of the round trip throughout (0.05--0.39\,s): the off-GPU cost is dominated by CPU compute, which the pipeline hides, not by data movement.

\paragraph{Caveats.} These timings upper-bound the overhead: no pinned memory, no fused CPU optimizer, no transfer/compute overlap. Host-side times vary up to ${\sim}2\times$ across nodes without NUMA pinning (the upper ends of the off-GPU ranges). The delayed update applies to the vocabulary rows only.

\section{Inference and Memory Efficiency Across Scales: Details}
\label{appendix:scale_efficiency}

\begin{table*}[t]
\footnotesize\centering
\renewcommand{\arraystretch}{1.2}
\setlength{\tabcolsep}{2.8pt}
\begin{tabular}{c @{\hspace{13pt}} cccccc @{\hspace{20pt}} cccccc @{\hspace{20pt}} cccccc}


\toprule
 & \multicolumn{6}{c}{\textbf{Gemma-3 (262k)}} & \multicolumn{6}{c}{\textbf{Qwen-3 (152k)}} & \multicolumn{6}{c}{\textbf{Llama-3 (128k)}} \\
Lang & 0.5B & 1B & 2.5B & 7B & 12B & 30B & 0.5B & 1B & 2.5B & 7B & 12B & 30B & 0.5B & 1B & 2.5B & 7B & 12B & 30B \\
\midrule
BG & 2.15 & 1.97 & 1.82 & 1.63 & 1.59 & 1.51 & 2.46 & 2.33 & 2.22 & 2.07 & 2.03 & 2.00 & 2.07 & 1.99 & 1.92 & 1.84 & 1.81 & 1.83 \\
CS & 2.10 & 1.92 & 1.76 & 1.58 & 1.56 & 1.46 & 2.40 & 2.27 & 2.15 & 2.02 & 2.00 & 1.91 & 1.60 & 1.53 & 1.47 & 1.41 & 1.42 & 1.39 \\
DE & 1.72 & 1.58 & 1.45 & 1.31 & 1.28 & 1.20 & 1.70 & 1.61 & 1.52 & 1.43 & 1.42 & 1.36 & 1.57 & 1.51 & 1.44 & 1.40 & 1.39 & 1.37 \\
EL & 2.63 & 2.41 & 2.22 & 1.98 & 1.97 & 1.85 & 5.03 & 4.74 & 4.51 & 4.19 & 4.15 & 4.03 & 2.08 & 1.99 & 1.91 & 1.82 & 1.84 & 1.83 \\
EN & 1.49 & 1.37 & 1.26 & 1.12 & 1.11 & 1.05 & 1.24 & 1.17 & 1.11 & 1.03 & 1.03 & 1.00 & 1.15 & 1.10 & 1.06 & 1.00 & 1.01 & 1.01 \\
FR & 1.66 & 1.52 & 1.40 & 1.25 & 1.22 & 1.17 & 1.57 & 1.48 & 1.41 & 1.31 & 1.31 & 1.26 & 1.45 & 1.38 & 1.33 & 1.28 & 1.27 & 1.27 \\
HR & 2.05 & 1.88 & 1.73 & 1.55 & 1.52 & 1.43 & 2.15 & 2.03 & 1.93 & 1.80 & 1.78 & 1.72 & 1.90 & 1.81 & 1.75 & 1.68 & 1.67 & 1.66 \\
HU & 2.25 & 2.06 & 1.90 & 1.69 & 1.67 & 1.59 & 2.40 & 2.27 & 2.16 & 2.01 & 2.00 & 1.92 & 2.18 & 2.08 & 2.00 & 1.93 & 1.92 & 1.90 \\
NE & 2.21 & 2.03 & 1.88 & 1.66 & 1.64 & 1.54 & 5.37 & 5.07 & 4.83 & 4.51 & 4.48 & 4.29 & 2.85 & 2.74 & 2.63 & 2.51 & 2.51 & 2.49 \\
PL & 2.03 & 1.86 & 1.71 & 1.53 & 1.52 & 1.41 & 2.01 & 1.90 & 1.80 & 1.68 & 1.66 & 1.61 & 1.95 & 1.88 & 1.80 & 1.73 & 1.73 & 1.69 \\
RO & 1.98 & 1.81 & 1.68 & 1.50 & 1.47 & 1.39 & 1.97 & 1.86 & 1.77 & 1.65 & 1.65 & 1.58 & 1.80 & 1.73 & 1.66 & 1.60 & 1.58 & 1.58 \\
SI & 2.72 & 2.50 & 2.29 & 2.04 & 2.03 & 1.92 & 6.95 & 6.56 & 6.21 & 5.82 & 5.76 & 5.58 & 7.82 & 7.47 & 7.15 & 6.88 & 6.89 & 6.85 \\
SL & 2.15 & 1.98 & 1.81 & 1.62 & 1.59 & 1.51 & 2.16 & 2.04 & 1.94 & 1.80 & 1.79 & 1.75 & 1.90 & 1.83 & 1.74 & 1.68 & 1.67 & 1.65 \\
SO & 2.36 & 2.17 & 1.99 & 1.78 & 1.75 & 1.65 & 2.26 & 2.13 & 2.03 & 1.89 & 1.87 & 1.82 & 2.05 & 1.97 & 1.89 & 1.82 & 1.80 & 1.80 \\
SW & 2.39 & 2.19 & 2.03 & 1.81 & 1.77 & 1.67 & 2.39 & 2.26 & 2.15 & 2.02 & 2.00 & 1.93 & 2.17 & 2.08 & 2.00 & 1.93 & 1.90 & 1.90 \\
TE & 2.46 & 2.26 & 2.09 & 1.84 & 1.82 & 1.73 & 8.18 & 7.72 & 7.34 & 6.86 & 6.70 & 6.56 & 8.69 & 8.31 & 7.97 & 7.65 & 7.57 & 7.64 \\
\midrule
 & \multicolumn{6}{c}{\textbf{TinyAya (261k)}} & \multicolumn{6}{c}{\textbf{Full (387k, ours)}} & \multicolumn{6}{c}{\textbf{Extracted 24k (ours)}} \\
\midrule
BG & 1.92 & 1.76 & 1.63 & 1.49 & 1.46 & 1.47 & 1.62 & 1.45 & 1.32 & 1.15 & 1.12 & 1.10 & 1.00 & 1.00 & 1.00 & 1.00 & 0.99 & 1.01 \\
CS & 1.71 & 1.56 & 1.44 & 1.32 & 1.31 & 1.29 & 1.55 & 1.39 & 1.25 & 1.11 & 1.09 & 1.05 & 1.00 & 1.00 & 1.00 & 1.00 & 1.01 & 1.00 \\
DE & 1.55 & 1.42 & 1.31 & 1.22 & 1.19 & 1.17 & 1.58 & 1.42 & 1.27 & 1.13 & 1.10 & 1.06 & 1.02 & 1.02 & 1.02 & 1.03 & 1.02 & 1.02 \\
EL & 1.99 & 1.82 & 1.67 & 1.54 & 1.51 & 1.50 & 1.60 & 1.43 & 1.29 & 1.13 & 1.11 & 1.08 & 1.00 & 1.00 & 1.00 & 0.99 & 1.00 & 1.00 \\
EN & 1.37 & 1.25 & 1.15 & 1.06 & 1.05 & 1.04 & 1.58 & 1.42 & 1.28 & 1.12 & 1.11 & 1.07 & 1.01 & 1.01 & 1.01 & 1.01 & 1.00 & 1.01 \\
FR & 1.48 & 1.36 & 1.25 & 1.14 & 1.14 & 1.13 & 1.58 & 1.42 & 1.28 & 1.12 & 1.10 & 1.07 & 1.01 & 1.00 & 1.01 & 1.00 & 1.00 & 1.01 \\
HR & 1.78 & 1.62 & 1.51 & 1.38 & 1.36 & 1.35 & 1.54 & 1.38 & 1.25 & 1.10 & 1.06 & 1.04 & 1.03 & 1.03 & 1.04 & 1.04 & 1.04 & 1.04 \\
HU & 1.92 & 1.76 & 1.62 & 1.49 & 1.46 & 1.45 & 1.59 & 1.43 & 1.29 & 1.13 & 1.11 & 1.08 & 1.03 & 1.03 & 1.03 & 1.03 & 1.03 & 1.03 \\
NE & 1.98 & 1.81 & 1.67 & 1.54 & 1.52 & 1.50 & 1.59 & 1.42 & 1.28 & 1.11 & 1.11 & 1.07 & 1.01 & 1.02 & 1.02 & 1.01 & 1.02 & 1.01 \\
PL & 1.84 & 1.68 & 1.55 & 1.43 & 1.42 & 1.38 & 1.57 & 1.42 & 1.27 & 1.12 & 1.10 & 1.06 & 1.03 & 1.03 & 1.03 & 1.03 & 1.04 & 1.03 \\
RO & 1.75 & 1.61 & 1.47 & 1.36 & 1.35 & 1.32 & 1.57 & 1.41 & 1.27 & 1.12 & 1.09 & 1.06 & 1.02 & 1.02 & 1.02 & 1.03 & 1.02 & 1.02 \\
SI & 12.7 & 11.6 & 10.6 & 9.78 & 9.71 & 9.65 & 1.55 & 1.39 & 1.25 & 1.09 & 1.09 & 1.05 & 1.00 & 1.00 & 1.00 & 1.00 & 1.00 & 1.00 \\
SL & 1.82 & 1.67 & 1.52 & 1.41 & 1.39 & 1.39 & 1.55 & 1.39 & 1.25 & 1.10 & 1.07 & 1.05 & 1.04 & 1.04 & 1.04 & 1.04 & 1.04 & 1.04 \\
SO & 2.26 & 2.06 & 1.90 & 1.75 & 1.72 & 1.72 & 1.57 & 1.41 & 1.27 & 1.11 & 1.09 & 1.06 & 1.02 & 1.02 & 1.02 & 1.02 & 1.02 & 1.02 \\
SW & 1.96 & 1.79 & 1.65 & 1.52 & 1.49 & 1.49 & 1.61 & 1.44 & 1.30 & 1.15 & 1.12 & 1.09 & 1.03 & 1.03 & 1.03 & 1.03 & 1.03 & 1.03 \\
TE & 2.35 & 2.15 & 1.99 & 1.82 & 1.78 & 1.78 & 1.58 & 1.41 & 1.27 & 1.12 & 1.09 & 1.07 & 1.00 & 1.00 & 1.00 & 1.00 & 0.99 & 1.01 \\
\bottomrule
\end{tabular}
\caption{Inference-time grid: forward time on FLORES relative to a dedicated 24k monolingual tokenizer, per language and model scale (configurations in Tab.~\ref{tab:mem_params_app}).}
\label{tab:exec_scale_full}
\end{table*}

\begin{table}[t]
\footnotesize\centering
\renewcommand{\arraystretch}{1.6}
\setlength{\tabcolsep}{3.5pt}
\begin{tabular}{l cccccc}
\toprule
Scale & 0.5B & 1B & 2.5B & 7B & 12B & 30B \\
\midrule
Width $d$ & 1536 & 2048 & 3072 & 4096 & 5120 & 7168 \\
Layers    & 19 & 21 & 24 & 40 & 40 & 52 \\
\midrule
\multicolumn{7}{l}{Total-parameter reduction, \emph{256k $\to$ 24k vocab.:}} \\

\quad untied emb. & 55\% & 47\% & 35\% & 20\% & 16\% & 10\% \\
\quad tied emb.   & 40\% & 31\% & 21\% & 11\% & 9\% & 5\% \\
\bottomrule
\end{tabular}
\caption{Benchmark configurations (scales named by non-embedding parameter count; head dim.\ 128, GQA~2, MLP $2.75d$; 2.5B matches our trained models) and the total-parameter reduction when serving a language with its 24k sub-vocabulary instead of a 256k one, for untied and tied embeddings.}

\label{tab:mem_params_app}
\end{table}

\paragraph{Setup.} Inference time is the mean forward pass over the full FLORES corpus, context 4096, batch 1, bf16, on H100 GPUs right-sized per scale: one GPU up to 12B, two (tensor-parallel) at 30B, where the monolingual reference runs on the same hardware, keeping ratios comparable. Model configurations are given in Tab.~\ref{tab:mem_params_app}; the 2.5B configuration matches our trained models. Tokenizers: Gemma-3 (262k), Qwen-3 (152k), \emph{Full} (our sequential multilingual BPE, 387k), and \emph{Extracted} (its per-language 24k subtokenizers), all against dedicated 24k monolingual BPE references.We measure 16 of the 21 languages: the six low-resource ones and ten from the conservative set; the five omitted (\textsc{es, pt, it, da, nl}) fall within the reported range, and their ratios at scale follow from their NSL (see below).

\paragraph{Inference time.}\emph{Full}'s overhead is nearly language-independent — within $\pm 0.04$ of each scale's mean, reflecting its pure output-layer cost — and decays from ${\sim}1.58\times$ to ${\sim}1.07\times$ (Tab.~\ref{tab:exec_scale_full}); open-weight overheads contain an additional compression factor that persists at every scale and is largest on low-resource languages and non-Latin scripts. At 30B, where the output-layer factor has faded, the normalized time converges to the tokenizer's NSL (Gemma-3 on \textsc{te}: $1.73$ vs.\ an NSL of $1.61$; on \textsc{si}: $1.92$ vs.\ $1.79$) — Tab.~\ref{tab:nsl_lowres} thus predicts large-scale inference cost for any language. \emph{Extracted} matches the monolingual references ($0.99$--$1.04$) everywhere.

\paragraph{Memory.} For parameters (Tab.~\ref{tab:mem_params_app}): a 24k sub-vocabulary instead of a 256k one removes 90.6\% of the embedding parameters at any width; the share of \emph{total} parameters saved shrinks as the backbone grows, and is lower with tied embeddings (last two rows). For compression (Tab.~\ref{tab:token_per_char}): comparing an open-weight row with the \emph{Extracted} row gives the sequence-length cut, which equally cuts decode steps and KV cache: on \textsc{hr}, Gemma-3 costs 0.314 tokens per character and ours 0.242, a 23\% cut. Cuts span $-3\%$ (\textsc{en}) to 44\% (low-resource). These cuts hold at every size and architecture; only the per-token cache cost varies (58--745\,KB from 0.5B to 30B; bf16, GQA~2, head dim.\ 128).

\end{document}